\documentclass[]{youtu}

\usepackage[nohelv]{newpxtext}
\usepackage{graphicx}
\usepackage[numbers]{natbib}
\newcommand{\todo}[1]{{\color{red}#1}}
\newcommand{\TODO}[1]{\textbf{\color{red}[TODO: #1]}}
\renewcommand{\TODO}[1]{}
\renewcommand{\todo}[1]{}

\usepackage{xurl}

\usepackage{subcaption}
\usepackage{caption}
\usepackage{newfloat}
\usepackage{listings}
\usepackage{textcomp}
\usepackage{xspace}

\usepackage{graphbox}
\usepackage{graphicx}
\usepackage{multirow}
\usepackage{tabularx}
\usepackage{wrapfig}
\usepackage{tikz}
\usepackage{float}
\usepackage{algorithm}
\usepackage{algpseudocode}

\usepackage{adjustbox}
\usepackage{booktabs}
\usepackage{tabularx}
\usepackage{stackengine}
\usepackage{tablefootnote}
\usepackage{threeparttable}

\usepackage{amsmath}
\usepackage{amssymb}
\usepackage{bbm}
\usepackage{dsfont}
\usepackage{mathtools}

\usepackage{color}
\usepackage{colortbl}
\usepackage[normalem]{ulem}
\usepackage{xcolor}

\tcbuselibrary{breakable,skins,listings}
\tcbset{schemaboxstyle/.style={
  breakable,
  listing only,
  listing options={
    basicstyle=\ttfamily\small,
    columns=fullflexible,
    keepspaces=true,
    breaklines=true,
    breakatwhitespace=true,
    breakindent=2em,
    extendedchars=false,
  },
  colback=white,
  colframe=gray!50,
  boxrule=0.5pt,
  arc=0pt, outer arc=0pt,
  left=1em, right=1em, top=0.6em, bottom=0.6em,
  before skip=\baselineskip,
  after skip=\baselineskip,
}}
\newtcblisting{schemabox}{schemaboxstyle,
  left skip=0.09\textwidth, right skip=0.09\textwidth}
\newtcblisting{schemaboxwide}{schemaboxstyle,
  left skip=0.07\textwidth, right skip=0.07\textwidth}

\usepackage{CJKutf8}

\newcommand{\hfilll}{\hspace{0pt plus 1filll}}

\usepackage[T1]{fontenc}
\usepackage[scale=0.9]{sourcecodepro}
\usepackage{amsmath,amssymb,booktabs,multirow,array,xcolor}
\usepackage{tikz}
\usetikzlibrary{positioning,arrows.meta,fit,calc,backgrounds,
             decorations.pathreplacing}

\makeatletter
\DeclareRobustCommand\onedot{\futurelet\@let@token\@onedot}
\def\@onedot{\ifx\@let@token.\else.\null\fi\xspace}

\makeatother

\DeclareFloatingEnvironment[fileext=lolst,name=Listing,placement=tbp]{listing}
\makeatletter

\def\OurMethod{Youtu-Parsing-Omni\xspace}

\def\OurMethodBF{\textbf{Youtu-Parsing-Omni}\xspace}

\def\OurEncoder{Youtu-Omni-Encoder\xspace}

\def\OurMethodV1{Youtu-Parsing\xspace}
\def\Logics{Logics-Parsing-Omni\xspace}
\def\OmniParsingBench{OmniParsingBench\xspace}
\def\OmniDocBench{OmniDocBench\xspace}

\makeatother

\definecolor{babyblue}{rgb}{0.54, 0.81, 0.94}
\definecolor{babyblueeyes}{rgb}{0.63, 0.79, 0.95}
\makeatother

\usepackage{pifont}
\newcommand{\yes}{{\color{black!70}\ding{51}}}
\newcommand{\no}{{\color{black!25}--}}

\usepackage[normalem]{ulem}

\makeatletter

\newif\ifrelease \releasefalse
\newif\ifpreview \previewfalse
\ifdefined\PREVIEW\previewtrue\fi
\newcommand{\tbd}[1][]{\ifrelease\errmessage{Unresolved TBD: #1}\else\ifpreview{\color{black!40}\textsf{\scriptsize[pending]}}\else{\color{red!75!black}\textbf{[TBD}\ifx\relax#1\relax\else: #1\fi\textbf{]}}\fi\fi}
\newcommand{\best}[1]{\textbf{#1}}
\newcommand{\second}[1]{\underline{#1}}

\newcommand{\oursmark}{\textcolor{youtuBlue}{\rule[-0.2ex]{2pt}{1.9ex}}\hspace{3pt}}
\newcommand{\tabhead}{\rowcolor{black!6}}
\newlength{\tabheadsavedaboverulesep}
\newlength{\tabheadsavedbelowrulesep}

\usepackage{needspace}
\pretocmd{\section}{\Needspace{6\baselineskip}}{}{}
\pretocmd{\subsection}{\Needspace{7\baselineskip}}{}{}
\pretocmd{\subsubsection}{\Needspace{5\baselineskip}}{}{}
\renewcommand{\topfraction}{0.85}
\renewcommand{\textfraction}{0.12}
\renewcommand{\floatpagefraction}{0.75}
\def\OSAD{OSAD\xspace}
\def\OmniSchema{OmniSchema\xspace}
\providecommand{\indic}{\mathds{1}}
\makeatletter
\renewcommand{\paragraph}{\@startsection{paragraph}{4}{\z@}{0.2ex \@plus .2ex}{-1em}{\normalfont\normalsize\bfseries}}
\makeatother

\newtcolorbox{appintro}{enhanced, colback=youtuBlue!3, colframe=youtuBlue!65, boxrule=0pt, leftrule=2.2pt,
  arc=0pt, outer arc=0pt, left=7pt, right=7pt, top=4pt, bottom=4pt, before skip=4pt, after skip=10pt,
  fontupper=\small}
\newtcolorbox{promptpanel}[2]{breakable, enhanced, colback=white, colframe=black!22, colbacktitle=youtuBlue!6,
  coltitle=youtuText, fonttitle=\sffamily\bfseries\small, boxrule=0.45pt, arc=1.5pt, left=6pt, right=6pt,
  top=4pt, bottom=4pt, before skip=8pt, after skip=10pt,
  title={\refstepcounter{table}\label{#2}Table~\thetable.\enspace Prompt for the #1}}
\usepackage{flafter}

\usepackage{newpxmath}

\title{\raisebox{-0.32\height}{\includegraphics[height=1.75em]{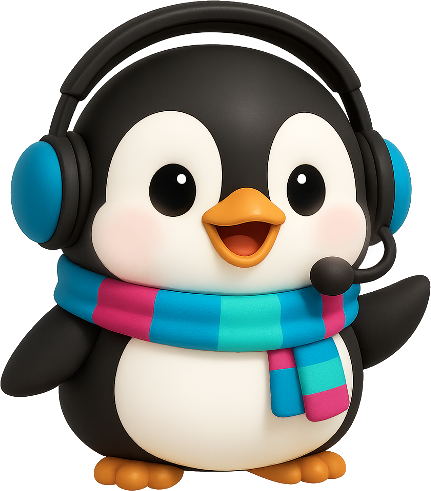}}\hspace{0.3em}\OurMethod:\\[2pt] One Encoder, One Schema, All Modalities}

\author{\OurMethod Team$^*$}

\sourcecode{https://github.com/TencentCloudADP/youtu-parsing}
\model{https://huggingface.co/tencent/Youtu-Parsing-Omni}

\begin{document}

\abstract{
    We present \textbf{Youtu-Parsing-Omni}, an omni-modal parser that converts documents, images, audio, and video into structured records grounded in space and time, providing a perception layer for agents that read, locate, and cite evidence across modalities. Its design rests on one encoder for every modality and one schema for every task. Youtu-Omni-Encoder, initialized from a pre-trained language model to strengthen text reading, encodes all modalities with shared weights in a unified space-time coordinate system, and a few fusion layers let video frames attend to concurrent audio, bringing cross-modal fusion and temporal alignment into perception itself. OmniSchema unifies the outputs of all ten parsing tasks, allowing a single self-evolving post-training procedure, OmniSchema-Aware On-Policy Distillation (OSAD), to provide a shared token-role-based objective across tasks. In each round, the model is trained on its own rollouts against token-level targets from a reference-conditioned copy of itself, which is then refreshed with the trained weights, so that the parser and its supervision co-evolve. With a single $5$\,B-parameter checkpoint, Youtu-Parsing-Omni achieves state-of-the-art performance on OmniDocBench v1.6 ($96.96$ Overall) and leads all open-weight models on OmniParsingBench ($75.08$ Avg.), with its advantage concentrated on audio and video. It also achieves the best overall chart-parsing score ($95.02$) and the highest exact chemical-fingerprint agreement among compared systems, and matches a dedicated optical-music model in character-level transcription of typeset sheet music while staying far ahead of general-purpose systems. Code and model weights are publicly available.
}

\maketitle

\begingroup
\setlength{\intextsep}{2pt}
\setlength{\parskip}{0pt}
\begin{figure}[H]
\centering
\captionsetup{skip=4pt}
\includegraphics[width=\linewidth]{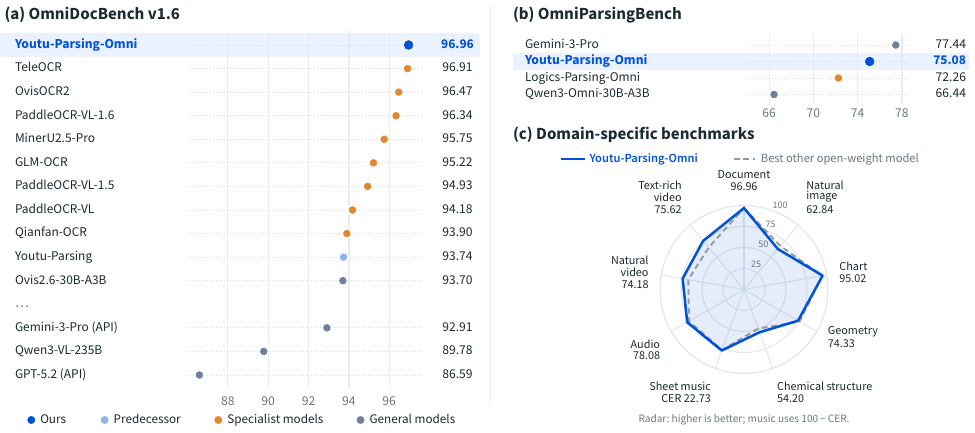}
\caption{\OurMethod{} ranks first on \OmniDocBench{} v1.6 (a), is the strongest open-weight model on \OmniParsingBench{} (b), and also leads on most domain-specific benchmarks (c).}
\label{fig:teaser}
\label{fig:teaser_omnidocbench}
\label{fig:teaser_omnimodal_parsing}
\end{figure}
\endgroup

\begin{NoHyper}
\renewcommand{\thefootnote}{*}
\footnotetext{Author list and contributions are given at the end of the main text.}
\end{NoHyper}

\vspace{-.1em}

\begin{figure}[!t]
\centering
\includegraphics[width=\linewidth]{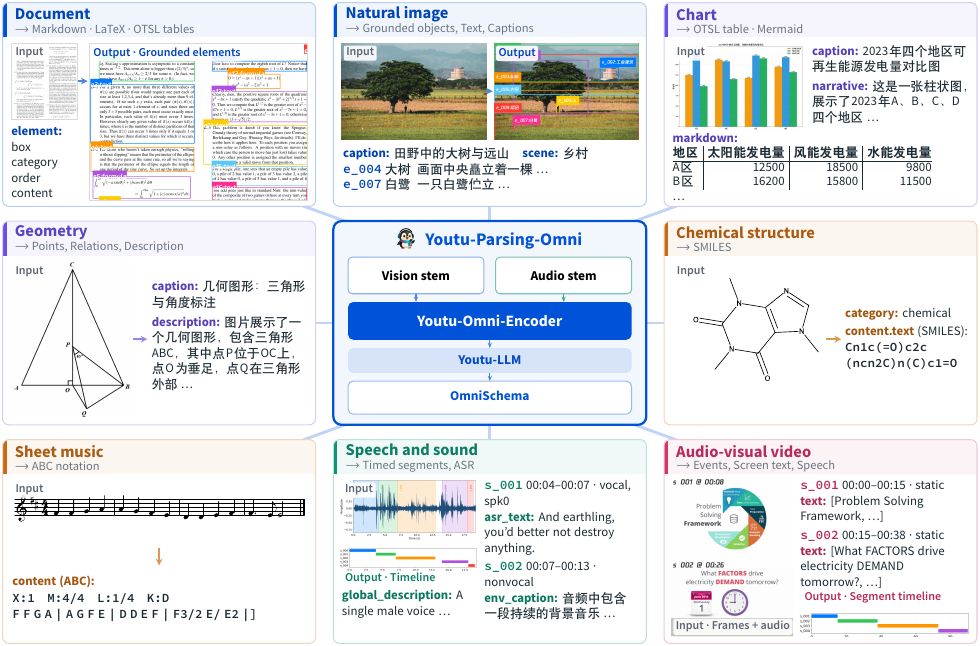}
\caption{
    \textbf{Omni-modal parsing with one encoder and one schema.}
    \OurMethod{} parses documents, natural images, audio, and video with \OurEncoder{} shared across all modalities, and unifies their parsing tasks in \OmniSchema{}.
}
\label{fig:overview}
\end{figure}

\section{Introduction}
\label{sec:intro}

Large language model (LLM) agents increasingly act on information beyond their own parameters: they retrieve evidence before answering~\citep{RAG}, call tools during reasoning~\citep{ReAct}, and operate computers through the screen~\citep{OSWorld}.
Much of this information arrives not as plain text but as documents, images, audio, and video.
Having the model that drives the agent consume such inputs directly is both \emph{costly and hard to audit}: each time the agent revisits the material, that model must re-encode long sequences of visual and acoustic tokens, and an answer drawn from raw pixels or waveforms does not by itself indicate which region or interval supports it.
\emph{Omni-modal parsing} addresses both issues by converting each input once into a structured text record of \emph{what} it contains and \emph{where} or \emph{when} each part occurs.
Any LLM, including a text-only one, can then read and act on this record and trace a claim back to its source through the anchors it cites; parsing thus provides a perception layer between non-text inputs and language-model reasoning.

Current parsers, however, cover this range of inputs only partially.
Document parsers have become strong on standard pages~\citep{Youtu-Parsing,MinerU2.5,PaddleOCR-VL-1.6}, but symbolic content such as chemical structures and sheet music is still left largely to domain-specific models~\citep{ChemDFM-X,LEGATO}.
Beyond documents, parsing is even less mature.
Omni-modal models are built mainly for captioning and question answering rather than for grounded, structured output~\citep{Qwen2.5-Omni,Qwen3-Omni}, and parsing across images, audio, and video has only recently been formulated as a task with a dedicated benchmark~\citep{Logics-Parsing-Omni}.
As a result, different input types are typically served by different models with different output formats.
Yet the target output is the same across input types, namely a structured record whose units are anchored in space or time, which suggests building a single parser with one output schema rather than one specialist per input type.

Building such a parser raises two challenges.
\textbf{(C1) Joint perception across modalities.}
Omni-modal models~\citep{Qwen3-Omni,VITA-1.5} and the omni-modal parser \Logics~\citep{Logics-Parsing-Omni} pair two separately pre-trained encoders, one for vision and one for audio, an arrangement with three consequences.
First, each modality is encoded in isolation: no cross-modal fusion takes place during perception, and the language model must infer cross-modal relations from separate representations.
Second, the two encoders index time independently, so a frame and the sound of the same instant do not share temporal positions within perception; temporal alignment, where modeled, is deferred to the language model~\citep{Qwen2.5-Omni}.
Third, each encoder requires its own architecture, training objective, and pre-training, so perception parameters are not shared across modalities and each new modality requires another tower.
The first two consequences weigh more on parsing than on captioning or question answering, because a structured parse must state explicitly which content co-occurs across modalities and when.
Encoders shared across vision and audio have been explored~\citep{VATT,TVLT,OmniEncoder}, but, to our knowledge, not for grounded parsing.
\textbf{(C2) Joint optimization across tasks.}
The same model must also be optimized jointly for content accuracy, spatial and temporal grounding, and schema validity across ten tasks whose output fields differ.
Supervised fine-tuning (SFT) trains only on reference prefixes, never on the prefixes the model itself produces at inference~\citep{DAgger,ScheduledSampling}.
Reinforcement learning with verifiable rewards (RLVR), including Group Relative Policy Optimization (GRPO)~\citep{DeepSeekMath}, is increasingly used in document parsing~\citep{olmOCR-2,MinerU2.5-Pro,PaddleOCR-VL-1.6}; it trains on the model's own outputs but requires a scalar reward that is hard to define across such heterogeneous criteria and provides only sequence-level feedback.
On-policy distillation supplies token-level targets on the model's own outputs using accessible teacher distributions~\citep{GKD}. Reference-conditioned self-distillation can instead obtain these targets from the same model under an enriched context~\citep{OPSD}; the remaining question is how to adapt this supervision to heterogeneous schema roles and refresh it as the parser learns.

To address both challenges, we present \OurMethod, \emph{an omni-modal parser built on one encoder for every modality and one schema for every task} (Fig.~\ref{fig:overview}).
For C1, \textbf{a unified perception encoder}, \OurEncoder, replaces the two towers with one shared Transformer initialized from a pre-trained text language model to strengthen text reading.
Only lightweight input stems and output projectors remain modality-specific, so almost all perception parameters are shared across modalities.
Visual and acoustic tokens share one space-time positional encoding and interact in a few fusion layers, so a frame and the sound of the same instant are aligned and fused within short audio-visual windows inside perception, leaving only dependencies across windows to the language model.
For C2, the shared schema allows \textbf{a single self-evolving post-training procedure}, OmniSchema-Aware On-Policy Distillation (\OSAD), to use the same token-role-based objective across tasks after SFT and schema-routed RLVR.
Instead of a sequence-level reward, the model learns from token-level targets on its own parses, provided by a frozen copy of itself that additionally conditions on the reference parse~\citep{OPSD,LUPI}.
Because every output follows \OmniSchema, each token can be identified as content or structure and supervised with an objective suited to its role.
After each round, the trained student becomes the next teacher, so the supervision evolves with the parser; \OSAD itself thus requires neither a hand-designed reward nor a separately trained teacher.

With this design, a single $5$\,B-parameter \OurMethod checkpoint serves all ten parsing tasks and achieves \emph{state-of-the-art document parsing}, scoring $96.96$ on \OmniDocBench v1.6 (Fig.~\ref{fig:teaser}a).
On \OmniParsingBench, it leads open-weight models with a score of $75.08$, surpassing \Logics ($72.26$), with its largest gains on audio and video (Fig.~\ref{fig:teaser}b).
The same checkpoint also excels at \emph{symbolic transcription} (Fig.~\ref{fig:teaser}c).
On chemical structures, it reaches an exact fingerprint-match rate (Tani@1.0) of $54.2$, outperforming DeepSeek-OCR~2~\cite{DeepSeek-OCR-2} ($49.6$).
On typeset sheet music, it achieves a character error rate (CER) of $22.73$, on par with the specialist LEGATO~\citep{LEGATO} ($23.30$, as reported in its original paper). Together, these results show that \emph{one encoder and one schema can support broad multimodal parsing while retaining the precision of specialized models}.

\begin{figure}[!t]
\centering
\includegraphics[width=\textwidth]{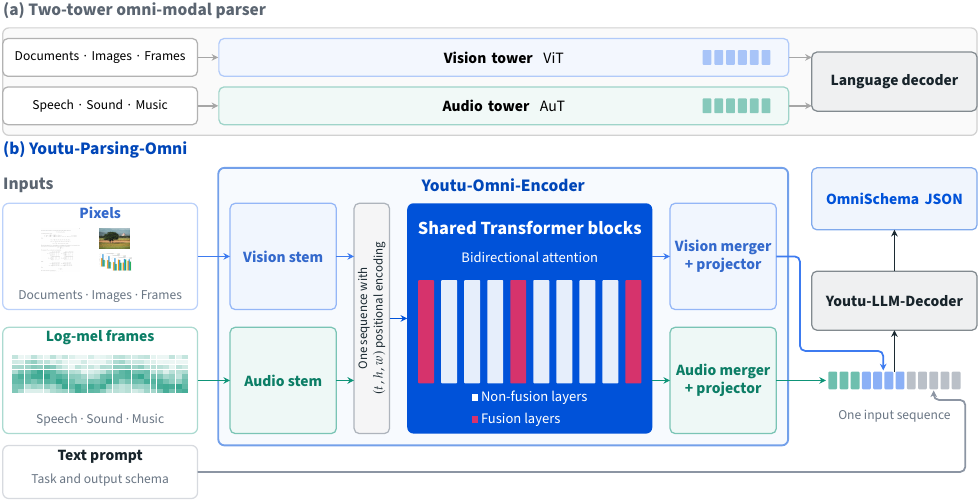}
\caption{
    \textbf{Overview of omni-modal parsing architectures.}
    We compare how visual and audio inputs are encoded before reaching the language model in the conventional design and in \OurMethod{}.
    (a) Existing omni-modal parsers encode pixels and audio with two separately pre-trained towers, whose streams first meet inside the language model.
    (b) \OurMethod{} replaces both towers with a single encoder, \OurEncoder. Thin modality-specific stems map pixels and log-mel frames to tokens; a shared bidirectional block then contextualizes an arbitrary packed mix of images, audio chunks, and videos under one $(t,h,w)$ positional encoding; and a few of its layers (red) attend across modalities within short audio-visual windows (Fig.~\ref{fig:encoder-mechanics}).
    As a result, pixels and sound share weights and positional encoding, and interact already inside the encoder.
}
\label{fig:arch-framework}
\end{figure}

\section{Model}
\label{sec:model}
\label{sec:arch}

\OurMethod is a decoder-only language model fed by one perception encoder (Fig.~\ref{fig:arch-framework}).
It accepts text, images, audio, and video (with its audio track when present) and emits text only.
The decoder is a dense text language model pre-trained following the recipe of Youtu-LLM-2B~\citep{Youtu-LLM}; it keeps 1D rotary positions, since spatial and temporal structure is already resolved inside the encoder (App.~\ref{subsec:app-arch-llm}).
Further architectural details, including the variants that the final model does not use, are given in App.~\ref{sec:app-arch}, and the complete configuration in Tab.~\ref{tab:arch-config-appx}.
Two principles shape the encoder.
First, every modality is read by the same weights with the same space-time positional encoding (Subsec.~\ref{subsec:omni-encoder}--\ref{subsec:omni-video}).
Second, these weights start from a model that already maps context to symbols (Subsec.~\ref{subsec:eval-ablation-encinit}).
The first principle thus amortizes perception capacity across modalities rather than duplicating it in one tower per modality.

\subsection{\OurEncoder}
\label{subsec:omni-encoder}
\label{subsec:omni-overview}

Vision and audio inputs pass through a three-part pipeline whose middle part is shared: a modality-specific stem turns raw input into $d$-dimensional tokens, a shared Transformer contextualizes them, and a modality-specific merger and projector down-sample the result and map it to the decoder width.
Only the stems and the merger--projector heads depend on the modality; together they hold a small fraction of the encoder's parameters (App.~\ref{subsec:app-omni-stems}), so most of the encoder is shared.
Architecturally, adding a modality therefore requires a new stem and projector rather than a second tower with its own pre-training lineage, although the shared weights must still be trained on the new input.

\paragraph{Stems.}
\label{subsec:omni-stems}
The two stems convert raw signals into tokens with minimal machinery.
Images and video frames are cut into fixed-size patches at their native resolution under a token budget~\citep{Qwen2-VL}, and a single linear map turns each patch into one token; no further convolution, absolute position table, or class token is used.
Audio is converted to a log-mel spectrogram, and a few strided convolutions down-sample it in time before a linear map to width $d$, giving a fixed audio token rate of $r_a$ tokens per second.

\begin{figure}[!t]
\centering
\includegraphics[width=\textwidth]{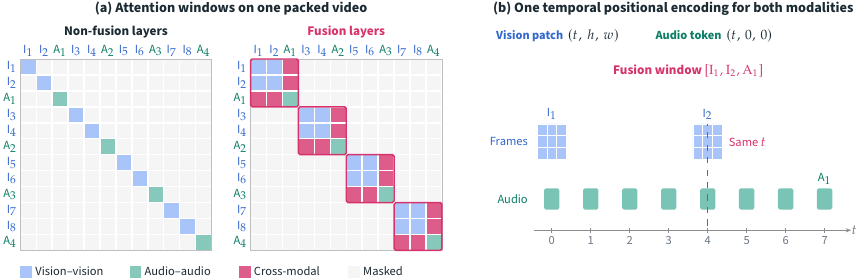}
\caption{
    \textbf{Concurrent visual and audio content share fusion windows and a common temporal coordinate system.}
    (a) A video with eight frames (I) and four audio chunks (A), packed in temporal order, with each chunk placed after the frames it overlaps.
    A non-fusion layer gives every frame and every chunk its own window;
    a fusion layer starts a new window at every audio-to-frame boundary, so the video forms four $(\mathrm{I}^{+}\mathrm{A}^{*})$ windows; red cells mark the cross-modal pairs, and off-diagonal blue cells the frame-to-frame pairs, that only fusion layers admit.
    (b) Vision tokens take $t$ from the frame index and $(h,w)$ from the patch grid; audio tokens take $t$ from their offset and write into the same temporal rotary slots.
    Inside a fusion window both advance on the common time scale of Equ.~\ref{equ:time-align}.
    }
\label{fig:encoder-mechanics}
\label{fig:fusion-layers}
\label{fig:rope-axes}
\end{figure}

\paragraph{Shared Transformer blocks.}
\label{subsec:omni-block}
The shared Transformer is a stack of standard pre-norm blocks with RMSNorm~\citep{RMSNorm}, SwiGLU~\citep{SwiGLU}, grouped-query attention~\citep{grouped-query-attention}, and QK-Norm~\citep{QKNorm}.
Attention is bidirectional.
Rotary position embeddings (RoPE)~\citep{RoFormer} are applied from a matrix of per-token rotation angles rather than read from one position table shared by the whole sequence; this is what makes the axis-factorized positional encoding of Subsec.~\ref{subsec:omni-video} expressible.
Any mix of images, audio chunks, and videos is packed into one variable-length sequence whose attention windows are given by a cumulative-length vector $\mathbf{c}$~\citep{FlashAttention}, so one encoder pass handles the whole mix without padding.

\paragraph{Token merger and projector.}
\label{subsec:omni-merge}
After the shared Transformer, neighboring vision tokens within a frame are concatenated along the feature axis into one token, and adjacent audio tokens are merged in the same way along time.
Each modality then passes through an RMSNorm and a two-layer MLP that projects it to the decoder width.

\subsection{Unified Positional Encoding and Audio-Visual Fusion}
\label{subsec:omni-video}

\paragraph{One positional encoding for all modalities.}
\label{subsec:omni-rope}

Sharing one Transformer across modalities makes positional encoding the central design question: one rotary budget of frequency slots has to express 2D layout, 1D time, and the correspondence between the two.
We partition the budget into contiguous segments of near-equal size, one for each axis $T$, $H$, and $W$, each with its own inverse-frequency table normalized by its own width, so every axis spans the full frequency range regardless of how many slots it receives (Equ.~\ref{equ:chunked}, App.~\ref{subsec:app-omni-rope}).
A vision token takes $t$ from its frame index and $(h,w)$ from its patch position; an audio token takes $t$ from its time offset and sets $h=w=0$.
Both modalities therefore write into the same $T$ segment, expressing their temporal positions on a common time scale.
Modality identity is left to the features: the two stems produce readily separable feature distributions.

The shared timeline makes frame and audio positions comparable, but it does not by itself let them interact.
Fusion provides this interaction by letting the two attend to each other, so that the correspondence a video parse must output (for instance, which words are spoken while a given slide is on screen) is computed in perception rather than reconstructed by the decoder.

\paragraph{Fused and independent routes.}
A video reaches the encoder by either of two routes that share all weights and return outputs of identical shape.
On the \emph{independent} route, each frame and each audio chunk forms its own attention window, as if the input were a burst of images and an unrelated recording.
On the \emph{fused} route, the frames and audio chunks of one video are packed in temporal order, and in the fusion layers audio-visual windows let a frame attend to the sound of the same instant; the other layers keep one window per frame or chunk.
During training, a video takes the fused route with probability $p_{\mathrm{fus}}$, set below $1$ from Stage~2 on (Sec.~\ref{sec:training}), which keeps both routes usable at inference.

\paragraph{Windows and time alignment.}
On the fused route, the packed event list (e.g., $[\mathrm{I}_1,\mathrm{I}_2,\mathrm{A}_1,\mathrm{I}_3,\dots]$) is cut into $(\mathrm{I}^{+}\mathrm{A}^{*})$ groups, i.e., one or more frames followed by the audio chunks that precede the next frame, with a new group starting at every audio-to-frame boundary; in the fusion layers, each group becomes one attention window (Fig.~\ref{fig:encoder-mechanics}a).
Inside a group, vision and audio advance on a common time scale tied to the audio token rate,
\begin{equation}
\label{equ:time-align}
\Delta t_{\text{audio}}=1\ \text{per token},\qquad
\Delta t_{\text{vision}}=\frac{r_a}{\mathrm{fps}}\ \text{per frame},
\end{equation}
where $\mathrm{fps}$ is the sampled video frame rate and $r_a$ is the audio token rate of the audio stem, before merging. This places frames and audio tokens on a common time scale, without requiring every sampled frame to coincide with a discrete audio-token position (Fig.~\ref{fig:encoder-mechanics}b).

\paragraph{Sparse fusion.}
Because windows are specified only by $\mathbf{c}$, fusion is a per-layer choice: the same packed features carry a per-group segmentation $\mathbf{c}^{\text{fus}}$ and a per-event (non-fusion) segmentation $\mathbf{c}^{\text{nof}}$, and each layer uses one of them.
The two segmentations assign different time origins to the same token, which is harmless because a rotary attention score depends only on the offset $t_q-t_k$ between query and key positions, and queries never meet keys from another window.

\section{Omni-Modal Parsing Schema}
\label{sec:task}
\label{sec:data}

\begin{listing}[!t]
\centering
\begin{lstlisting}[basicstyle=\ttfamily\scriptsize,columns=fixed,keepspaces=true,frame=single,framesep=6pt,rulecolor=\color{black!25}]
{
  "modality": "image | audio | video",
  "subtype": "document | natural_image                      // image, video
              | natural_video | text_rich_video",
  "elements": [                                             // image inputs
    {
      "id": "e1",
      "category": "text | title | header | footer | caption | code | formula
                   | table | chart | flowchart | figure | geometric | chemical
                   | seal | score | entity",
      "bbox": "<box><x_1><y_1><x_2><y_2></box>",
      "tag": "...",                                         // natural image
      "content": {
        "text": "...",                                      // text, SMILES, ABC, in-box OCR
        "latex": "...",                                     // formula
        "otsl": "...",                                      // table
        "markdown": "...",                                  // chart
        "mermaid": "...",                                   // flowchart
        "geometric": "...",                                 // geometric figure
        "caption": "..."                                    // visual element, image region
      },
      "attributes": {"language": "zh | en | ..."}           // documents
    }
  ],
  "segments": [                                             // audio, video inputs
    {
      "id": "s_001",
      "kind": "vocal | nonvocal",                           // audio
      "start_timestamp": "HH:MM:SS",
      "end_timestamp": "HH:MM:SS",
      "visual": {                                           // video
        "elements": [{"category": "object | text", "ocr_type": "title | ... | null",
                      "content": {"text": "..."}}],
        "atomic_actions": ["..."],
        "object_actor_interactions": ["..."],
        "camera_motion": {"l1": "...", "l2": "..."},
        "event_description": "..."
      },
      "audio": {
        "speaker_id": "spk1",                               // vocal audio, video
        "asr_text": "...",                                  // vocal audio, video
        "timbre_caption": "...",                            // audio, vocal
        "instruct_caption": "...",                          // audio, vocal
        "scene_caption": "...",                             // audio, vocal
        "env_caption": "...",                               // audio, nonvocal
        "mode": "asr",                                      // video
        "acoustic_events": ["..."]
      },
      "language": "zh | en | unk"                           // audio, vocal
    }
  ],
  "reading_order": ["e1"],                                  // documents
  "global_description": {
    "caption": "...", "narrative": "...", "scene": "...",
    "language": "...", "tag": "..."
  },
  "structured_report": "..."                               // text-rich video; empty for natural video
}
\end{lstlisting}
\caption{\textbf{Overview of the \OmniSchema output.} The listing is schematic: \texttt{|} separates alternative values, \texttt{"..."} stands for elided content, and comments (\texttt{//}) are annotations rather than part of the output. Each task uses the subset of fields it needs; comments mark the inputs that use a field, and an unmarked field applies to every input that uses its enclosing field. Every output has \texttt{modality}, a list of \texttt{elements} or \texttt{segments}, and \texttt{global\_description}.}
\label{lst:omnischema-json}
\end{listing}

As discussed in Sec.~\ref{sec:intro}, parsing results of different modalities share a common form: a set of units, each localized in space or time and paired with the content extracted from the corresponding region or interval.
Motivated by this observation, we design a unified output schema, \OmniSchema, which covers ten parsing tasks across three modalities: images (document pages and natural images), audio, and video.
Because every element or segment is localized by an explicit spatial or temporal anchor, each parsed result is traceable to its source region or time interval.

\paragraph{Schema structure.}
Lst.~\ref{lst:omnischema-json} gives an overview of the schema.
Each output is serialized as a JSON object with three common required fields: \texttt{modality}, a list of localized units with their content (\texttt{elements} for images, \texttt{segments} for audio and video), and a global description of the input (\texttt{global\_description}). Additional fields depend on the task: \texttt{subtype} and \texttt{reading\_order} are included where applicable, and video outputs contain \texttt{structured\_report}, populated for text-rich video and left empty for natural video.
The unit type depends on the modality.
For images, the unit is an \emph{element}, a spatial region delimited by a bounding box \texttt{<box><x\_1><y\_1><x\_2><y\_2></box>} whose coordinates are normalized to $0$--$1{,}000$ relative to the image width and height; for audio and video, it is a \emph{segment}, a time interval delimited by start and end timestamps (\texttt{HH:MM:SS}, i.e., at one-second resolution, relative to the input onset), and each speech segment additionally includes one speaker identifier that is consistent across the input.
The content fields of each unit contain the information extracted from the corresponding region or interval.
For example, \texttt{text} contains the recognized text of a text element, \texttt{otsl} encodes the structure and cell content of a table, and \texttt{asr\_text} contains the speech transcript of an audio or video segment.

\paragraph{Multi-task parsing.}
\OurMethod supports ten parsing tasks: on images, document parsing, chemical structure recognition, sheet music recognition, chart parsing, flowchart parsing, geometric figure parsing, and natural image parsing; on audio, audio parsing; and on video, natural video parsing and text-rich video parsing.
In the image tasks, each parsed region is an element delimited by a bounding box.
In \emph{document parsing}, each element on the page is labeled with a layout category, placed in the reading order, and transcribed according to its category, e.g., as plain text, a \LaTeX{} formula, or a table serialized in Optimized Table Structure Language (OTSL).
In \emph{chemical structure recognition}, a molecule is parsed as a document element whose content is a SMILES~\citep{SMILES} string.
In \emph{sheet music recognition}, a musical score is parsed as a document element whose content is ABC~\citep{ABC} notation.
In \emph{chart parsing}, \emph{flowchart parsing}, and \emph{geometric figure parsing}, the whole figure forms a single element whose content is its underlying data as a Markdown table, its nodes and edges as a Mermaid graph, or its geometric primitives and their relations as a geometry string, respectively; the element also contains a caption.
In \emph{natural image parsing}, each salient object forms an element with a bounding box, a tag naming the object, the text inside the box, and a region caption, and the global description adds an image caption, a narrative, and a scene description.
In the audio and video tasks, the input is divided into a sequence of time segments.
Each audio segment is labeled as vocal or nonvocal: a vocal segment contains a speaker identifier, a transcript, the language of the speech, acoustic-event labels, and captions of timbre, delivery, and scene, whereas a nonvocal segment contains acoustic-event labels and a caption of the acoustic environment.
Each video segment contains the objects and on-screen text visible in its frames, where each text item is labeled with an optical character recognition (OCR) type such as title.
It also contains atomic actions, actor-object interactions, camera motion at a coarse and a fine level, an event description, and the speaker identifier, transcript, and acoustic-event labels of the accompanying audio.
\emph{Natural video parsing} and \emph{text-rich video parsing} share this segment structure; the output of text-rich video parsing additionally contains a structured report in Markdown that summarizes the whole video, a field left empty for natural video.
Full example outputs for each input type are given in App.~\ref{sec:app-examples}.

\paragraph{Token roles.}
Because every output follows \OmniSchema, each token of a serialized output can be assigned one of two roles, which \OSAD supervises with different objectives (Subsec.~\ref{subsec:osad}).
Structure tokens belong to the JSON syntax, namely keys and punctuation such as braces, brackets, colons, and commas, and follow a constrained grammar; content tokens belong to field values, which can admit multiple valid continuations.

\section{Training}
\label{sec:training}

\begin{figure}[!t]
\centering
\includegraphics[width=\textwidth]{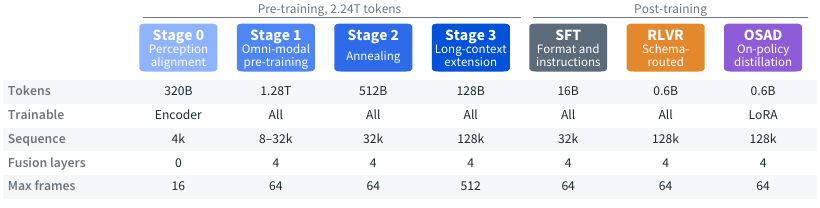}
\caption{\textbf{Perception first, output format second, on-policy refinement last.} Four pre-training stages (Stage~0: perception alignment; Stage~1: omni-modal pre-training; Stage~2: annealing; Stage~3: long-context extension) are followed by three post-training stages: supervised fine-tuning, schema-routed RLVR, and \OSAD. Rows give each stage's token budget, trainable parameters, sequence-length limit, number of encoder fusion layers, and maximum frames per training video. Stage~1 uses sequence lengths of $8$\,k--$32$\,k. LoRA denotes low-rank adaptation of frozen base weights.}
\label{fig:recipe}
\label{fig:pt-stages}
\end{figure}

Training first develops unified perception through four pre-training stages and then refines structured parsing through three post-training stages (Fig.~\ref{fig:recipe}).
Supervised fine-tuning establishes the output format and instruction following; two subsequent on-policy stages refine the model's own parses using verifiable rewards and then reference-conditioned token-level supervision.

\subsection{Pre-Training}
\label{subsec:pretrain}
\label{subsec:pt-curriculum}

Pre-training turns a text-only decoder and a text-initialized encoder into an omni-modal perceiver, which raises three coupled problems.
First, the randomly initialized stems and projectors must learn to map pixels and mel frames into a representation that the pre-trained Transformer weights can read.
Second, the model must acquire dense, native-resolution perception of images, audio, and video without eroding the decoder's language ability.
Third, it must handle long inputs such as multi-page documents and long videos.
Addressing all three at once is both costly and unstable: gradients through randomly initialized projectors would perturb the pre-trained decoder, and long native-resolution sequences are expensive to train on before the features are aligned.
We therefore adopt a staged curriculum, and progressively expand four axes across Stages~0--3 (Fig.~\ref{fig:recipe}):
the trainable parameters (encoder only, then all parameters);
the input scale (from $4$\,k-token sequences with at most $256$ tokens per image to $128$\,k-token sequences with up to $512$ video frames);
cross-modal interaction (from no fusion layers to four);
and input perturbation (from none to stochastic routing and augmentation).

We initialize the decoder from our text-only checkpoint and the shared encoder block from Qwen3-0.6B-Base~\citep{Qwen3} without its embedding layer and LM head; stems and projectors are randomly initialized.

\paragraph{Stage~0: Perception Alignment.}
The first stage bridges the randomly initialized stems and projectors to the pre-trained weights on either side of them.
Only the encoder (stems, shared block, and projectors) is trained, while the decoder stays frozen.
Freezing the decoder serves two purposes: it shields the language model from the noisy gradients of the new projectors, and it forces the encoder to produce features in the input space that the fixed language model already interprets.
The supervision is dense: captions and transcripts describe the whole input, so every region of an image and every second of a recording contributes to the loss, whereas a short answer about one region, as in visual question answering, would provide little supervision for the remaining content.
Inputs are kept small (at most $256$ tokens per image, $16$ frames per video, and $4$\,k tokens per sequence) and no fusion layer is enabled, so that alignment is learned cheaply over a large token budget before native-resolution inputs make each step expensive.

\paragraph{Stage~1: Omni-Modal Pre-Training.}
Once the encoder emits features that the decoder can read, the bottleneck shifts from alignment to capacity: a frozen decoder cannot learn to emit the long, dense symbol sequences required by parsing.
Stage~1 therefore unfreezes all parameters and is the largest stage, consuming $1.28$\,T tokens.
The image budget grows from $256$ to $20$\,k tokens so that the model sees pages at native resolution, videos grow to $64$ frames of up to $512$ tokens each, and sequence lengths range from $8$\,k to $32$\,k tokens.
Four fusion layers are enabled at this point, so that frames attend to the co-occurring audio (Subsec.~\ref{subsec:omni-video}) only after each modality has been aligned on its own.
Text-only data are mixed in to limit catastrophic forgetting of the decoder's language ability.
The learning rate is held constant so the total budget need not be fixed in advance and any checkpoint can be handed to the decay stage.

\paragraph{Stage~2: Annealing.}
Under a constant learning rate, the loss plateaus above the level that a subsequent learning-rate decay reaches, and annealing is a standard final phase of large-scale pre-training.
Stage~2 decays the learning rate from $10^{-4}$ to $10^{-5}$ with a cosine schedule over $512$\,B tokens, with sequences fixed at $32$\,k tokens and slightly larger image and per-frame budgets.
Because annealing consolidates what the model has learned, we also introduce two mechanisms here that protect or broaden the learned perception.
First, layer-wise learning-rate decay scales the learning rate of encoder depth $\ell$ by $\gamma^{L+1-\ell}$ with $\gamma=0.9$, from the stems ($\ell=0$) to the projectors ($\ell=L+1$, no decay); the projectors, the newest components, keep adapting, whereas the stems and lower layers, which already encode generic low-level features, are protected from being overwritten.
Second, audio-visual videos are routed stochastically with fused-route probability $p_{\mathrm{fus}}=0.8$, so they take the independent route with probability $0.2$ and both routes remain usable at inference.

\paragraph{Stage~3: Long-Context Extension.}
Multi-page documents and long videos require far longer sequences than Stages~0--2 provide.
Stage~3 trains on sequences of up to $128$\,k tokens, with up to $102{,}400$ tokens per image or video and videos of up to $512$ frames, for $128$\,B tokens at a constant learning rate of $10^{-5}$.

\subsection{Post-Training}
\label{subsec:posttrain}

Pre-training yields a model that perceives all three modalities but does not yet emit \OmniSchema reliably.
Post-training proceeds through supervised fine-tuning (SFT) for \OmniSchema serialization, schema-routed reinforcement learning with verifiable rewards (RLVR) for sequence-level optimization, and OmniSchema-Aware On-Policy Distillation (\OSAD) for token-adaptive supervision and iterative self-evolution.

\subsubsection{Supervised Fine-Tuning}
\label{subsec:sft}
Starting from the Stage~3 checkpoint, we train all parameters on prompt--response pairs to establish reliable \OmniSchema serialization and instruction following.
Although the pre-training mixture already contains parsing data, the Stage~3 model does not yet consistently respond to a task prompt with a complete, schema-conformant object; SFT makes this output format the default response to each parsing instruction.

The SFT mixture combines two sources.
Parsing data cover all ten tasks in Sec.~\ref{sec:task}: each sample pairs a task prompt and its image, audio, or video input with a reference parse serialized in \OmniSchema, so a single model learns to emit the fields each task requires.
General-purpose data, reused from Youtu-LLM~\citep{Youtu-LLM}, Youtu-VL~\citep{Youtu-VL}, Long-VITA~\citep{Long-VITA}, VITA-1.5~\citep{VITA-1.5}, and VITA-Audio~\citep{VITA-Audio}, span text-only, visual, and audio instructions and preserve the conversational and instruction-following abilities that a parsing-only mixture would erode.
SFT uses a budget of $16$\,B tokens, sequences of up to $32$\,k tokens, and at most $64$ frames per video, and keeps the four fusion layers of Stages~1--3 (Fig.~\ref{fig:recipe}).

SFT also prepares the model for the two on-policy stages that follow, both of which learn from the model's own rollouts.
RLVR assigns zero reward to undecodable outputs, and with group-mean-centered advantages, a group in which every rollout fails yields no gradient; \OSAD likewise distills only from rollouts that pass its admission checks.
A policy that rarely produces well-formed parses would therefore supply few useful learning signals to either stage.
SFT, however, supervises the model only under reference prefixes, a limitation that the on-policy stages address (Subsec.~\ref{subsec:osad}).

\subsubsection{Schema-Routed RLVR}
\label{subsec:srvr}
For an input query $x$, let $y$ denote the generated parse, $y^{\star}$ its reference parse, and $\nu$ the task index.
Here, $x$ includes the task prompt and its multimodal input.
We optimize the policy on its own generated parses with Group Relative Policy Optimization (GRPO)~\citep{DeepSeekMath}. The reward is gated by decodability and the presence of a primary payload, and combines shared well-formedness with task-specific content accuracy:
\begin{gather}
\label{equ:srvr-reward}
R(y,y^{\star},\nu)=\indic[y\text{ decodes}]\,\indic[y\text{ has a payload}]\cdot\mathrm{clip}_{[0,1]}\big(\Omega(y,y^{\star})\cdot\mathrm{core}_{\nu}(y,y^{\star})\big),\\
\Omega(y,y^{\star})=\mathrm{Fmt}(y)\,\big(1-0.6\,P_{\mathrm{repeat}}\big)\,\big(1-0.7\,P_{\mathrm{under}}\big).\nonumber
\end{gather}
The indicator $\indic[\cdot]$ equals one when its condition holds and zero otherwise; $\mathrm{clip}_{[0,1]}$ restricts its argument to $[0,1]$.
If either gate fails, the reward is zero; otherwise, it equals the clipped product of the well-formedness and content factors. Thus the final reward lies in $[0,1]$.
The shared factor $\Omega$ combines schema validity, scored by $\mathrm{Fmt}(y)$, with penalties for repetition ($P_{\mathrm{repeat}}$) and incomplete coverage ($P_{\mathrm{under}}$).
Both penalties compare $y$ with $y^{\star}$; their arguments are suppressed for brevity, and $0.6$ and $0.7$ are their respective coefficients.
The task-specific factor $\mathrm{core}_{\nu}$ averages grounded-field accuracy and reference-description similarity when reference descriptions are available; otherwise, it uses grounded-field accuracy alone.
App.~\ref{subsec:app-srvr} details the validity checks, penalties, and task-specific scorers.

\subsubsection{OmniSchema-Aware On-Policy Distillation}
\label{subsec:osad}

\begin{figure}[!t]
\centering
\includegraphics[width=\textwidth]{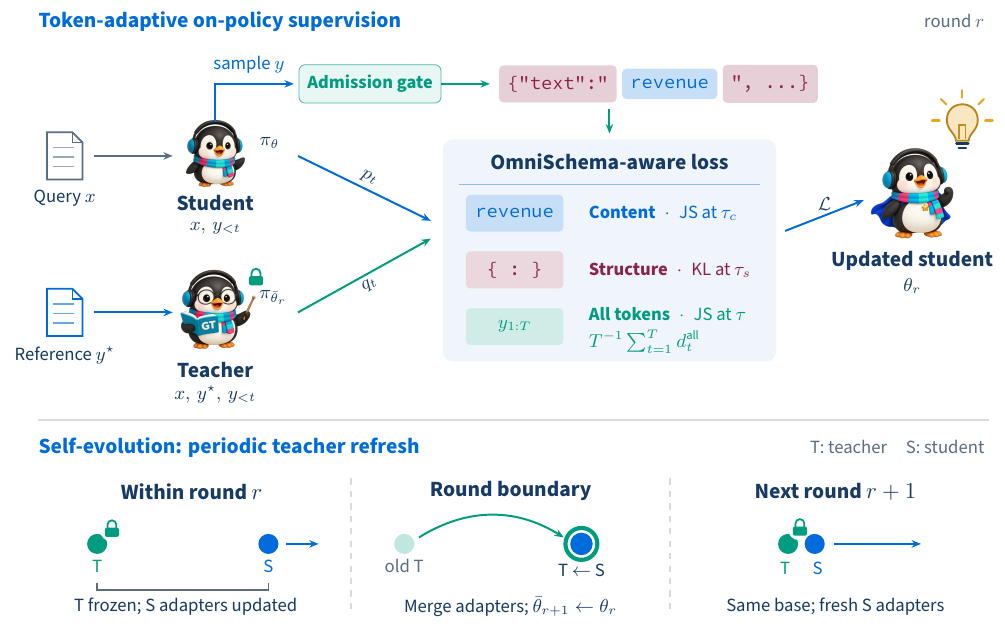}
\caption{\textbf{\OSAD combines token-adaptive on-policy supervision with teacher refresh at each round boundary.} Admitted student rollouts provide shared prefixes for a LoRA-trained student and a frozen, reference-conditioned teacher. The objective combines content-specific JS divergence at $\tau_c$, structure-specific forward KL divergence at $\tau_s$, and an all-token JS term at $\tau$. Within each round, only the student's adapters are updated. At the round boundary, the learned adapters are merged into the base weights to initialize the next teacher and student; the student receives fresh adapters, while only the teacher conditions on the reference. The lower panel illustrates the update and refresh mechanism rather than a measured model-state distance.}
\label{fig:osad}
\end{figure}

Starting from the RLVR checkpoint, \OSAD refines generated parses through token-adaptive supervision and self-evolution through teacher refresh (Fig.~\ref{fig:osad}).
The first matches supervision to each token's schema role; the second updates the teacher as the student learns.

\paragraph{Token-adaptive on-policy supervision.}
SFT trains on reference prefixes, whereas inference uses model-generated prefixes~\citep{DAgger,ScheduledSampling}.
RLVR trains on these generated prefixes but assigns one reward to the entire parse.
\OSAD complements this sequence-level optimization with reference-conditioned targets at each token, building on on-policy distillation~\citep{GKD,MiniLLM} and reference-conditioned self-distillation~\citep{OPSD}, with the reference serving as privileged information~\citep{LUPI}.

\OSAD proceeds in rounds. Within round $r$, a frozen teacher with parameters $\bar\theta_r$ and a student with effective parameters $\theta$ start from the same checkpoint.
The student trains low-rank adapters (LoRA)~\citep{LoRA} on frozen base weights and samples a response $y=(y_1,\ldots,y_T)$ of length $T$ for each query $x$.
Only rollouts passing JSON-validity, schema-conformance, box-syntax, and repetition checks enter distillation; rejected rollouts are masked (App.~\ref{subsec:app-osad}).
Both models score the same generated prefix $y_{<t}=(y_1,\ldots,y_{t-1})$, but only the teacher receives the reference parse through $\tilde x=[y^{\star};x]$:
\begin{equation}
p_t^{(\tau)}=\mathrm{softmax}\big(f_{\theta}(x,y_{<t})/\tau\big),\qquad
q_t^{(\tau)}=\mathrm{softmax}\big(f_{\bar\theta_r}(\tilde x,y_{<t})/\tau\big).
\label{equ:osad-pq}
\end{equation}
Here, $p_t^{(\tau)}$ and $q_t^{(\tau)}$ are student and teacher next-token distributions, $f$ returns vocabulary logits, and $\tau>0$ denotes distillation temperature.
Shared prefixes thus provide aligned targets on the student's own trajectories without exposing the reference to the student.

Following the token roles in Sec.~\ref{sec:data}, let $\mathcal{C}(y)$ and $\mathcal{S}(y)$ denote the content and structure positions of a response, respectively.
Content fields may admit several valid continuations, whereas JSON keys and punctuation follow a constrained grammar.
We therefore use the symmetric, bounded Jensen--Shannon (JS) divergence for content and the forward Kullback--Leibler (KL) divergence for structure:
\begin{equation}
d_t^{\mathcal{C}}=\mathrm{JS}\big(q_t^{(\tau_c)}\,\big\|\,p_t^{(\tau_c)}\big),\qquad
d_t^{\mathcal{S}}=\mathrm{KL}\big(q_t^{(\tau_s)}\,\big\|\,p_t^{(\tau_s)}\big),\qquad \tau_s>\tau_c.
\label{equ:osad-div}
\end{equation}
The per-position divergences $d_t^{\mathcal{C}}$ and $d_t^{\mathcal{S}}$ use content and structure temperatures $\tau_c$ and $\tau_s$, respectively, and are computed over the full vocabulary. Round indices on temperatures and role weights are omitted here for brevity.
JS bounds content disagreement, while forward KL encourages coverage of the teacher's structural predictions, softened by the higher structure temperature.

The objective combines a shared all-token term with additional role-specific terms.
For each nonempty role $\mathcal{R}\in\{\mathcal{C},\mathcal{S}\}$, define its mean divergence as $\bar d^{\,\mathcal{R}}(y)=|\mathcal{R}(y)|^{-1}\sum_{t\in\mathcal{R}(y)}d_t^{\mathcal{R}}$.
The all-token term applies JS uniformly at temperature $\tau$ and averages over all $T$ response positions:
\[
d_t^{\mathrm{all}}=\mathrm{JS}\big(q_t^{(\tau)}\,\big\|\,p_t^{(\tau)}\big),\qquad
\bar d^{\,\mathrm{all}}(y)=\frac{1}{T}\sum_{t=1}^{T}d_t^{\mathrm{all}}.
\]
With nonnegative content and structure weights $\alpha_c$ and $\alpha_s$, the per-rollout loss and minibatch objective are
\begin{equation}
\mathcal{L}_{\mathrm{row}}(y)=\frac{\bar d^{\,\mathrm{all}}(y)+\alpha_c\,\bar d^{\,\mathcal{C}}(y)+\alpha_s\,\bar d^{\,\mathcal{S}}(y)}{1+\alpha_c+\alpha_s},
\qquad
\mathcal{L}=\frac{1}{|\mathcal{B}^{+}|}\sum_{y\in\mathcal{B}^{+}}\mathcal{L}_{\mathrm{row}}(y).
\label{equ:osad-loss}
\end{equation}
Here, $\mathcal{B}^{+}$ contains all admitted rollout instances in the global minibatch across data-parallel ranks; updates are skipped when this global collection is empty.
For an absent role, its weighted loss and coefficient are omitted from the numerator and denominator, respectively.
The all-token mean weights response positions equally, while role averaging controls the additional supervision independently of role length. Averaging over rollouts gives each admitted response equal weight. The role-weight schedule is given in App.~\ref{subsec:app-osad}.

\paragraph{Self-evolution through teacher refresh.}
Token-adaptive supervision remains tied to the starting checkpoint if the teacher stays fixed throughout training.
At each round boundary, \OSAD merges the learned adapters into the student's base weights to obtain $\theta_r$, which initializes the next teacher and student:
\begin{equation}
\bar\theta_{r+1}\leftarrow\operatorname{stopgrad}(\theta_r),
\qquad
\theta_{\mathrm{init}}^{(r+1)}\leftarrow\bar\theta_{r+1}.
\label{equ:osad-refresh}
\end{equation}
Here, $\operatorname{stopgrad}$ blocks gradients, and $\theta_{\mathrm{init}}^{(r+1)}$ denotes the next student's initial base weights, to which fresh low-rank adapters are added.
Teacher parameters are fixed within each round and refreshed between rounds, while reference conditioning remains exclusive to the teacher.
This alternation updates both the parser and its supervision on a fixed source pool.
We run six rounds of \OSAD. App.~\ref{sec:app-posttrain} provides the training settings and pseudocode (Alg.~\ref{alg:osad}); Subsec.~\ref{subsec:eval-posttrain} evaluates the successive checkpoints.

\begin{table}[!t]
    \centering
    \caption{\textbf{Evaluation protocol.}
    Size is the number of evaluated samples. Every benchmark is scored with its official code (the LEGATO protocol for PDMX-Synth), except OmniParsingBench, which we score with our own toolkit (App.~\ref{sec:app-eval}). \emph{Baselines}: \emph{re-scored} means that every system is run and scored by us with the same code; \emph{leaderboard} means that baseline numbers are taken from the official public leaderboard. $^*$ChemDFM-X scores on ChemOCR are taken from TinyChemVL~\citep{TinyChemVL}; LEGATO scores on PDMX-Synth are taken from its original report~\citep{LEGATO} (Tab.~\ref{tab:symbolic}).}    \label{tab:eval-protocol}
    \small
    \setlength{\tabcolsep}{5pt}
    \renewcommand{\arraystretch}{1.12}
    \begin{tabularx}{\textwidth}{l >{\raggedright\arraybackslash\hsize=1.1\hsize}X r >{\raggedright\arraybackslash\hsize=0.9\hsize}X l}
    \toprule
     Benchmark & Split / task & Size & Reported metric & Baselines \\
    \midrule
    OmniDocBench v1.6~\hfilll~\cite{OmniDocBench} & full set & 1{,}651 & Overall, Edit, CDM, TEDS & leaderboard \\
    PureDocBench~\hfilll~\cite{PureDocBench} & clean, degraded, recaptured & 1{,}475\,$\times$\,3 & per-track Overall, Avg$_3$ & leaderboard \\
    \midrule
    \multirow{5}{*}{OmniParsingBench~\hfilll~\cite{Logics-Parsing-Omni}}
    & natural image           & 1{,}000 & \multirow{5}{=}{Overall: parse rate $\times$ mean of Perception and Cognition} & \multirow{5}{*}{re-scored} \\
      & chart, geometry         & 1{,}000 & & \\
      & audio                   & 938     & & \\
      & natural video (general) & 550     & & \\
      & text-rich video         & 250     & & \\
    \midrule
    Video-MME-v2~\hfilll~\cite{Video-MME-v2} & without subtitles & 3{,}200 & group score, L1--L3 & leaderboard \\
    \midrule
    ChemOCR~\hfilll~\cite{ChemVLM} & molecule $\rightarrow$ SMILES & 990 & Avg.\ Sim., Tani@1.0 & re-scored$^*$ \\
    PDMX-Synth~\hfilll~\cite{PDMX,LEGATO} & score $\rightarrow$ ABC & 800 & CER, SER, LER & re-scored$^*$ \\
    \bottomrule
    \end{tabularx}
\end{table}

\begin{table}[tp]
\centering
\caption{\textbf{Performance comparison of \OurMethod with other document parsing models and proprietary models on OmniDocBench v1.6.} \textbf{Bold} and \underline{underline}: best and second best in each column. $^\dagger$Proprietary model.
}
\label{tab:omnidocbench}
\small\setlength{\tabcolsep}{4.5pt}
\resizebox{\textwidth}{!}{%
\begin{tabular}{l r cccccc}
\toprule
 Model & Params & Overall$\uparrow$ & Text Edit$\downarrow$ & Formula CDM$\uparrow$ & TEDS$\uparrow$ & TEDS-S$\uparrow$ & Read.\ Order Edit$\downarrow$ \\
\midrule
\multicolumn{8}{@{}l}{\emph{Multi-stage specialists}} \\
TeleOCR~\hfilll~\cite{TeleOCR} & $1.2$\,B & \second{96.91} & \second{0.0267} & 96.59 & \best{96.82} & \second{98.18} & 0.1184 \\
PaddleOCR-VL-1.6~\hfilll~\cite{PaddleOCR-VL-1.6} & $0.9$\,B & 96.34 & 0.0326 & \best{97.53} & 94.76 & 97.10 & 0.1278 \\
MinerU2.5-Pro~\hfilll~\cite{MinerU2.5-Pro} & $1.2$\,B & 95.75 & 0.0360 & 97.45 & 93.42 & 95.92 & 0.1200 \\
GLM-OCR~\hfilll~\cite{GLM-OCR} & $0.9$\,B & 95.22 & 0.0440 & 97.18 & 92.83 & 95.39 & 0.1330 \\
PaddleOCR-VL-1.5~\hfilll~\cite{PaddleOCR-VL-1.5} & $0.9$\,B & 94.93 & 0.0380 & 96.89 & 91.67 & 94.37 & 0.1300 \\
\OurMethodV1~\hfilll~\cite{Youtu-Parsing} & $2.5$\,B & 93.74 & 0.0440 & 93.63 & 92.02 & 95.00 & 0.1160 \\
Dolphin-v2~\hfilll~\cite{Dolphin} & $3$\,B & 89.50 & 0.0690 & 91.01 & 84.40 & 87.44 & 0.1500 \\
MonkeyOCR-pro-3B~\hfilll~\cite{MonkeyOCR} & $3$\,B & 88.57 & 0.0740 & 88.74 & 84.35 & 88.62 & 0.1890 \\
\midrule
\multicolumn{8}{@{}l}{\emph{End-to-end specialists}} \\
OvisOCR2~\hfilll~\cite{OvisOCR2} & $0.8$\,B & 96.47 & \best{0.0265} & \second{97.49} & 94.58 & 96.98 & \second{0.1120} \\
HunyuanOCR-1.5~\hfilll~\cite{HunyuanOCR-1.5} & $1$\,B & 94.74 & 0.0390 & 94.50 & 93.67 & 94.71 & 0.1290 \\
Qianfan-OCR~\hfilll~\cite{Qianfan-OCR} & $4$\,B & 93.90 & 0.0400 & 95.08 & 90.53 & 93.31 & 0.1300 \\
Logics-Parsing-v2~\hfilll~\cite{Logics-Parsing} & $4$\,B & 93.33 & 0.0410 & 95.65 & 88.42 & 91.98 & 0.1370 \\
FireRed-OCR~\hfilll~\cite{FireRed-OCR} & $2$\,B & 93.26 & 0.0370 & 95.44 & 88.04 & 91.06 & 0.1310 \\
OpenDoc-0.1B~\hfilll~\cite{OpenDoc-0.1B} & $0.1$\,B & 90.67 & 0.0490 & 93.02 & 83.88 & 87.45 & 0.1400 \\
dots.ocr~\hfilll~\cite{dots.ocr} & $3$\,B & 90.77 & 0.0480 & 89.95 & 87.18 & 90.58 & 0.1380 \\
DeepSeek-OCR~2~\hfilll~\cite{DeepSeek-OCR-2} & $3$\,B & 90.25 & 0.0500 & 91.84 & 83.89 & 87.75 & 0.1440 \\
HunyuanOCR~\hfilll~\cite{HunyuanOCR} & $1$\,B & 89.95 & 0.0880 & 87.68 & 91.01 & 93.23 & 0.1710 \\
OCRVerse~\hfilll~\cite{OCRVerse} & $4$\,B & 88.60 & 0.0630 & 89.61 & 82.44 & 86.27 & 0.1630 \\
\midrule
\multicolumn{8}{@{}l}{\emph{General VLMs}} \\
Ovis2.6-30B-A3B~\hfilll~\cite{Ovis2.5} & $30$\,B-A$3$\,B & 93.70 & 0.0350 & 95.17 & 89.44 & 92.40 & 0.1350 \\
Gemini-3-Pro$^\dagger$~\hfilll~\cite{Gemini-3} & -- & 92.91 & 0.0640 & 95.99 & 89.15 & 92.96 & 0.1650 \\
Qwen3-VL-235B~\hfilll~\cite{Qwen3-VL} & $235$\,B-A$22$\,B & 89.78 & 0.0630 & 92.55 & 83.07 & 86.75 & 0.1660 \\
GPT-5.2$^\dagger$~\hfilll~\cite{GPT-5.2} & -- & 86.59 & 0.1140 & 88.21 & 82.95 & 87.93 & 0.1930 \\
InternVL3.5-241B~\hfilll~\cite{InternVL3.5} & $241$\,B-A$28$\,B & 83.76 & 0.1300 & 89.95 & 74.35 & 79.78 & 0.2150 \\
\midrule
\rowcolor{youtuBlue!7}\oursmark\OurMethodBF & $5$\,B & \best{96.96} & 0.0271 & 96.80 & \second{96.79} & \best{98.20} & \best{0.1104} \\
\bottomrule
\end{tabular}}
\end{table}

\section{Evaluation}
\label{sec:eval}

We evaluate a single checkpoint under a single inference configuration and organize the evaluation around two questions: what this checkpoint achieves across modalities, and how the two ideas behind the model, the text-initialized unified encoder and self-evolving post-training, contribute.
Subsec.~\ref{subsec:eval-setup} describes the benchmarks, baselines, and protocol, and Subsec.~\ref{subsec:eval-main-omnidocbench}--\ref{subsec:eval-main-symbolic} report the results on each benchmark.
Subsec.~\ref{subsec:eval-analysis} then studies the two principles of the encoder, one set of shared weights and initialization from a language model, in a controlled setting, and measures the contribution of self-evolving post-training on the final model.

\subsection{Setup}
\label{subsec:eval-setup}

\paragraph{Benchmarks and baselines.}
Tab.~\ref{tab:eval-protocol} lists every benchmark with its size, its metric, and the source of the baseline numbers.
\OmniDocBench v1.6~\citep{OmniDocBench} evaluates page-level document parsing; its Overall score averages text accuracy ($1-\text{edit distance}$), formula CDM, and table TEDS.
PureDocBench~\citep{PureDocBench} evaluates identical pages in clean, digitally degraded, and physically recaptured tracks, which isolates the cost of input degradation; it reports an Overall score per track and their mean, Avg$_3$.
\OmniParsingBench~\citep{Logics-Parsing-Omni} evaluates parsing of natural images, charts, geometry, audio, natural video, and text-rich video; each split is scored by a \emph{Perception} component over grounded fields, including localization, element extraction, OCR, and automatic speech recognition (ASR), and by a \emph{Cognition} component from multiple-choice questions that an LLM judge answers by reading only the parse.
Overall is the per-split leaderboard score: the mean of Perception and Cognition multiplied by the post-repair parse rate (Tab.~\ref{tab:omniparsing-image}--\ref{tab:omniparsing-textvideo}).
On \OmniParsingBench, every system is run and scored by us: we compare with the parsing-specialized \Logics, the open-weight generalists Qwen3-Omni-30B-A3B~\citep{Qwen3-Omni} and Qwen3.5-397B-A17B~\citep{Qwen3.5}, and the API models Gemini-3-Pro~\citep{Gemini-3} and GPT-5.4~\citep{GPT-5.4}; Qwen3.5 and GPT-5.4 are evaluated only on image splits because neither model accepts audio input.
Video-MME-v2~\citep{Video-MME-v2} evaluates general video understanding with $3{,}200$ multiple-choice questions arranged in groups of four; its group score rewards answering the related questions of a group consistently, and it is also reported for three capability levels (L1--L3). We use its setting without subtitles, in which models receive visual frames only.
ChemOCR~\citep{ChemVLM} evaluates chemical structure recognition from molecule images to SMILES, scored by the mean Tanimoto similarity of molecular fingerprints (Avg.\ Sim.) and the exact fingerprint-match rate (Tani@1.0).
PDMX-Synth~\citep{PDMX,LEGATO} evaluates optical music recognition from rendered score images to ABC notation, scored by character, symbol, and line error rates (CER, SER, and LER).
For \OmniDocBench, PureDocBench, and Video-MME-v2, the baseline numbers are taken from the official leaderboards; on ChemOCR and PDMX-Synth, we run every baseline with the official evaluator, except ChemDFM-X~\citep{ChemDFM-X} and LEGATO~\citep{LEGATO}, whose scores come from TinyChemVL~\citep{TinyChemVL} and LEGATO\citep{LEGATO}, respectively.

\paragraph{Protocol.}
\OmniDocBench, PureDocBench, Video-MME-v2, ChemOCR, and PDMX-Synth are scored with their official evaluation code or protocol.
\OmniParsingBench is released without a complete evaluation toolkit, so we implement one from the metric definitions of the benchmark (App.~\ref{sec:app-eval}) and re-score every system with it on one internal leaderboard.
These numbers, for our model and for every baseline, are thus comparable with each other, but not directly with those reported by the benchmark authors.
The toolkit repairs schema violations before scoring, so that formatting errors are not charged as content errors, and it uses corrected ground truth for a small number of verified annotation errors.
We set aside the camera-video subset because of annotation issues and exclude $136$ samples with unavailable media, leaving $3{,}738$ of $3{,}874$ samples.

\subsection{\OmniDocBench v1.6}
\label{subsec:eval-main-omnidocbench}

\begin{figure}[tp]
\centering
\includegraphics[width=\textwidth]{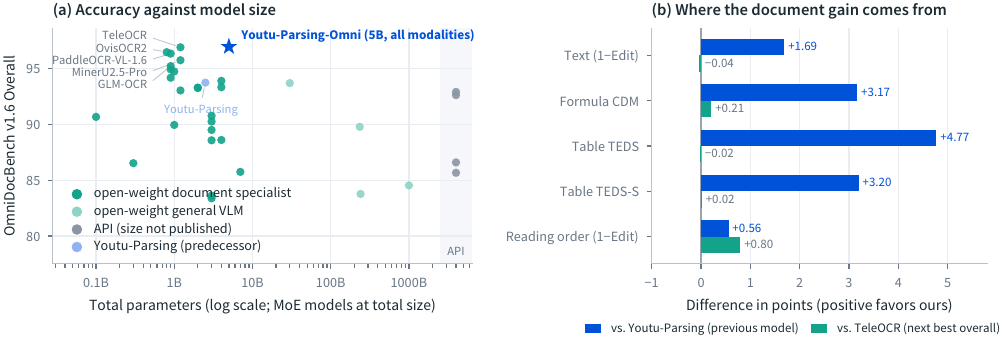}
\caption{\textbf{\OurMethod is the only model above $96$ on \OmniDocBench v1.6 that also parses images, audio, and video, and its gain over \OurMethodV1 comes mostly from tables and formulas.}
(a) \OmniDocBench v1.6 Overall against total parameters (teal: open-weight), with API systems in a strip at the right; the document specialists at our level have $0.8$--$1.2$\,B parameters.
(b) Per-metric difference to \OurMethodV1~\cite{Youtu-Parsing}, our previous model, and to TeleOCR~\cite{TeleOCR}, the next-best system overall; edit distances are shown as $1-\text{Edit}$ so that a positive difference always favors \OurMethod.}
\label{fig:docbench}
\end{figure}

We first investigate whether broadening a model's capabilities compromises its depth in document parsing, a domain in which specialized models are particularly strong.

The results indicate otherwise: while supporting a substantially broader range of input modalities and tasks, a single omni-modal model retains performance on par with specialized document parsers.
On \OmniDocBench v1.6, \OurMethod achieves the highest Overall score of $96.96$ (Tab.~\ref{tab:omnidocbench}).
More importantly, this result does not rely on an isolated advantage in any single metric, but instead reflects balanced parsing performance: the differences from TeleOCR on text, formulas, and tables are all within $0.21$ points, while \OurMethod attains the best reading-order accuracy (Fig.~\ref{fig:docbench}b).
Compared with \OurMethodV1, the improvements are concentrated on tables and formulas, with gains of $4.8$ TEDS and $3.2$ CDM, respectively.
Both content types require the model to recover their structure accurately and serialize it correctly.
This pattern suggests that the new model improves not only visual recognition, but also the organization of recognized content into coherent, structured outputs.
Notably, this specialist-level document parsing performance is delivered by the same model that processes natural images, audio, and video across ten tasks (Fig.~\ref{fig:docbench}a).
Thus, the broader modality coverage is achieved without weakening document understanding or fine-grained parsing capabilities.

\subsection{PureDocBench}
\label{subsec:eval-main-puredocbench}

\begin{table}[!t]
\centering
\caption{\textbf{Comparison with existing document parsing models and proprietary models on PureDocBench under clean and degraded conditions.}
Numbers are from the public leaderboard~\cite{PureDocBench}.
$^\dagger$Proprietary model.
\textbf{Bold} and \underline{underline}: best and second best of each Overall column.
}
\label{tab:puredocbench}
\small
\setlength{\tabcolsep}{3pt}
\resizebox{\textwidth}{!}{%
\begin{tabular}{l r cccc cccc cccc c}
\toprule
 & & \multicolumn{4}{c}{Clean} & \multicolumn{4}{c}{Digitally degraded} & \multicolumn{4}{c}{Physically recaptured} & \\
\cmidrule(lr){3-6} \cmidrule(lr){7-10} \cmidrule(lr){11-14}
 Model & Params & Overall$\uparrow$ & Text Edit$\downarrow$ & Formula CDM$\uparrow$ & Table TEDS$\uparrow$ & Overall$\uparrow$ & Text Edit$\downarrow$ & Formula CDM$\uparrow$ & Table TEDS$\uparrow$ & Overall$\uparrow$ & Text Edit$\downarrow$ & Formula CDM$\uparrow$ & Table TEDS$\uparrow$ & Avg$_3\uparrow$ \\
\midrule
\multicolumn{15}{@{}l}{\emph{Multi-stage specialists}} \\
TeleOCR~\hfilll~\cite{TeleOCR} & $1.2$\,B & \best{86.90} & 0.111 & 81.01 & 91.09 & \second{77.47} & 0.206 & 72.59 & 80.45 & \second{70.85} & 0.302 & 65.11 & 77.66 & \best{78.41} \\
dots.mocr~\hfilll~\cite{dots.mocr} & $3$\,B & 76.27 & 0.151 & 66.23 & 77.65 & 73.16 & 0.198 & 64.32 & 74.95 & 61.73 & 0.312 & 54.39 & 61.97 & 70.39 \\
MinerU2.5-Pro~\hfilll~\cite{MinerU2.5-Pro} & $1.2$\,B & 75.87 & 0.222 & 65.14 & 84.68 & 71.77 & 0.272 & 61.79 & 80.73 & 62.56 & 0.375 & 52.70 & 72.47 & 70.07 \\
\OurMethodV1~\hfilll~\cite{Youtu-Parsing} & $2.5$\,B & 75.02 & 0.230 & 67.34 & 80.74 & 69.66 & 0.270 & 61.44 & 74.49 & 60.29 & 0.360 & 52.20 & 64.69 & 68.32 \\
PaddleOCR-VL-1.5~\hfilll~\cite{PaddleOCR-VL-1.5} & $0.9$\,B & 73.01 & 0.266 & 63.53 & 82.12 & 66.73 & 0.339 & 58.03 & 76.07 & 60.50 & 0.398 & 54.00 & 67.33 & 66.75 \\
GLM-OCR~\hfilll~\cite{GLM-OCR} & $0.9$\,B & 68.65 & 0.314 & 57.89 & 79.44 & 63.06 & 0.383 & 53.23 & 74.21 & 58.31 & 0.433 & 50.34 & 67.83 & 63.34 \\
Dolphin-v2~\hfilll~\cite{Dolphin} & $3$\,B & 65.90 & 0.342 & 59.80 & 72.12 & 60.24 & 0.393 & 52.20 & 67.86 & 44.92 & 0.553 & 39.98 & 50.04 & 57.02 \\
MonkeyOCR-pro-3B~\hfilll~\cite{MonkeyOCR} & $3$\,B & 62.23 & 0.346 & 48.46 & 72.83 & 57.40 & 0.397 & 45.57 & 66.32 & 46.49 & 0.511 & 38.18 & 52.43 & 55.37 \\
MonkeyOCR-pro-1.2B~\hfilll~\cite{MonkeyOCR} & $1.2$\,B & 61.09 & 0.358 & 47.43 & 71.60 & 55.72 & 0.416 & 43.91 & 64.83 & 43.82 & 0.556 & 36.94 & 50.07 & 53.54 \\
\midrule
\multicolumn{15}{@{}l}{\emph{End-to-end specialists}} \\
OvisOCR2~\hfilll~\cite{OvisOCR2} & $0.8$\,B & 81.55 & -- & -- & -- & 77.09 & -- & -- & -- & 66.56 & -- & -- & -- & 75.06 \\
FD-RL~\hfilll~\cite{FD-RL} & $4$\,B & 78.38 & 0.193 & 68.21 & 86.22 & 76.33 & 0.214 & 67.16 & 83.22 & 67.04 & 0.298 & 58.82 & 72.08 & 73.92 \\
Logics-Parsing-v2~\hfilll~\cite{Logics-Parsing} & $4$\,B & 76.35 & 0.213 & 67.67 & 82.67 & 73.85 & 0.248 & 67.33 & 79.02 & 67.64 & 0.304 & 61.65 & 71.64 & 72.61 \\
OCRVerse~\hfilll~\cite{OCRVerse} & $4$\,B & 73.18 & 0.273 & 63.78 & 83.09 & 71.36 & 0.302 & 63.95 & 80.36 & 63.66 & 0.363 & 57.03 & 70.30 & 69.40 \\
FireRed-OCR~\hfilll~\cite{FireRed-OCR} & $2$\,B & 70.81 & 0.287 & 63.86 & 77.23 & 68.49 & 0.319 & 62.64 & 74.77 & 57.42 & 0.415 & 51.60 & 62.16 & 65.57 \\
dots.ocr~\hfilll~\cite{dots.ocr} & $3$\,B & 72.01 & 0.248 & 61.37 & 79.51 & 65.95 & 0.307 & 56.67 & 71.86 & 55.68 & 0.403 & 47.70 & 59.63 & 64.55 \\
olmOCR-2-7B~\hfilll~\cite{olmOCR-2} & $7$\,B & 69.36 & 0.284 & 56.89 & 79.59 & 65.87 & 0.318 & 54.57 & 74.81 & 56.10 & 0.417 & 48.79 & 61.25 & 63.78 \\
HunyuanOCR~\hfilll~\cite{HunyuanOCR} & $1$\,B & 65.61 & 0.269 & 55.74 & 68.02 & 61.49 & 0.308 & 51.62 & 63.68 & 54.58 & 0.421 & 48.30 & 57.54 & 60.56 \\
olmOCR-7B~\hfilll~\cite{olmOCR} & $7$\,B & 62.56 & 0.388 & 58.69 & 67.77 & 57.84 & 0.436 & 55.44 & 61.66 & 47.30 & 0.542 & 46.26 & 49.80 & 55.90 \\
OpenDoc-0.1B~\hfilll~\cite{OpenDoc-0.1B} & $0.1$\,B & 60.28 & 0.411 & 53.09 & 68.86 & 52.46 & 0.501 & 48.41 & 59.04 & 44.27 & 0.547 & 38.46 & 49.06 & 52.34 \\
Qianfan-OCR~\hfilll~\cite{Qianfan-OCR} & $4$\,B & 57.22 & 0.370 & 49.79 & 58.83 & 50.85 & 0.438 & 44.41 & 51.96 & 45.06 & 0.494 & 39.08 & 45.53 & 51.04 \\
DeepSeek-OCR~2~\hfilll~\cite{DeepSeek-OCR-2} & $3$\,B & 55.53 & 0.354 & 46.00 & 56.01 & 49.41 & 0.412 & 40.78 & 48.67 & 43.60 & 0.486 & 37.30 & 42.06 & 49.51 \\
DeepSeek-OCR~\hfilll~\cite{DeepSeek-OCR} & $3$\,B & 53.50 & 0.419 & 45.39 & 57.06 & 46.95 & 0.478 & 39.99 & 48.64 & 40.48 & 0.537 & 34.04 & 41.12 & 46.98 \\
OCRFlux-3B~\hfilll~\cite{OCRFlux} & $3$\,B & 47.14 & 0.454 & 38.35 & 48.46 & 41.82 & 0.486 & 31.90 & 42.17 & 37.21 & 0.559 & 32.65 & 34.87 & 42.06 \\
\midrule
\multicolumn{15}{@{}l}{\emph{General VLMs}} \\
Gemini-3.1-Pro$^\dagger$~\hfilll~\cite{Gemini-3.1-Pro} & -- & 70.04 & 0.306 & 65.63 & 75.08 & 69.28 & 0.322 & 65.81 & 74.24 & \best{71.98} & 0.300 & 68.62 & 77.26 & 70.43 \\
Kimi K2.6~\hfilll~\cite{Kimi-K2.6} & $1{,}000$\,B-A$32$\,B & 72.32 & 0.303 & 66.93 & 80.30 & 69.95 & 0.322 & 64.69 & 77.31 & 68.02 & 0.335 & 62.44 & 75.14 & 70.10 \\
GPT-5.6$^\dagger$~\hfilll~\cite{GPT-5.6} & -- & 74.15 & 0.354 & 74.96 & 82.87 & 71.73 & 0.380 & 72.83 & 80.33 & 61.59 & 0.472 & 62.26 & 69.75 & 69.16 \\
Qwen3.5-397B-A17B~\hfilll~\cite{Qwen3.5} & $397$\,B-A$17$\,B & 69.12 & 0.233 & 65.26 & 65.40 & 68.34 & 0.244 & 63.91 & 65.53 & 62.70 & 0.287 & 60.70 & 56.12 & 66.72 \\
\midrule
\rowcolor{youtuBlue!7}\oursmark\OurMethodBF & $5$\,B & \second{82.61} & 0.164 & 76.37 & 87.89 & \best{78.13} & 0.208 & 72.43 & 82.79 & 68.13 & 0.312 & 62.31 & 73.22 & \second{76.29} \\
\bottomrule
\end{tabular}}
\end{table}

\begin{table}[!t]
\centering
\caption{
    \textbf{Comparison with open-weight and proprietary models on each split of OmniParsingBench.}
    Scores are the per-split Overall (Subsec.~\ref{subsec:eval-setup}); higher is better. \textbf{Bold} and \underline{underline}: best and second best in each column.
    Avg.\ is the average over five columns, in which chart and geometry enter together as one graphics column.
    All systems are scored with the same toolkit.
    Perception and Cognition are in Tab.~\ref{tab:omniparsing-image}--\ref{tab:omniparsing-textvideo}.
    }
\label{tab:omniparsing}
\small\setlength{\tabcolsep}{4.5pt}
\begin{tabular}{l cccccc c}
\toprule
 Model & Chart & Geometry & Nat.\ image & Audio & Nat.\ video & Text-rich video & Avg. \\
\midrule
\multicolumn{8}{l}{\emph{Proprietary / API}} \\
Gemini-3-Pro~\hfilll~\cite{Gemini-3} & 92.79 & \best{85.43} & \best{69.73} & \second{76.74} & \best{74.82} & \best{76.80} & \best{77.44} \\
GPT-5.4~\hfilll~\cite{GPT-5.4} & 90.92 & \second{81.16} & 60.50 & -- & -- & -- & -- \\
\midrule
\multicolumn{8}{l}{\emph{Open-weight}} \\
Logics-Parsing-Omni~\hfilll~\cite{Logics-Parsing-Omni} & \second{94.65} & 76.34 & \second{67.53} & 75.58 & 67.34 & 65.36 & 72.26 \\
Qwen3-Omni-30B-A3B~\hfilll~\cite{Qwen3-Omni} & 87.07 & 58.60 & 59.35 & 75.23 & 65.19 & 59.59 & 66.44 \\
Qwen3.5-397B-A17B~\hfilll~\cite{Qwen3.5} & 93.51 & 72.16 & 63.87 & -- & -- & -- & -- \\
\midrule
\rowcolor{youtuBlue!7}\oursmark\OurMethodBF & \best{95.02} & 74.33 & 62.84 & \best{78.08} & \second{74.18} & \second{75.62} & \second{75.08} \\
\bottomrule
\end{tabular}
\end{table}

We evaluate the model's robustness to degradation introduced during acquisition on PureDocBench, which comprises clean, digitally degraded, and physically recaptured inputs.
\OurMethod achieves the highest Avg$_3$ among end-to-end models ($76.29$; Tab.~\ref{tab:puredocbench}) and outperforms every end-to-end specialist on all three tracks.
Its consistent lead over end-to-end specialists across both clean and degraded conditions indicates that the advantage stems from robust parsing rather than specialization to a particular acquisition condition.
Compared with \OurMethodV1, \OurMethod gains approximately $8$ points on each of the three tracks.
The gains span all three component metrics and are more pronounced for formulas and tables, indicating that the model better preserves its ability to parse structured content under degradation rather than merely improving text recognition.
Nevertheless, most systems exhibit a marked performance decline on physically recaptured inputs, indicating that distortions and artifacts introduced during real-world acquisition remain a major challenge for document parsing.

\subsection{\OmniParsingBench}
\label{subsec:eval-main-omniparsing}

\begin{table}[!t]
\centering
\caption{\textbf{Detailed evaluation results on the Natural Image module of OmniParsingBench.}
Perception Avg. is the mean of the three Content columns (Entity, Know., and Text). Overall is the mean of Perception and Cognition multiplied by the parse rate. Parse rate (\%) is the fraction of outputs successfully parsed after repair. \textbf{Bold} and \underline{underline}: best and second best in each column.}
\label{tab:omniparsing-image}
\setlength{\tabcolsep}{3pt}
\begin{adjustbox}{max width=0.99\textwidth}
\begin{tabular}{@{}lccccccccccccccccccc@{}}
\toprule
\multirow{3}{*}{Model} & \multirow{3}{*}{Overall} & \multirow{3}{*}{Parse rate} & \multicolumn{7}{c}{Perception} & \multicolumn{10}{c}{Cognition} \\
\cmidrule(lr){4-10}\cmidrule(lr){11-20}
 &  &  & \multicolumn{3}{c}{Location} & \multicolumn{3}{c}{Content} & \multirow{2}{*}{Avg.} & \multirow{2}{*}{Theme} & \multirow{2}{*}{Emot.} & \multirow{2}{*}{Scene} & \multirow{2}{*}{Count} & \multirow{2}{*}{Attr.} & \multirow{2}{*}{Loc.} & \multirow{2}{*}{Text} & \multirow{2}{*}{Style} & \multirow{2}{*}{Comp.} & \multirow{2}{*}{Avg.} \\
\cmidrule(lr){4-6}\cmidrule(lr){7-9}
 &  &  & Entity & Know. & Text & Entity & Know. & Text &  &  &  &  &  &  &  &  &  &  &  \\
\midrule
Gemini-3-Pro & \textbf{69.73} & \textbf{100.00} & \underline{67.37} & 83.16 & \textbf{72.54} & 63.70 & \textbf{32.89} & \textbf{48.59} & \textbf{48.39} & \textbf{98.94} & \textbf{99.02} & \underline{96.62} & \underline{77.15} & \underline{91.48} & \textbf{88.83} & \textbf{83.76} & \underline{96.68} & \underline{87.17} & \underline{91.07} \\

Qwen3-Omni-30B-A3B & 59.35 & \underline{99.30} & 62.78 & 82.48 & 50.71 & 59.03 & 16.71 & 30.00 & 35.25 & 98.13 & \underline{95.00} & 94.05 & 70.79 & 82.30 & 76.83 & 80.21 & 85.04 & 76.26 & 84.29 \\
Logics-Parsing-Omni & \underline{67.53} & \textbf{100.00} & 66.77 & \underline{89.84} & \underline{67.61} & \underline{64.05} & \underline{21.39} & \underline{44.37} & \underline{43.27} & \underline{98.41} & \textbf{99.02} & \textbf{98.87} & \textbf{79.12} & \textbf{93.67} & \underline{87.59} & \underline{81.73} & \textbf{98.89} & \textbf{88.78} & \textbf{91.79} \\
\midrule
\rowcolor{youtuBlue!7}\oursmark\OurMethodBF & 62.84 & \textbf{100.00} & \textbf{74.60} & \textbf{96.26} & 52.11 & \textbf{68.39} & 13.90 & 30.28 & 37.52 & 95.23 & \textbf{99.02} & 96.34 & 72.48 & 89.02 & 85.46 & 74.11 & 96.12 & 85.57 & 88.15 \\
\bottomrule
\end{tabular}%
\end{adjustbox}
\end{table}

\begin{table}[!t]
\centering
\caption{\textbf{Detailed evaluation results on the Graphics -- Chart module of OmniParsingBench.} Parse rate (\%) is the fraction of outputs successfully parsed after repair. \textbf{Bold} and \underline{underline}: best and second best in each column.}
\label{tab:omniparsing-chart}
\setlength{\tabcolsep}{3pt}
\begin{adjustbox}{max width=0.99\textwidth}
\begin{tabular}{@{}lccccccccc@{}}
\toprule
\multirow{2}{*}{Model} & \multirow{2}{*}{Overall} & \multirow{2}{*}{Parse rate} & \multicolumn{3}{c}{Perception} & \multicolumn{4}{c}{Cognition} \\
\cmidrule(lr){4-6}\cmidrule(lr){7-10}
 &  &  & Stat.\ Charts ($F_1$) & Struct.\ Diagrams ($F_1$) & Avg. & OCR \& Data & Visual Elements & Logical Reasoning & Avg.\ Acc. \\
\midrule
Gemini-3-Pro & 92.79 & \textbf{100.00} & 92.42 & \textbf{99.27} & \underline{95.85} & \textbf{99.29} & 82.24 & 87.67 & 89.74 \\
Qwen3-Omni-30B-A3B & 87.07 & \textbf{100.00} & 79.79 & 94.47 & 87.13 & 95.77 & 79.91 & 85.39 & 87.02 \\
Logics-Parsing-Omni & \underline{94.65} & \textbf{100.00} & \underline{93.26} & 95.78 & 94.52 & \underline{99.12} & \textbf{93.46} & \textbf{91.78} & \textbf{94.79} \\
\midrule
\rowcolor{youtuBlue!7}\oursmark\OurMethodBF & \textbf{95.02} & \underline{99.80} & \textbf{97.16} & \underline{97.82} & \textbf{97.49} & \underline{99.12} & \underline{91.59} & \underline{88.07} & \underline{92.93} \\
\bottomrule
\end{tabular}%
\end{adjustbox}
\end{table}

\begin{table}[!t]
\centering
\caption{\textbf{Detailed evaluation results on the Graphics -- Geometry module of OmniParsingBench.} Perception Avg. weights the Positioning, Existence, and Relation group means by $0.3$, $0.3$, and $0.4$, respectively. Parse rate (\%) is the fraction of outputs successfully parsed after repair. \textbf{Bold} and \underline{underline}: best and second best in each column.}
\label{tab:omniparsing-geometry}
\setlength{\tabcolsep}{3pt}
\begin{adjustbox}{max width=0.99\textwidth}
\begin{tabular}{@{}lccccccccccccccc@{}}
\toprule
\multirow{3}{*}{Model} & \multirow{3}{*}{Overall} & \multirow{3}{*}{Parse rate} & \multicolumn{9}{c}{Perception} & \multicolumn{4}{c}{Cognition} \\
\cmidrule(lr){4-12}\cmidrule(lr){13-16}
 &  &  & \multicolumn{2}{c}{Positioning} & \multicolumn{4}{c}{Existence} & \multicolumn{2}{c}{Relation} & \multirow{2}{*}{Avg.} & \multirow{2}{*}{{\renewcommand{\arraystretch}{0.9}\begin{tabular}{@{}c@{}}Basic\\Elements\end{tabular}}} & \multirow{2}{*}{{\renewcommand{\arraystretch}{0.9}\begin{tabular}{@{}c@{}}Geometric\\Relations\end{tabular}}} & \multirow{2}{*}{{\renewcommand{\arraystretch}{0.9}\begin{tabular}{@{}c@{}}Quantitative\\Relations\end{tabular}}} & \multirow{2}{*}{Avg.} \\
\cmidrule(lr){4-5}\cmidrule(lr){6-9}\cmidrule(lr){10-11}
 &  &  & Coord. & Radius & Point & Line/Arc & Circle & Membership & Geo. & Quant. &  &  &  &  &  \\
\midrule
Gemini-3-Pro & \textbf{85.43} & \underline{99.80} & \textbf{81.60} & \underline{42.56} & \textbf{99.02} & \textbf{89.64} & \textbf{98.61} & 63.09 & \textbf{80.70} & \textbf{82.93} & \textbf{77.63} & \underline{90.30} & \textbf{93.49} & \textbf{96.91} & \textbf{93.57} \\
Qwen3-Omni-30B-A3B & 58.60 & 96.60 & 24.58 & 15.86 & 91.18 & \underline{78.69} & 92.86 & 56.57 & 35.19 & 29.94 & 43.04 & 86.61 & 71.93 & 76.34 & 78.29 \\
Logics-Parsing-Omni & \underline{76.34} & \textbf{100.00} & 41.80 & 40.16 & \underline{98.79} & 75.32 & 65.28 & \underline{79.89} & \underline{64.24} & \underline{56.14} & 60.32 & \textbf{95.90} & \underline{86.35} & \underline{94.85} & \underline{92.37} \\
\midrule
\rowcolor{youtuBlue!7}\oursmark\OurMethodBF & 74.33 & \textbf{100.00} & \underline{56.42} & \textbf{55.34} & 97.35 & 75.26 & \underline{94.44} & \textbf{94.31} & 46.77 & 49.59 & \underline{63.14} & 88.36 & 76.45 & 91.75 & 85.52 \\
\bottomrule
\end{tabular}%
\end{adjustbox}
\end{table}

\paragraph{Overall results.}
We next ask whether a single encoder can serve images, audio, and video at a competitive level, and where any advantage appears.
Under our evaluation toolkit, \OurMethod is the strongest open-weight model on \OmniParsingBench, with an average of $75.08$ against $72.26$ for \Logics; among the models evaluated on every split, only the API model Gemini-3-Pro scores higher ($77.44$; Tab.~\ref{tab:omniparsing}, Fig.~\ref{fig:teaser}b).
It leads the other open-weight models on all three temporal splits (audio, natural video, and text-rich video) and attains the highest chart score of all systems ($95.02$).
The advantage, however, is not uniform across splits.
Relative to \Logics, \OurMethod leads by $2.5$ points on audio, $6.8$ on natural video, and $10.3$ on text-rich video ($6.5$ on average), whereas it trails by $2.8$ points on average on the two static splits (natural images and graphics), mainly on natural images ($-4.7$) and geometry ($-2.0$).
These group means follow the leaderboard weighting, combining chart and geometry into one graphics score before averaging.
Relative to Gemini-3-Pro, it is on par on the temporal splits ($75.96$ against $76.12$ on average) but trails on the static ones ($73.76$ against $79.42$).
\Logics pairs a $1.1$\,B two-tower front end with a $30$\,B mixture-of-experts (MoE) decoder that activates about $3$\,B parameters per token: fewer active parameters than our dense $5$\,B model, but about six times as many in total.
The contrast between static and temporal splits is consistent with encoding all modalities on a shared timeline (Sec.~\ref{sec:model}), whose benefit should be largest where a parse must align events in time and across modalities.
Overall, with a single encoder, \OurMethod is the strongest open-weight model on \OmniParsingBench, and its advantage is concentrated on the temporal splits.

\paragraph{Natural images.}
\OurMethod leads in entity localization and plain-entity recognition (Tab.~\ref{tab:omniparsing-image}).
Knowledge recognition is weaker ($13.90$ versus $21.39$ for \Logics), possibly reflecting our smaller total parameter budget (e.g., \Logics uses a $30$\,B-A$3$\,B decoder).
Counting remains challenging across models, with all Count scores below $80$ despite strong Theme and Scene scores.

\paragraph{Graphics chart and geometry.}
\OurMethod achieves the highest chart Perception ($97.49$) and Overall ($95.02$) scores in the breakdown (Tab.~\ref{tab:omniparsing-chart}), with its clearest advantage in statistical-chart extraction.
In terms of Cognition score, our method is lower than \Logics ($92.93$ versus $94.79$), with the gap concentrated in visual-element and logical-reasoning questions. This is mainly because our captions tend to provide globalized descriptions and weaken the semantic descriptions of color and position, since such information contains limited effective content and is prone to introducing semantic errors.
For geometry (Tab.~\ref{tab:omniparsing-geometry}), \OurMethod also exceeds \Logics in Perception ($63.14$ versus $60.32$), with gains in positioning, circle detection, and membership.
Geometric relations remain weaker in both Perception ($46.77$ versus $64.24$) and Cognition ($76.45$ versus $86.35$).
Together, these results suggest that richer descriptions of graphical elements and their relationships could improve how the parses support downstream questions.

\begin{table}[!t]
\centering
\caption{\textbf{Detailed evaluation results on the Audio module of OmniParsingBench.} Parse rate (\%) is the fraction of outputs successfully parsed after repair. \textbf{Bold} and \underline{underline}: best and second best in each column.}
\label{tab:omniparsing-audio}
\setlength{\tabcolsep}{3pt}
\begin{adjustbox}{max width=0.99\textwidth}
\begin{tabular}{@{}lccccccccc@{}}
\toprule
\multirow{2}{*}{Model} & \multirow{2}{*}{Overall} & \multirow{2}{*}{Parse rate} & \multicolumn{4}{c}{Perception} & \multicolumn{3}{c}{Cognition} \\
\cmidrule(lr){4-7}\cmidrule(lr){8-10}
 &  &  & Time & Key & ASR & Avg. & Info.\ Ext. & Rec. & Avg. \\
\midrule
Gemini-3-Pro & \underline{76.74} & \underline{99.89} & \underline{51.53} & \underline{61.30} & \underline{84.38} & \textbf{65.74} & \textbf{97.32} & 78.47 & 87.90 \\
Qwen3-Omni-30B-A3B & 75.23 & \textbf{100.00} & 35.02 & \textbf{67.29} & \textbf{84.57} & 62.29 & 96.43 & 79.90 & 88.17 \\
Logics-Parsing-Omni & 75.58 & \textbf{100.00} & 44.78 & 46.36 & 83.56 & 58.23 & \underline{96.88} & \underline{89.00} & \textbf{92.94} \\
\midrule
\rowcolor{youtuBlue!7}\oursmark\OurMethodBF & \textbf{78.08} & \textbf{100.00} & \textbf{53.89} & 59.48 & 77.94 & \underline{63.77} & 91.96 & \textbf{92.82} & \underline{92.39} \\
\bottomrule
\end{tabular}%
\end{adjustbox}
\end{table}

\begin{table}[!t]
\centering
\caption{\textbf{Detailed evaluation results on the Natural Video module of OmniParsingBench.} Parse rate (\%) is the fraction of outputs successfully parsed after repair. \textbf{Bold} and \underline{underline}: best and second best in each column.}
\label{tab:omniparsing-video}
\setlength{\tabcolsep}{3pt}
\begin{adjustbox}{max width=0.99\textwidth}
\begin{tabular}{@{}lcccccccccccc@{}}
\toprule
\multirow{3}{*}{Model} & \multirow{3}{*}{Overall} & \multirow{3}{*}{Parse rate} & \multicolumn{5}{c}{Perception} & \multicolumn{5}{c}{Cognition} \\
\cmidrule(lr){4-8}\cmidrule(lr){9-13}
 &  &  & \multirow{2}{*}{Time} & \multirow{2}{*}{Key} & \multirow{2}{*}{ASR} & \multirow{2}{*}{Visual} & \multirow{2}{*}{Avg.} & \multicolumn{2}{c}{Visual} & \multicolumn{2}{c}{Audio-Visual} & \multirow{2}{*}{Avg.} \\
\cmidrule(lr){9-10}\cmidrule(lr){11-12}
 &  &  &  &  &  &  &  & \shortstack{Object/\\Attribute} & Event & Emot. & \shortstack{Caus.\\Rsn.} &  \\
\midrule
Gemini-3-Pro & \textbf{74.82} & \textbf{100.00} & \underline{59.47} & \textbf{43.94} & \textbf{84.06} & \underline{44.80} & \textbf{58.07} & 85.11 & \underline{88.24} & \underline{96.45} & \textbf{96.49} & \underline{91.57} \\
Qwen3-Omni-30B-A3B & 65.19 & \textbf{100.00} & 35.08 & 29.96 & 62.46 & 38.80 & 41.58 & 80.85 & \textbf{89.08} & 90.53 & \underline{94.74} & 88.80 \\
Logics-Parsing-Omni & 67.34 & \textbf{100.00} & 34.62 & 31.04 & \underline{83.55} & 28.31 & 44.38 & \underline{88.65} & 84.87 & 94.67 & 92.98 & 90.30 \\
\midrule
\rowcolor{youtuBlue!7}\oursmark\OurMethodBF & \underline{74.18} & \textbf{100.00} & \textbf{63.68} & \underline{41.40} & 67.80 & \textbf{45.82} & \underline{54.67} & \textbf{92.91} & 86.55 & \textbf{98.82} & \textbf{96.49} & \textbf{93.69} \\
\bottomrule
\end{tabular}%
\end{adjustbox}
\end{table}

\begin{table}[!t]
\centering
\caption{\textbf{Detailed evaluation results on the Text-Rich Video module of OmniParsingBench.} Perception OCR measures on-screen text transcription. Cognition Avg. equally weights the mean of the seven parsing dimensions and the mean of Core, Outline, and Deep. Parse rate (\%) is the fraction of outputs successfully parsed after repair. \textbf{Bold} and \underline{underline}: best and second best in each column.}
\label{tab:omniparsing-textvideo}
\setlength{\tabcolsep}{3pt}
\begin{adjustbox}{max width=0.99\textwidth}
\begin{tabular}{@{}lcccccccccccccccccc@{}}
\toprule
\multirow{3}{*}{Model} & \multirow{3}{*}{Overall} & \multirow{3}{*}{Parse rate} & \multicolumn{4}{c}{Perception} & \multicolumn{12}{c}{Cognition} \\
\cmidrule(lr){4-7}\cmidrule(lr){8-19}
 &  &  & \multirow{2}{*}{Time} & \multirow{2}{*}{OCR} & \multirow{2}{*}{ASR} & \multirow{2}{*}{Avg.} & \multicolumn{1}{c}{Audio} & \multicolumn{4}{c}{Visual} & \multicolumn{2}{c}{Audio-Visual} & \multicolumn{4}{c}{In-Depth Str.\ Cap.} & \multirow{2}{*}{Avg.} \\
\cmidrule(lr){8-8}\cmidrule(lr){9-12}\cmidrule(lr){13-14}\cmidrule(lr){15-18}
 &  &  &  &  &  &  & \shortstack{Info.\\Ext.} & Object & OCR & \shortstack{Attr.\&\\States} & Event & \shortstack{Info.\\Syn.} & \shortstack{Caus.\\Rsn.} & Core & Outline & Deep & Avg. &  \\
\midrule
Gemini-3-Pro & \textbf{76.80} & \textbf{100.00} & \textbf{61.56} & 42.50 & \textbf{77.61} & \textbf{60.56} & \underline{97.41} & 73.83 & 92.68 & \underline{62.50} & 84.25 & \textbf{98.05} & \textbf{100.00} & \textbf{100.00} & \textbf{98.67} & \underline{98.73} & \textbf{99.13} & 93.05 \\
Qwen3-Omni-30B-A3B & 59.59 & \underline{98.40} & 13.94 & 26.83 & 49.39 & 30.06 & 93.42 & \underline{74.76} & 88.18 & 55.94 & 80.95 & 92.57 & \underline{99.51} & \textbf{100.00} & 97.15 & 98.37 & 98.51 & 91.06 \\
Logics-Parsing-Omni & 65.36 & \textbf{100.00} & 23.05 & \textbf{52.68} & 34.28 & 36.67 & 96.55 & 71.96 & \textbf{95.61} & \textbf{75.69} & \underline{88.19} & 96.10 & \textbf{100.00} & \textbf{100.00} & 97.93 & \textbf{98.93} & \underline{98.96} & \underline{94.06} \\
\midrule
\rowcolor{youtuBlue!7}\oursmark\OurMethodBF & \underline{75.62} & \textbf{100.00} & \underline{50.70} & \underline{42.67} & \underline{76.52} & \underline{56.63} & \textbf{98.71} & \textbf{78.50} & \underline{95.12} & \textbf{75.69} & \textbf{88.98} & \underline{97.56} & \textbf{100.00} & \textbf{100.00} & \underline{98.33} & 97.33 & 98.56 & \textbf{94.60} \\
\bottomrule
\end{tabular}%
\end{adjustbox}
\end{table}

\paragraph{Audio.}
\OurMethod's lead over \Logics comes from more accurate temporal localization and event keywords, with gains of $9.1$ and $13.1$ points, respectively (Tab.~\ref{tab:omniparsing-audio}).
These improvements compensate for an ASR score of $77.94$, the lowest among the compared models, revealing a stronger ability to locate and describe acoustic events than to transcribe speech relative to the baselines.
With Cognition close to \Logics ($92.39$ versus $92.94$), the highest Overall score ($78.08$) is chiefly supported by these perceptual gains.

\begin{table}[!t]
  \centering
  \caption{
    \textbf{Comparison with open-weight models on Video-MME-v2 without subtitles.}
    Baseline numbers are taken from the official leaderboard~\citep{Video-MME-v2}.
    Higher is better; \textbf{bold} and \underline{underline}: best and second best in each column.
  }
  \label{tab:video-bench}
  \small
  \setlength{\tabcolsep}{5pt}
  \begin{adjustbox}{max width=0.99\textwidth}
    \begin{tabular}{l cccc}
     \toprule
  
      & \multicolumn{4}{c}{{Video-MME-v2 without subtitles}} \\
  
     \cmidrule{2-5}
  
     \multirow{-2}{*}{Model} & Group score & L1 & L2 & L3 \\
  
     \midrule
  
     Qwen3-Omni-30B-A3B-Think~\hfilll~\cite{Qwen3-Omni}    & \textbf{12.6} & \underline{15.8} & \textbf{12.3} & \textbf{11.0} \\
  
     Qwen3-VL-4B-Think~\hfilll~\cite{Qwen3-VL}             & \underline{11.6} & 15.0 & \underline{11.3} & 9.8 \\
  
     Qwen3.5-4B-Think~\hfilll~\cite{Qwen3.5}               & \underline{11.6} & 14.9 & 11.1 & \underline{9.9} \\
  
     InternVL3.5-8B-Instruct~\hfilll~\cite{InternVL3.5}    & 11.4 & 14.8 & 10.8 & 9.8 \\
  
     Qwen3-Omni-30B-A3B-Instruct~\hfilll~\cite{Qwen3-Omni} & 10.9 & 15.0 & 10.2 & 8.9 \\
  
     InternVL3.5-4B-Instruct~\hfilll~\cite{InternVL3.5}    & 9.9  & 13.5 & 9.2  & 8.2 \\
  
     \midrule
     \rowcolor{youtuBlue!7}\oursmark\OurMethodBF & 11.5 & \textbf{20.0} & 10.4 & 9.2 \\
  
     \bottomrule
    \end{tabular}
  \end{adjustbox}
  \end{table}

\paragraph{Natural and text-rich video.}
Across both natural and text-rich video settings, \OurMethod achieves the best Overall performance among open-weight models by virtue of its powerful temporally grounded parsing and robust semantic preservation.
In natural videos (Tab.~\ref{tab:omniparsing-video}), it obtains a superior temporal localization score of $63.68$, outperforming Gemini-3-Pro, \Logics, and Qwen3-Omni-30B-A3B by $4.21$, $29.06$, and $28.60$ points, respectively, while reaching the highest Cognition average of $93.69$ across all baselines, with leading results in object and attribute understanding ($92.91$), audio-visual emotion understanding ($98.82$), and joint-best audio-visual causal reasoning ($96.49$).
Notably, even with a lower ASR score ($67.80$) than Gemini-3-Pro ($84.06$), its Overall score of $74.18$ surpasses \Logics by $6.84$ points and approaches the proprietary Gemini-3-Pro within only $0.64$ points, demonstrating that accurate temporal structural parsing effectively compensates for speech transcription weaknesses and maintains high-level semantic fidelity.
In more challenging text-rich videos (Tab.~\ref{tab:omniparsing-textvideo}), the model's advantages are further strengthened, yielding an Overall score of $75.62$ that expands the lead over \Logics to $10.26$ points, driven by substantial improvements in temporal localization ($50.70$ versus $23.05$) and speech transcription ($76.52$ versus $34.28$).
It also retains the highest Cognition average ($94.60$) with state-of-the-art audio information extraction ($98.71$) and object understanding ($78.50$), while matching the best audio-visual causal reasoning score ($100.00$).
Consistent superior Cognition results across both video splits, evaluated purely based on parsed outputs, verify that our unified temporal modeling paradigm effectively preserves multimodal semantics for downstream understanding, confirming its effectiveness in achieving both temporally precise and semantically reliable video parsing.

\subsection{Video-MME-v2}
\label{subsec:eval-main-videomme}

Training centered on parsing may come at the expense of the general video understanding required for question answering.
We examine this on Video-MME-v2 (Tab.~\ref{tab:video-bench}) in the setting without subtitles. Our model receives $64$ visual frames; baseline scores are taken from the official leaderboard under each system's reported configuration.

At the level of the group score, general video understanding is largely preserved.
Without a thinking mode, \OurMethod performs on par with larger non-thinking models and comes within $0.1$ points of the $4$\,B thinking models.
The most comparable baseline is Qwen3-Omni-30B-A3B-Instruct, which is also an omni-modal model evaluated without thinking and has about six times as many total parameters; \OurMethod scores higher than it on the group score and on each of the three levels.

\begin{table}[!t]
\centering
\caption{\textbf{Comparison with existing specialized, general, and proprietary models on ChemOCR and PDMX-Synth.} Left: ChemOCR ($1{,}000$ molecules; $990$ scored after excluding $10$ invalid references). Right: PDMX-Synth ($800$ score images). \textbf{Bold} and \underline{underline}: best and second best in each column. $^*$ChemDFM-X~\citep{ChemDFM-X} scores on ChemOCR are taken from TinyChemVL~\citep{TinyChemVL}; LEGATO scores on PDMX-Synth are taken from its original report~\citep{LEGATO}.
}
\label{tab:symbolic}
\label{tab:chemocr-results}
\label{tab:pdmx-results}
\small
\setlength{\tabcolsep}{4pt}
\begin{minipage}[t]{0.55\textwidth}\centering
\begin{tabular}[t]{l r cc}
\toprule
 Model & Params & Avg.\ Sim.$\uparrow$ & Tani@1.0$\uparrow$ \\
\midrule
DeepSeek-OCR~2~\hfilll~\cite{DeepSeek-OCR-2} & $3$\,B & \best{74.9} & \second{49.6} \\
ChemDFM-X$^*$~\hfilll~\cite{ChemDFM-X}   & $13$\,B & 70.9 & 36.5 \\
GPT-4o~\hfilll~\cite{GPT-4o}             & --  & 36.8 & 3.4 \\
Qwen2.5-VL-7B~\hfilll~\cite{Qwen2.5-VL}  & $7$\,B  & 25.5 & 0.4 \\
InternVL2.5-4B~\hfilll~\cite{InternVL2.5} & $4$\,B & 4.4  & 0.0 \\
\midrule
\rowcolor{youtuBlue!7}\oursmark\OurMethodBF & $5$\,B & \second{74.6} & \best{54.2} \\
\bottomrule
\end{tabular}
\end{minipage}\hfill
\begin{minipage}[t]{0.43\textwidth}\centering
\begin{tabular}[t]{l ccc}
\toprule
 Model & CER$\downarrow$ & SER$\downarrow$ & LER$\downarrow$ \\
\midrule
GOT-OCR2.0~\hfilll~\cite{GOT-OCR2.0}     & 93.98 & 98.54 & 97.76 \\
Logics-Parsing-Omni~\hfilll~\cite{Logics-Parsing-Omni}               & 72.44 & 81.93 & 81.34 \\
Gemini-3-Pro~\hfilll~\cite{Gemini-3}            & {54.26} & {68.10} & {82.33} \\
GPT-5.6~\hfilll~\cite{GPT-5.6}            & {45.02} & {54.74} & {87.52} \\
LEGATO$^*$~\hfilll~\cite{LEGATO}          & \second{23.30} & \best{25.80} & \best{31.70} \\
\midrule
\rowcolor{youtuBlue!7}\oursmark\OurMethodBF & \best{22.73} & \second{26.73} & \second{33.20} \\
\bottomrule
\end{tabular}
\end{minipage}
\end{table}

\paragraph{L1 score.}
The level-wise results indicate that this outcome stems mainly from L1, i.e., information aggregation across frames.
For the baselines, L1 varies little with either model size or inference mode: it lies within a $2.3$-point range from InternVL3.5-4B-Instruct to Qwen3-Omni-30B-A3B-Think, and enabling thinking in Qwen3-Omni raises it by only $0.8$ points.
\OurMethod exceeds every non-thinking baseline on L1 by at least $5$ points, a margin that neither additional parameters nor thinking yields among the compared models.
We attribute this margin primarily to the parsing objective, which requires the model to locate and transcribe the visual content of each frame exhaustively and thus directly exercises cross-frame aggregation.

\paragraph{L2 and L3 scores.}
On L2 and L3, which target temporal understanding and complex reasoning, \OurMethod performs comparably to the non-thinking baselines and shows no comparable advantage.
Its gap to Qwen3-Omni-30B-A3B-Think on these levels ($1.9$ and $1.8$ points) is close to the improvement that thinking brings to Qwen3-Omni itself ($2.1$ points on each), whereas the same change contributes less than one point on L1.
This suggests that the remaining deficit arises from multi-step reasoning over the extracted evidence rather than from perception, and that an explicit reasoning mode, which yields a gain of similar size for Qwen3-Omni, may help close it.
It may also explain why the large lead on L1 yields only a modest lead in group score, since the group-based metric requires consistent answers across related questions, many of which involve higher-level reasoning.

In short, parsing-centric training preserves general video understanding and specifically improves cross-frame information aggregation, while higher-level temporal reasoning remains limited in the absence of a thinking mode.

\subsection{ChemOCR and PDMX-Synth}
\label{subsec:eval-main-symbolic}

ChemOCR and PDMX-Synth test whether a general parser can emit an exact formal language (SMILES and ABC notation) rather than a plausible description, and the two benchmarks separate the systems in different ways (Tab.~\ref{tab:symbolic}).

\paragraph{ChemOCR.}
On chemical structures, \OurMethod's advantage is in exact RDK-fingerprint agreement rather than average fingerprint similarity.
On ChemOCR, it attains the highest exact fingerprint-match rate (Tani@1.0) among the compared models, $54.2$ against $49.6$ for DeepSeek-OCR~2, while its average Tanimoto similarity (Avg.\ Sim.) is marginally lower ($74.6$ against $74.9$). 
A higher average fingerprint similarity does not necessarily imply that a greater number of predictions share the same fingerprint. In common scenarios, the latter is considered to be more practically meaningful, so we regard Tani@1.0 as the main metric.

\paragraph{PDMX-Synth.}
On typeset sheet music, \OurMethod attains the lowest character error rate of all compared systems, $22.73$ CER, edging out the specialized LEGATO~\citep{LEGATO} ($23.30$) and remaining second to it at the symbol and line levels ($26.73$ against $25.80$ SER, $33.20$ against $31.70$ LER).
The margin over the other general systems is of a different order: the strongest of them reach $45$--$54$ CER (GPT-5.6 and Gemini-3-Pro), and the remaining parsers lie at $72$--$94$ CER.
Taken together, \OurMethod leads every compared model on ChemOCR's Tani@1.0 metric and matches a dedicated optical-music model in character-level transcription of typeset sheet music, far ahead of the other general systems.

\subsection{Ablation Study}
\label{subsec:eval-analysis}
\label{subsec:eval-ablation}

\begin{figure}[!t]
\centering
\includegraphics[width=\textwidth]{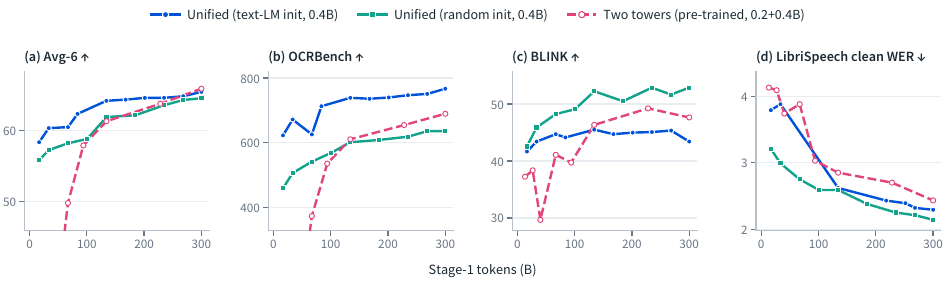}
\caption{
    \textbf{Encoder architecture and initialization studies.}
    (a) General multimodal (Avg-6); (b) OCRBench; (c) BLINK; (d) LibriSpeech test-clean WER.
    Blue, green, and pink denote the text-initialized unified encoder, randomly initialized unified encoder, and pre-trained SigLIP-2~\citep{SigLIP-2} and SenseVoice-small~\citep{FunAudioLLM} towers, respectively.
}
\label{fig:controlled}
\end{figure}

Having established what the checkpoint achieves, we now examine how the two ideas behind the model contribute: the text-initialized unified encoder and \OSAD.
We study the two principles of the encoder (Sec.~\ref{sec:model}) in a setting that uses general multimodal data and no parsing data.
This setting pairs a Qwen3-4B decoder with an encoder of the Qwen3-0.6B architecture ($0.4$\,B without its embedding layer and LM head).
General multimodal performance is summarized by Avg-6, the mean of MMBench~\citep{MMBench} (English and Chinese averaged), MMStar~\citep{MMStar}, MMMU-VAL~\citep{MMMU}, HallusionBench~\citep{HallusionBench}, AI2D~\citep{AI2D}, and OCRBench~\citep{OCRBench} (scaled by $1/10$).
Fig.~\ref{fig:controlled} summarizes the two encoder studies, and App.~\ref{sec:app-ablations} provides additional vision and ASR curves for these runs.

\paragraph{Unified encoder versus separate towers.}
\label{subsec:eval-ablation-encarch}
With two thirds of the perception parameters and no second tower, a randomly initialized unified encoder trails pre-trained SigLIP-2~\citep{SigLIP-2} and SenseVoice-small~\citep{FunAudioLLM} towers by $1.3$ points on Avg-6 at $300$\,B Stage~1 tokens (Fig.~\ref{fig:controlled}a).
At the same budget, it trails on OCRBench by $53.4$ points (Fig.~\ref{fig:controlled}b) but achieves a lower word error rate (WER) on LibriSpeech test-clean, $2.14$ against $2.43$ (Fig.~\ref{fig:controlled}d).
Text-LM initialization substantially reduces the vision gap: at the same endpoint, the text-initialized encoder comes within $0.4$ points of the towers on Avg-6 and surpasses them by $77.2$ points on OCRBench.
These results suggest that initialization accounts for an important part of the vision gap.

\paragraph{Effect of text-LM initialization.}
\label{subsec:eval-ablation-encinit}
At matched training budgets, initializing the shared block from Qwen3-0.6B-Base rather than from scratch gives the encoder \emph{a reading advantage that persists throughout the plotted range} (Fig.~\ref{fig:controlled}b).
The two vision branches run identical Stage~0 and Stage~1 schedules and differ only in this initialization.
The initialized encoder leads on OCRBench at every checkpoint evaluated in both branches; at $300$\,B tokens, its advantage is $130.6$ points ($767.0$ against $636.4$).
It reaches $672$ early in training, a score the scratch encoder never attains within the plotted range.
On Avg-6, the benefit is mainly one of sample efficiency: the initialized encoder reaches $64.1$ with half the Stage~1 tokens needed by the scratch encoder to reach a comparable score, and it leads by $0.9$ points at $300$\,B tokens ($65.37$ against $64.48$; Fig.~\ref{fig:controlled}a).
The advantage is concentrated on reading: at the same $300$\,B budget, the scratch encoder leads on $8$ of the $16$ vision benchmarks, most clearly on BLINK ($+9.5$, $52.89$ against $43.41$; Fig.~\ref{fig:controlled}c), which tests spatial and relational perception.
This dissociation is consistent with the second principle of the encoder (Sec.~\ref{sec:model}): weights that already map context to symbols help perception output symbols, not reason about space.
Because parsing depends on reading more than on any other capability, this is the encoder result most relevant to our task.

\paragraph{Contribution of \OSAD.}
\label{subsec:eval-posttrain}

\begin{figure}[!t]
\centering
\includegraphics[width=\textwidth]{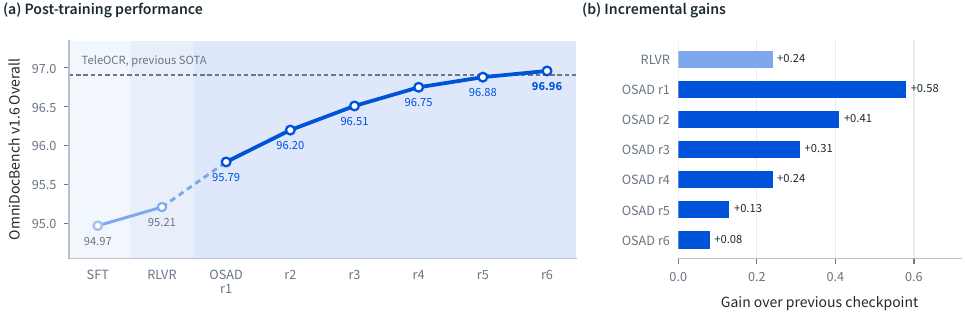}
\caption{\textbf{\OSAD supplies $1.75$ of the $1.99$ points gained after SFT ($88\%$), with positive but diminishing increments.} \OmniDocBench v1.6 Overall (higher is better). (a) Score after SFT, RLVR, and each \OSAD round; the dashed line marks TeleOCR, the strongest other system ($96.91$). (b) Gain over the preceding checkpoint.}

\label{fig:osad-progress}
\end{figure}

\OSAD provides \emph{most of the improvement beyond SFT}, contributing $1.75$ of the $1.99$ points gained on \OmniDocBench v1.6 ($88\%$), compared with $0.24$ from RLVR (Fig.~\ref{fig:osad-progress}).
Its six rounds raise the Overall score from $95.21$ to $96.96$ (Fig.~\ref{fig:osad-progress}), with gains at every round.
This continued refinement is consistent with the self-evolving design, which refreshes supervision as the parser learns. The refinement targets \emph{content accuracy within structured outputs}: a schema-valid parse can still misread words, formula symbols, or table-cell values.
\OSAD provides reference-conditioned feedback at these content positions while separately supervising JSON keys and delimiters.
For pages combining text, formulas, and tables, this couples precise transcription with consistent serialization, providing direct supervision for local errors that sequence-level rewards capture only indirectly.

\FloatBarrier

\section{Related Work}
\label{sec:related}

\subsection{Parsing as a Perception Interface for Agents}
\label{subsec:related-interface}

Retrieval-augmented generation~\citep{RAG} and tool-using agents~\citep{ReAct} motivate reusable interfaces between external evidence and language-model reasoning.
For multimodal inputs, this interface may consist of extracted content, visual embeddings, or grounded structured records.
ColPali~\citep{ColPali} and visual retrieval systems such as VisRAG and M3DocRAG~\citep{VisRAG,M3DocRAG} retrieve directly from document images, while Pix2Struct~\citep{Pix2Struct} uses screenshot parsing to learn representations for visual language understanding.
MMLongBench-Doc~\citep{MMLongBench-Doc} evaluates understanding across long multimodal documents, and OSWorld~\citep{OSWorld} and TheAgentCompany~\citep{TheAgentCompany} place perception within interactive computer and workplace tasks.
These settings motivate interfaces that retain both semantic content and its source location.
\OmniSchema serves this purpose through a common representation of entities, relations, and spatial or temporal anchors across documents, images, audio, and video.

\subsection{Omni-Modal Parsing and Perception Encoders}
\label{subsec:related-parsing}

\paragraph{Document parsing.}
Document understanding spans OCR-based pipelines and models that generate structured outputs directly from images.
Donut~\citep{Donut} established an OCR-free encoder-decoder approach, while systems such as MinerU and PaddleOCR~\citep{MinerU,PaddleOCR-3.0} organize recognition and layout analysis into specialized components.
Recent parsers explore different allocations of these computations: MinerU2.5~\citep{MinerU2.5} separates low-resolution layout analysis from high-resolution recognition; dots.ocr and MonkeyOCR~\citep{dots.ocr,MonkeyOCR} integrate layout and content prediction; and \OurMethodV1~\citep{Youtu-Parsing} decodes regions in parallel from shared encoder features.
MinerU2.5-Pro and PaddleOCR-VL~\citep{MinerU2.5-Pro,PaddleOCR-VL,PaddleOCR-VL-1.6} further improve document parsing through data and training refinements.
GOT-OCR2.0 and DeepSeek-OCR~\citep{GOT-OCR2.0,DeepSeek-OCR} also emphasize visual encoding tailored to text recognition and document compression.
Our work extends this encoder-centric approach to grounded parsing across visual and auditory inputs.

\paragraph{Omni-modal models and structured parsing.}
Omni-modal language models integrate vision and audio for multimodal understanding and generation~\citep{Qwen2.5-Omni,Qwen3-Omni,Baichuan-Omni-1.5,MiniCPM-o-4.5,VITA-1.5,M2-omni,Ming-Omni}.
A common architecture uses separate vision and audio encoders, followed by fusion in the language model.
Qwen2.5-Omni, for example, uses TMRoPE to align audio and video tokens through temporal positional encoding~\citep{Qwen2.5-Omni}.
\Logics~\citep{Logics-Parsing-Omni} develops an explicit omni-modal parsing framework that connects detection, recognition, and interpretation through evidence grounding, and introduces \OmniParsingBench, which we adopt.
Building on this task formulation, \OurMethod combines a shared output schema with a unified perception encoder, \OurEncoder, placing audio-visual temporal alignment and selective cross-modal interaction before the decoder.

\paragraph{Shared representations and unified encoders.}
Multimodal alignment and encoder sharing address complementary aspects of perception.
ImageBind~\citep{ImageBind} aligns six modalities in a common embedding space using image-paired data.
Architectures such as VATT~\citep{VATT}, Omnivore~\citep{Omnivore}, and TVLT~\citep{TVLT} instead explore shared Transformer computation across different combinations of visual, auditory, textual, and depth inputs.
Meta-Transformer~\citep{Meta-Transformer} and OneLLM~\citep{OneLLM} extend parameter sharing across modalities with modality-specific input processing.
\OurEncoder targets grounded parsing, including native-resolution pages, with a text-initialized shared backbone and a common temporal positional encoding for audio and video.
Cross-modal attention is confined to selected layers and local windows, combining shared computation with explicit temporal interaction.

\paragraph{Language-model initialization for perception.}
Pre-trained language-model weights provide a complementary starting point to visual or contrastive pre-training.
FPT~\citep{FPT} studies transfer from text Transformers to non-text tasks, and frozen language-model blocks can enrich visual representations~\citep{FrozenLLMVision}.
TrOCR~\citep{TrOCR} uses pre-trained text components for recognition, while encoder-free models such as Fuyu, EVE, and Mono-InternVL~\citep{Fuyu,EVE,Mono-InternVL} process image patches within the language-model architecture.
DeepSeek-OCR~2~\citep{DeepSeek-OCR-2} is particularly relevant: its visual encoder incorporates a Qwen2-initialized Transformer with bidirectional visual attention and causal query tokens for learned reading order.
We apply text initialization to a separate perception backbone shared by vision and audio, preserving a distinct decoder and evaluating the initialization against an otherwise identical encoder trained from scratch (Subsec.~\ref{subsec:eval-ablation-encinit}).

\subsection{Post-Training for Structured Parsing}
\label{subsec:related-posttrain}

\paragraph{Verifiable rewards.}
Reinforcement learning extends parsing supervision from reference sequences to model-generated outputs.
GRPO~\citep{DeepSeekMath} provides a group-relative optimization framework, and OCR systems instantiate task-specific rewards: olmOCR-2~\citep{olmOCR-2} uses programmatic unit tests, while MinerU2.5-Pro and PaddleOCR-VL-1.6 incorporate reinforcement learning into their training pipelines~\citep{MinerU2.5-Pro,PaddleOCR-VL-1.6}.
Our schema-routed RLVR combines shared validity checks with task-specific content scores under one policy.
The subsequent distillation stage supplies token-level feedback for structured outputs whose heterogeneous fields are difficult to supervise with a single scalar reward.

\paragraph{On-policy and reference-conditioned distillation.}
Training on learner-generated trajectories addresses the mismatch between reference prefixes and the states encountered at inference~\citep{DAgger,ScheduledSampling}.
GKD~\citep{GKD} applies teacher feedback to student-generated sequences and permits different distribution-matching objectives; MiniLLM~\citep{MiniLLM} develops on-policy optimization of a reverse-KL distillation objective.
A separate line of work transfers information available only during training: learning with privileged information and generalized distillation formalize this setting~\citep{LUPI,GeneralizedDistillation}, while context distillation internalizes the benefits of additional instructions or reasoning context~\citep{ContextDistillation}.
OPSD~\citep{OPSD} brings these ideas together by using the same model as teacher and student under different contexts, with reference-conditioned token targets evaluated on student rollouts.
\OSAD builds directly on this construction and uses the output schema to adapt supervision to token roles, assigning different divergences to content and structure within a shared all-token objective.

\paragraph{Self-distillation and teacher evolution.}
Born-again networks~\citep{BornAgain} and Noisy Student~\citep{NoisyStudent} improve models through successive teacher-student generations, while Mean Teacher and DINO~\citep{MeanTeachers,DINO} update a teacher through an exponential moving average of student parameters.
These approaches establish teacher evolution as a general mechanism for refining supervision.
\OSAD combines periodic teacher refresh with reference-conditioned on-policy targets: the teacher remains frozen within each round, then inherits the trained student's checkpoint for the next round.
This design couples schema-aware token supervision with a supervision source that incorporates the student's accumulated learning, as described in Subsec.~\ref{subsec:osad}.

\section{Limitations and Conclusion}
\label{sec:conclusion}

\paragraph{Limitations.}
\label{subsec:conclusion-limitations}
Compared with its predecessor \OurMethodV1~\citep{Youtu-Parsing}, \OurMethod does not yet implement inference acceleration.
\OurMethodV1 reuses shared encoder features across the regions of a page and applies token- and query-level parallel decoding, whereas \OurMethod decodes end-to-end and autoregressively with a larger decoder.
We therefore expect its throughput on dense structured content such as tables to be lower.
We will bring parallel decoding and related acceleration techniques to \OurMethod in future work.

\paragraph{Conclusion.}
\label{subsec:conclusion-summary}

We presented \OurMethod, an omni-modal parser that converts documents, images, audio, and video into structured records grounded in space and time.
Its design rests on \emph{one encoder for every modality and one schema for every task}.
The encoder, \OurEncoder, replaces separate vision and audio towers with one text-initialized Transformer, in which visual and acoustic tokens share a space-time positional encoding and interact within short audio-visual windows.
The schema, \OmniSchema, casts the outputs of all ten tasks into one grounded format, which allows a single self-evolving post-training procedure, \OSAD, to use the same token-role-based objective across tasks.
Together, the two components let a single $5$\,B-parameter checkpoint serve all ten parsing tasks with one inference configuration.
\OurMethod achieves the highest \OmniDocBench v1.6 score among all compared systems ($96.96$) and leads all open-weight models on \OmniParsingBench ($75.08$).

\section*{Contributions}
\label{sec:contributions}

\paragraph{Core Contributors}
Yunhang Shen\textsuperscript{\rm \dag}\quad
Zhangquan Chen\textsuperscript{\rm \dag}\quad
Xiaotian Li\textsuperscript{\rm \dag}\quad
Zhongpeng Cai\textsuperscript{\rm \dag}\quad
Qing Zhou\quad
Zekai Zhang\quad
Kun Zhang\quad
Zezhong Wu\quad
Jian Hu\quad
Haihan Gao\quad
Heting Gao\quad
Xin Li\quad
Yinsong Liu\quad
Deqian Jiang\quad
Di Yin\quad
Haoyu Cao\textsuperscript{\rm \#}\quad
Xing Sun\textsuperscript{\rm *}

\paragraph{Contributors}
Bing Liu\quad
Yanqiu Qu\quad
Yunfei Wu\quad
Kun Yin\quad
Zhixiang Wei\quad
Yongxiang Hua\quad
Qianyu Li\quad
Xiaoxing Hu\quad
Peiyuan Zhang\quad
Lijiang Li\quad
Dong Xu\quad
Zhenchuan Yang\quad
Yiliang Jiang\quad

\medskip
\textsuperscript{\rm \dag}Equal contribution\quad
\textsuperscript{\rm \#}Project lead\quad
\textsuperscript{\rm *}Corresponding author


\clearpage

\setcitestyle{numbers,square}
\setlength{\bibsep}{0.4pt plus 0.4pt}
\bibliography{library_format}

\clearpage
\appendix

\renewcommand{\topfraction}{0.95}
\renewcommand{\bottomfraction}{0.6}
\renewcommand{\textfraction}{0.05}
\renewcommand{\floatpagefraction}{0.5}
\setcounter{topnumber}{5}
\setcounter{bottomnumber}{3}
\setcounter{totalnumber}{8}
\makeatletter
\setlength{\@fptop}{0pt}
\setlength{\@fpsep}{14pt plus 2pt}
\setlength{\@fpbot}{0pt plus 1fil}
\makeatother

\section*{Appendix}
\addcontentsline{toc}{section}{Appendix}
\label{sec:app-overview}

The main text is self-contained; the appendix gives the complete specification behind it.

\begin{center}
\small
\renewcommand{\arraystretch}{1.15}
\begin{tabular}{@{}l >{\raggedright\arraybackslash}p{10.2cm} r@{}}
\toprule
\tabhead & Contents & Page \\
\midrule
\ref{sec:app-data}      & Data: output fields per task and the complete annotation pipelines & \pageref{sec:app-data} \\
\ref{sec:app-arch}      & Architecture: complete configuration and component details, including variants the final model does not use & \pageref{sec:app-arch} \\
\ref{sec:app-posttrain} & Post-training: reward implementation, rollout sampling, and the \OSAD round schedule & \pageref{sec:app-posttrain} \\
\ref{sec:app-eval}      & Evaluation protocol: metric definitions and the evaluation set & \pageref{sec:app-eval} \\
\ref{sec:app-ablations} & Additional encoder ablation curves & \pageref{sec:app-ablations} \\
\ref{sec:app-examples}  & Qualitative examples & \pageref{sec:app-examples} \\
\bottomrule
\end{tabular}
\end{center}

\FloatBarrier
\section{Data}
\label{sec:app-data}

\begin{appintro}
Tab.~\ref{tab:data-document-fields}--\ref{tab:data-video-fields} list the annotation fields for document, natural image, audio, and video parsing, respectively.
For general-purpose data, we reuse the training data of Youtu-LLM~\citep{Youtu-LLM}, Youtu-VL~\citep{Youtu-VL}, Long-VITA~\citep{Long-VITA}, VITA-1.5~\citep{VITA-1.5}, and VITA-Audio~\citep{VITA-Audio}.
\end{appintro}
\label{sec:app-data-common}

\subsection{Document Parsing}
\label{subsec:app-data-doc}

\begin{table}[H]
  \centering
  \small
  \caption{\textbf{Annotation fields of the document parsing dataset and their semantics.}}
  \label{tab:data-document-fields}
  \setlength{\tabcolsep}{3pt}
  \begin{tabular}{@{}llp{8.3cm}@{}}
  \toprule
  \textbf{Field} & \textbf{Scope} & \textbf{Semantics} \\
  \midrule
  \texttt{id} & element & Unique element identifier. \\
  \texttt{category} & element & Element type: text, table, formula, code, caption, header, footer, title, chemical, score, chart, flowchart, figure, seal, or geometric. \\
  \texttt{bbox} & element & Bounding box of the element in page coordinates. \\
  \texttt{content.text} & element & Transcription for text-like elements, code, chemical structures, and ABC notation. \\
  \texttt{content.latex} & formula & Formula transcription in \LaTeX{}. \\
  \texttt{content.otsl} & table & Table structure and content in OTSL. \\
  \texttt{content.markdown} & chart & Structured chart data in Markdown. \\
  \texttt{content.mermaid} & flowchart & Nodes and directed relations in Mermaid. \\
  \texttt{content.geometric} & geometric & Geometric primitives and relations. \\
  \texttt{content.caption} & visual element & Description of a chart, flowchart, figure, or geometric diagram. \\
  \texttt{attributes.language} & element & Language of the element content. \\
  \texttt{reading\_order} & page & Ordered list of element identifiers. \\
  \texttt{global\_description} & page & Page-level caption, narrative, language, and semantic tag. \\
  \bottomrule
  \end{tabular}
\end{table}

\label{subsec:data-doc-general}\label{subsec:data-doc-vertical}
The document data consist of two parts: full pages for layout and content parsing, and dedicated data for specialized elements.
For full pages, public document datasets are typically annotated for a single sub-task and contain label noise, so we re-annotate them with expert models and vision-language models (VLMs).
Pages that pass consistency checks and a vote among annotators with different architectures are retained directly; of the pages on which the annotators disagree, about $10\%$ are corrected by human experts and added back to the corpus.
Synthetic pages are rendered from layout templates of magazines, papers, textbooks, manuals, and advertisements, so their labels are exact by construction.
For specialized elements, source tables of public chart datasets~\citep{ChartQA,PlotQA,Chart-to-Text} are serialized to Markdown, and flowcharts, mind maps, and architecture diagrams generated from sampled graphs are serialized to Mermaid.
Molecules from PubChem~\citep{PubChem} and ChEMBL~\citep{ChEMBL} are rendered into $1.0$\,M pairs of images and SMILES~\citep{SMILES} strings, geometric figures are produced by a procedural generator with primitives and relations aligned to pixels by construction, and music scores from PDMX~\citep{PDMX} are converted to ABC and rendered with varied engines, fonts, and layouts.

\subsection{Natural Image Parsing}
\label{subsec:app-data-image}
\label{subsec:data-image-pipeline}

\begin{table}[!t]
  \centering
  \small
  \caption{\textbf{Annotation fields of the natural image parsing dataset and their semantics.}}
  \label{tab:data-image-fields}
  \setlength{\tabcolsep}{3pt}
  \begin{tabular}{@{}ll>{\raggedright\arraybackslash}p{8.0cm}@{}}
  \toprule
  \textbf{Field} & \textbf{Scope} & \textbf{Semantics} \\
  \midrule
  \texttt{id} & element & Unique element identifier. \\
  \texttt{category} & element & Element type, fixed to \texttt{entity}. \\
  \texttt{bbox} & element & Bounding box of a salient object. \\
  \texttt{tag} & element & A word naming the object inside the bounding box. \\
  \texttt{content.text} & element & OCR result inside the bounding box; \texttt{""} when the annotation pipeline returns no text. \\
  \texttt{content.caption} & element & Precise and concise caption of the region inside the bounding box. \\
  \texttt{global\_description.caption} & image & Precise and concise caption of the image. \\
  \texttt{global\_description.narrative} & image & Detailed description of the image, within $100$ characters. \\
  \texttt{global\_description.scene} & image & Short phrase describing where the image takes place. \\
  \bottomrule
  \end{tabular}
\end{table}

Each image is annotated in three stages: captioning, entity verification, and localization.
We first prompt a VLM to write a short caption restricted to the subject, the scene, and the action, followed by a longer narrative conditioned on that caption and the image.
Next, a large language model (LLM) extracts the entities named in the caption and the narrative, and a second VLM checks whether each entity is visible in the image.
If any caption entity is hallucinated, the sample is discarded; hallucinated narrative entities are instead removed, the narrative is rewritten from the remaining entities, and the entity lists of the caption and the narrative are merged.
Finally, each remaining entity is localized independently by SAM-3~\citep{SAM-3} and by a VLM.
When the IoU of the two boxes exceeds a threshold, their intersection is kept; otherwise, an LLM selects the box whose region caption matches the entity.
The region captions, kept as \texttt{content.caption}, and the scene phrase \texttt{global\_description.scene} are also generated by a VLM.
The text inside each box is transcribed by \OurMethodV1~\citep{Youtu-Parsing} and by a general-purpose OCR engine; when the two transcripts disagree, the element is discarded. An empty \texttt{text} records that neither OCR system returned text, rather than guaranteeing that the region contains none.

\subsection{Audio Parsing}
\label{subsec:app-data-audio}
\label{subsec:data-audio-pipeline}

\begin{table}[!t]
  \centering
  \small
  \caption{\textbf{Annotation fields of the audio parsing dataset and their semantics.}}
  \label{tab:data-audio-fields}
  \setlength{\tabcolsep}{3pt}
  \begin{tabular}{@{}llp{8.2cm}@{}}
  \toprule
   \textbf{Field} & \textbf{Scope} & \textbf{Semantics} \\
  \midrule
  \texttt{id} & segment & Unique segment identifier. \\
  \texttt{kind} & segment & Segment type: \texttt{vocal} or \texttt{nonvocal}. \\
  \texttt{audio.asr\_text} & vocal segment & Verbatim transcription of the speech in the segment. \\
  \texttt{audio.speaker\_id} & vocal segment & Speaker identifier, consistent across the chunk. \\
  \texttt{audio.timbre\_caption} & vocal segment & Voice timbre: gender, age, pitch, breathiness, etc. \\
  \texttt{audio.instruct\_caption} & vocal segment & Speech delivery characteristics: emotion, pace, prosody, etc. \\
  \texttt{audio.scene\_caption} & vocal segment & Mid-level scene context: time, location, activity, etc. \\
  \texttt{language} & vocal segment & Language of the segment (\texttt{zh}/\texttt{en}/\texttt{unk}). \\
  \texttt{audio.env\_caption} & nonvocal segment & Description of the background acoustic environment. \\
  \texttt{audio.acoustic\_events} & segment & Closed-set acoustic-event labels (e.g., \texttt{Human Voice}, \texttt{Music}). \\
  \texttt{start/end\_timestamp} & segment & Segment boundaries relative to the chunk onset (\texttt{HH:MM:SS}). \\
  \texttt{global\_description} & entire chunk & Holistic description aggregating information across segments. \\
  \bottomrule
  \end{tabular}
\end{table}

The audio data consist of in-house audiobooks, podcasts, and conversations.
Each clip is first segmented: a BEATs~\citep{BEATs} tagger trained with PretrainedSED~\citep{PretrainedSED} labels every frame with an AudioSet~\citep{Audio-Set} class, and the resulting vocal and nonvocal regions are split into segments of at most $30$\,s and packed into chunks of about $60$\,s, each of which forms one training sample.
Speech is then attributed and transcribed: DiariZen~\citep{DiariZen} assigns clip-level speaker identities, three ASR systems with different architectures transcribe each turn, and the transcript with the smallest total pairwise WER is retained, while the largest pairwise WER is recorded as a disagreement score.
The language of each segment is identified from the transcript by fastText~\citep{fastText} and from the audio by MMS-LID~\citep{MMS-LID}, and word timestamps are obtained by MMS forced alignment.
The timbre, delivery, and environment captions (\texttt{timbre\_caption}, \texttt{instruct\_caption}, and \texttt{env\_caption}) are generated by an audio-language model, and the scene caption and the chunk-level \texttt{global\_description} by an LLM.
For fine-tuning, a chunk is retained only if the two language identifiers agree and its disagreement score does not exceed $0.1$.
For reinforcement learning, we further discard chunks whose four offline rollouts receive uniformly high, uniformly low, or identical scores from an audio-language judge; this judge is used only for this offline filtering, and online training uses the rule-based verifiable reward of Equ.~\ref{equ:srvr-reward}.

\subsection{Video Parsing}
\label{subsec:app-data-video}
\label{subsec:data-video-pipeline}

\begin{table}[!t]
  \centering
  \small
  \caption{\textbf{Annotation fields of the video parsing dataset and their semantics.}}
  \label{tab:data-video-fields}
  \setlength{\tabcolsep}{3pt}
  \begin{tabular}{@{}lll>{\raggedright\arraybackslash}p{5.6cm}@{}}
  \toprule
  \textbf{Field} & \textbf{Scope} & \textbf{Category} & \textbf{Semantics} \\
  \midrule
  \texttt{id}
  & segment & shared
  & Unique segment identifier. \\

  \texttt{subtype}
  & video & shared
  & Video type: \texttt{natural\_video} or \texttt{text\_rich\_video}. \\

  \texttt{start/end\_timestamp}
  & segment & shared
  & Temporal boundaries relative to video onset. \\

  \texttt{visual.elements}
  & segment & shared
  & Salient objects and on-screen text; each item has a \texttt{category} (\texttt{object} or \texttt{text}) and, for text, an \texttt{ocr\_type} (e.g., \texttt{title}). \\

  \texttt{visual.atomic\_actions}
  & segment & shared
  & Fine-grained actions within the segment. \\

  \texttt{visual.object\_actor\_interactions}
  & segment & shared
  & Interactions between actors and objects. \\

  \texttt{visual.camera\_motion}
  & segment & shared
  & Camera motion at two granularities (\texttt{l1}: coarse, \texttt{l2}: fine). \\

  \texttt{visual.event\_description}
  & segment & shared
  & Summary of the segment-level visual event. \\

  \texttt{audio.speaker\_id}
  & segment & shared
  & Speaker identifier, consistent across the video. \\

  \texttt{audio.asr\_text}
  & segment & shared
  & Transcription of speech in the segment. \\

  \texttt{audio.acoustic\_events}
  & segment & shared
  & Acoustic-event labels of the segment audio. \\

  \texttt{global\_description}
  & video & shared
  & Global scene, narrative, caption, and semantic tag. \\

  \texttt{structured\_report}
  & video & shared
  & Markdown report of text-centric content for text-rich video; an empty string (\texttt{""}) for natural video. \\
  \bottomrule
  \end{tabular}
\end{table}

Each video is first segmented by a VLM, at event and scene changes for natural video and at slide, page, and interface transitions for text-rich video.
The VLM then annotates the visual fields of each segment; in parallel, an ASR system transcribes the corresponding audio, a speaker diarization model assigns speaker IDs that remain consistent across the entire video, and an acoustic event tagger labels its acoustic events; the visual and audio annotations are aligned only through the segment timestamps.
Finally, an LLM summarizes the segments into the \texttt{global\_description} and, for text-rich video, into the \texttt{structured\_report}.

\section{Architecture Details}
\label{sec:app-arch}

\begin{appintro}
This appendix lists every architectural setting (Tab.~\ref{tab:arch-config-appx}) and specifies the components of \OurEncoder that go beyond a standard Transformer, together with two rotary variants that the final model does not use.
\end{appintro}

\subsection{Model Specification}
\label{subsec:app-hparams}

Tab.~\ref{tab:arch-config-appx} gives the complete configuration of the encoder, the stems, the projector, the decoder, and the sequence settings.

\begin{table}[!t]
\centering
\small
\caption{\textbf{Complete configuration of \OurMethod.}}
\label{tab:arch-config-appx}
\begin{tabular}{lll}
\toprule
 Component & Quantity & Value \\
\midrule
\multirow{8}{*}{\OurEncoder}
 & layers $L$                        & $28$ \\
 & hidden size $d$                   & $1{,}024$ \\
 & FFN hidden size $d_{\mathrm{ffn}}$& $3{,}072$ \\
 & query heads $H$ / KV groups $H_{kv}$ & $16$ / $8$ \\
 & head dimension $d_h$              & $128$ \\
 & attention                         & GQA, bidirectional, packed \\
 & normalization                     & RMSNorm, $\varepsilon=10^{-6}$ \\
 & activation                        & SwiGLU \\
\midrule
\multirow{6}{*}{Encoder RoPE}
 & rotary slots $d_{\mathrm{rot}}$   & $64$ \\
 & axis set $\mathcal{A}$            & $\{T,H,W\}$; chunked segments \\
 & axis split $(d_T,d_H,d_W)$        & $(21,21,22)$ \\
 & base $b$                          & $10^{4}$ \\
\midrule
\multirow{5}{*}{Stems}
 & vision patch size $p$             & $16$ \\
 & mel bins $F$ / frame rate $f_{\mathrm{mel}}$ & $128$ / $100$\,Hz ($25$\,ms window, $10$\,ms hop) \\
 & audio convolutions                & $3\times$ stride-2, $C_a=512$ channels \\
 & audio window $n_w$                & $100$ mel frames ($1$\,s) \\
 & audio token rate $r_a$            & $13$ per second ($\lceil 100/8\rceil$ per window) \\
\midrule
\multirow{4}{*}{Projector}
 & merge factors $(m_v,m_a)$         & $(2,2)$ \\
 & merge normalization               & RMSNorm over $k_\mu d$ \\
 & projector hidden $d_{\mathrm{proj}}$ & $2{,}560$ \\
 & projector output $d_{\mathrm{lm}}$ & $2{,}560$ \\
\midrule
\multirow{9}{*}{LLM decoder}
 & layers $L_{\mathrm{lm}}$          & $40$, all dense \\
 & hidden size $d_{\mathrm{lm}}$     & $2{,}560$ \\
 & FFN hidden size                   & $9{,}728$ \\
 & attention                         & MLA, causal, packed \\
 & heads $H_{\mathrm{lm}}$           & $32$ \\
 & $(r_q,r_{kv})$                    & $(1536,512)$ \\
 & $(d^{C},d^{R},d^{V})$             & $(128,64,128)$ \\
 & normalization / activation        & RMSNorm, $\varepsilon=10^{-6}$ / SwiGLU \\
 & RoPE                              & 1D, base $1.6\times10^{6}$, no scaling \\
\midrule
\multirow{3}{*}{Sequence}
 & context length                    & $128$\,k tokens \\
 & video fusion layers               & $13$, $14$, $27$, $28$ of $28$ (1-indexed) \\
\bottomrule
\end{tabular}
\end{table}

\paragraph{Encoder paradigms of representative parsers.}
Tab.~\ref{tab:encoder-paradigm-comparison} compares the perception front ends of representative parsers: every other parser in the table uses a separate encoder per modality, whereas \OurMethod covers all three modalities with one encoder.

\begin{table}[!t]
\centering
\caption{\textbf{Every other representative parser uses a separate encoder per modality, and only Logics-Parsing-Omni covers all three modalities, with two towers; \OurMethod covers them with one encoder.}
\emph{Modality-specific}: every modality has its own encoder; \emph{Unified}: one encoder with shared weights for all modalities.
\emph{End-to-end}: a page or clip is parsed in one forward pass rather than by a layout-then-recognition pipeline.
}
\label{tab:encoder-paradigm-comparison}
\small
\setlength{\tabcolsep}{3.5pt}
\resizebox{\textwidth}{!}{%
\begin{tabular}{llccl ccccc c}
\toprule
 & & \multicolumn{2}{c}{Perception} & & \multicolumn{5}{c}{Parsing capability} & \\
\cmidrule(lr){3-4} \cmidrule(lr){6-10}
 \multirow{-2}{*}{Model} & \multirow{-2}{*}{\shortstack{Encoder\\paradigm}} & Vision & Audio & \multirow{-2}{*}{Language model} & Doc. & Image & Audio & Video & End-to-end & \multirow{-2}{*}{Params} \\
\midrule
MinerU2.5-Pro~\hfilll~\cite{MinerU2.5-Pro}             & Modality-specific & ViT $0.6$\,B & -- & Qwen2-0.5B      & \yes & \no & \no & \no & \no & $1.2$\,B \\
PaddleOCR-VL-1.6~\hfilll~\cite{PaddleOCR-VL-1.6}       & Modality-specific & ViT $0.4$\,B & -- & ERNIE-4.5-0.3B  & \yes & \no & \no & \no & \no & $0.9$\,B \\
TeleOCR~\hfilll~\cite{TeleOCR}                         & Modality-specific & ViT $0.6$\,B & -- & Qwen3-0.6B      & \yes & \no & \no & \no & \no & $1.2$\,B \\
DeepSeek-OCR~2~\hfilll~\cite{DeepSeek-OCR-2}           & Modality-specific & LLM $0.5$\,B & -- & DeepSeek-3B-MoE & \yes & \no & \no & \no & \yes & $3$\,B \\
\OurMethodV1~\hfilll~\cite{Youtu-Parsing}             & Modality-specific & ViT $0.4$\,B & -- & Youtu-LLM-2B    & \yes & \no & \no & \no & \no & $2.5$\,B \\
Ovis2.6-30B-A3B~\hfilll~\cite{Ovis2.5}                         & Modality-specific & ViT $0.4$\,B & -- & Qwen3-30B-A3B   & \yes & \yes & \no & \yes & \yes & $30$\,B-A$3$\,B \\
Logics-Parsing-Omni~\hfilll~\cite{Logics-Parsing-Omni} & Modality-specific & ViT $0.5$\,B & AuT $0.6$\,B & Qwen3-30B-A3B & \yes & \yes & \yes & \yes & \yes & $30$\,B-A$3$\,B \\
\midrule
\rowcolor{youtuBlue!7}\oursmark\OurMethodBF & Unified & \multicolumn{2}{c}{\OurEncoder $0.4$\,B} & Youtu-LLM-4.5B & \yes & \yes & \yes & \yes & \yes & $5$\,B \\
\bottomrule
\end{tabular}%
}
\end{table}

\subsection{\OurEncoder}
\label{subsec:app-omni-encoder}

\subsubsection{Modality Stems}
\label{subsec:app-omni-stems}

Both stems are thin: they only turn raw signals into tokens of the shared-block width $d$ and together hold less than $3\%$ of the encoder parameters.
The \emph{vision stem} is a single linear map from each $16\times16$ patch, taken at native resolution under a token budget~\citep{Qwen2-VL}, to one token, with no convolution, absolute position table or class token, so all spatial information enters through RoPE (Subsec.~\ref{subsec:omni-rope}).
The \emph{audio stem} applies three $3\times3$, stride-2 convolutions with GELU to the log-mel spectrogram, so each $1$\,s window of $100$ mel frames becomes $\lceil 100/8\rceil=13$ tokens, i.e., $13$ audio tokens per second.
It then folds the remaining frequency bins into the channels and projects them to width $d$ with a bias-free linear map, so that audio tokens have exactly the layout the shared block expects.

\subsubsection{Shared Transformer Block}
\label{subsec:app-omni-block}

The shared block is a standard bias-free pre-norm Transformer (RMSNorm~\citep{RMSNorm}, SwiGLU FFN~\citep{SwiGLU}, grouped-query attention~\citep{grouped-query-attention} with QK-Norm~\citep{QKNorm}) without a final normalization layer, since the merger supplies its own.
It departs from a language-model decoder in two ways: attention is bidirectional, and rotary embeddings are read from a per-token angle matrix, which makes the axis-factorized scheme of Subsec.~\ref{subsec:omni-rope} expressible and lets the same token be rotated differently in different layers on the video path (Subsec.~\ref{subsec:omni-video}).

\subsubsection{Positional Encoding}
\label{subsec:app-omni-rope}

Sharing one block across modalities makes positional encoding the central design question: a single frequency budget of $d_{\mathrm{rot}}=d_h/2$ rotary slots, with $d_h$ the per-head dimension of the shared block, has to express 2D image geometry, 1D audio time and, for fused video, the correspondence between the two.

A plain 2D-vision / 1D-audio RoPE~\citep{RoFormer} would place video frames and their audio under incomparable positional encodings, with no frame index and with audio time occupying the slots that vision uses for height.

\paragraph{Chunked axis-factorized RoPE.}

We instead partition the rotary budget across a set of semantic axes $\mathcal{A}$: each axis $A\in\mathcal{A}$ receives a contiguous segment of $d_A$ slots with its own inverse-frequency table, and the angle row $\boldsymbol\Phi[s]$ of token $s$ concatenates the per-axis contributions:
\begin{equation}
\label{equ:chunked}
\boldsymbol\Phi[s]=
\big\Vert_{A\in\mathcal{A}}\ x^{(A)}_s\,\boldsymbol\phi^{(A)},
\qquad
\phi^{(A)}_k=b_A^{-\,k/d_A},\ \ k=0,\dots,d_A-1,
\qquad
\sum_{A\in\mathcal{A}} d_A = d_{\mathrm{rot}},
\end{equation}
where $x^{(A)}_s$ is the position index of token $s$ along axis $A$ and $b_A$ its rotary base.
Because each segment is normalized by its own width $d_A$, every axis spans the full frequency range regardless of how many slots it owns.
\OurMethod uses $\mathcal{A}=\{T,H,W\}$ with a common base and $(d_T,d_H,d_W)=(21,21,22)$ at $d_{\mathrm{rot}}=64$ (Fig.~\ref{fig:rope-axes}); vision tokens take $t$ from the frame index and $(h,w)$ from the patch grid, while audio tokens take $t$ from the time offset and set $h=w=0$, so both modalities share one time axis (Subsec.~\ref{subsec:omni-video}).

\paragraph{Unused variants.}
The formulation admits two variants that the final model does not use.
First, an explicit \emph{modality} axis $M$ with a categorical positional encoding (image, audio, video frame, video audio) gave no gain with $d_M=4$ in preliminary runs, since the two stems already produce readily separable features.
Second, an \emph{interleaved} layout, in the style of the interleaved M-RoPE of Qwen3-VL~\citep{Qwen3-VL}, lets $T$, $H$, and $W$ share one full-length frequency table through a stride-3 assignment, so that each axis covers the whole spectrum without per-axis renormalization.
Because the per-axis normalization of Equ.~\ref{equ:chunked} already gives every axis the full frequency range, the two layouts differ only in where each axis samples that range; the interleaved variant is a drop-in replacement, but preliminary runs favored the chunked layout for videos of up to $64$ frames. Stage~3 training uses up to $512$ frames, while inference uses $64$ frames.

\subsubsection{Token Merger and Projector}
\label{subsec:app-omni-merge}

Down-sampling acts \emph{after} the shared block, so attention runs at full resolution while the decoder receives fewer tokens.
Vision tokens are merged by concatenating each $2\times2$ block within a frame along the feature axis ($4\times$ fewer tokens, e.g., $196$ decoder tokens for a $448\times448$ image), which is nearly free because adjacent patches are highly redundant.
Each modality then applies RMSNorm to the merged features, followed by a bias-free two-layer GELU MLP that maps them to the decoder width, so they enter the decoder as ordinary input embeddings.

\subsection{Language Model}
\label{subsec:app-arch-llm}
\label{subsec:arch-llm}

The encoder's tokens are consumed by a decoder that we pre-train for text.
It is a dense $40$-layer Transformer of width $2{,}560$ with multi-head latent attention (MLA)~\citep{DeepSeek-V2}, which compresses the key-value cache for long outputs, and with 1D rotary positions of base $1.6\times10^{6}$ and a $128$\,k context.
Its pre-training follows the recipe of Youtu-LLM-2B~\citep{Youtu-LLM}, scaled to $4.5$\,B parameters.
The decoder retains 1D positions because the encoder already encodes spatial and temporal structure; this lets us reuse the text checkpoint and its long-context recipe unchanged.

\section{Post-Training Details}
\label{sec:app-posttrain}

\begin{appintro}
This appendix supplements Subsec.~\ref{subsec:posttrain} with reward implementation, rollout sampling, distributed loss reduction, and the six-round training schedule.
\end{appintro}

\subsection{Schema-Routed RLVR}
\label{subsec:app-srvr}

The reward in Equ.~\ref{equ:srvr-reward} combines two factors: shared well-formedness $\Omega$ and schema-specific content accuracy $\mathrm{core}_{\nu}$.
We describe their components below and optimize the resulting reward with GRPO~\citep{DeepSeekMath}, using group-mean-centered rewards as advantages.

\paragraph{Well-formedness ($\Omega$).}
$\Omega$ combines three components:
\emph{(i) format validity} ($\mathrm{Fmt}$), which averages checks for decoding after repair, decoding without repair, correctly typed required fields, and absence of undeclared fields;
\emph{(ii) repetition} ($P_{\mathrm{repeat}}$), which penalizes excess top-level units, repeated field-scoped $16$-grams, and looping relative to the reference;
and \emph{(iii) incomplete coverage} ($P_{\mathrm{under}}$), which penalizes shortfalls in unit counts, semantic content, and available category or temporal fields.
Each penalty takes the largest violation across its checks, with task-specific thresholds estimated from clean rollouts.
Undecodable outputs and outputs without a primary payload receive zero reward.

\paragraph{Content accuracy ($\mathrm{core}_{\nu}$).}
\label{subsec:srvr-content}
The content factor combines a grounded-field score $\Pi_{\nu}$ with a description score $\Gamma_{\nu}$.
When the reference includes descriptions, $\mathrm{core}_{\nu}=(\Pi_{\nu}+\Gamma_{\nu})/2$; otherwise, $\mathrm{core}_{\nu}=\Pi_{\nu}$.
For the document and image tasks used in RLVR, the grounded-field scorer is selected as follows:
\emph{(i) charts and flowcharts} use RMS-$F_1$ and Mermaid $F_1$, respectively (App.~\ref{subsec:eval-protocol-omniparsing});
\emph{(ii) documents} average sequence similarity over reference-present text, formula, table, and reading-order categories;
\emph{(iii) natural images} combine entity text similarity and box IoU with weights $0.7$ and $0.3$, renormalized over available annotations;
and \emph{(iv) other structured fields} use sequence similarity for geometric expressions and normalized SMILES, and text similarity for OCR and music transcriptions.
The description score $\Gamma_{\nu}$ averages text similarity over reference-present descriptive fields.
For both entity and description scoring, text similarity is the mean of token-set $F_1$ and normalized character-sequence similarity after lowercasing and whitespace normalization. This training reward is distinct from the LLM-judged CaptionQA accuracy used to evaluate Cognition on OmniParsingBench (App.~\ref{subsec:eval-protocol-omniparsing}).

\subsection{OmniSchema-Aware On-Policy Distillation}
\label{subsec:app-osad}
\begin{algorithm}[!t]
\caption{\OSAD with self-evolution.}
\label{alg:osad}
\small
\begin{algorithmic}[1]
\Require RLVR policy $\theta_0$; data $\mathcal{D}=\{(x,y^{\star})\}$; rounds $K=6$; round settings $\mathcal{H}_r=\{\eta_r,\tau_r,\tau_{c,r},\tau_{s,r},\alpha_{c,r},\alpha_{s,r}\}$
\State $\bar\theta_1\leftarrow\theta_0$
\For{$r=1,\ldots,K$}
  \State freeze the teacher $\pi_{\bar\theta_r}$; initialize the student base weights from $\bar\theta_r$ and add fresh low-rank adapters
  \For{each global minibatch $\mathcal{B}\subset\mathcal{D}$, partitioned across data-parallel ranks}
    \State sample one rollout $y\sim\pi_\theta(\cdot\mid x)$ per prompt at sampling temperature $T_{\mathrm{samp}}=0.35$
    \State $\mathcal{B}^{+}\leftarrow$ admitted rollout instances pooled across ranks; $N\leftarrow|\mathcal{B}^{+}|$; skip this minibatch if $N=0$
    \State score each $y\in\mathcal{B}^{+}$: student on $x$, teacher on $[y^{\star};x]$ \Comment{Equ.~\ref{equ:osad-pq}}
    \State assign schema roles $\mathcal{C}(y)$ and $\mathcal{S}(y)$; compute the all-token and role-specific terms of $\mathcal{L}_{\mathrm{row}}(y)$ \Comment{Subsec.~\ref{subsec:osad}}
    \State update the adapters on the global objective $\mathcal{L}=N^{-1}\sum_{y\in\mathcal{B}^{+}}\mathcal{L}_{\mathrm{row}}(y)$ with learning rate $\eta_r$
  \EndFor
  \State merge the learned adapters into the student base weights to obtain $\theta_r$
  \State $\bar\theta_{r+1}\leftarrow\operatorname{stopgrad}(\theta_r)$
\EndFor
\State \Return $\theta_K$
\end{algorithmic}
\end{algorithm}

Starting from the RLVR checkpoint, \OSAD replaces sequence-level rewards with the token-level supervision defined in Subsec.~\ref{subsec:osad}.
Alg.~\ref{alg:osad} summarizes the training loop; below we specify its sampling, batch reduction, and round schedule.

\paragraph{Sampling and loss reduction.}
\label{subsec:osad-rollouts}
\label{subsec:osad-batch}
We sample one response per query with sampling temperature $T_{\mathrm{samp}}=0.35$, nucleus cutoff $0.9$, and top-$k=50$. This sampling temperature is distinct from the distillation temperatures in Equ.~\ref{equ:osad-pq} and Equ.~\ref{equ:osad-div}.
The admission gate then admits a response to the loss only if it passes JSON-validity, schema-conformance, box-syntax, and repetition checks; rejected rollouts are masked in place rather than repaired.
To make the global averaging in Equ.~\ref{equ:osad-loss} explicit, let $\mathcal{B}_k^{+}$ be the admitted rollout instances on data-parallel rank $k$, with count $n_k=|\mathcal{B}_k^{+}|$ and loss sum $s_k=\sum_{y\in\mathcal{B}_k^{+}}\mathcal{L}_{\mathrm{row}}(y)$. For a nonzero global count $N=\sum_k n_k$, the objective is
\[
\mathcal{L}=\frac{\sum_k s_k}{N}.
\]
This is a mean over admitted instances, not an unweighted mean of per-rank means when admission counts differ. A rank with no admitted rollout contributes zero to both sums in this expression. An update is skipped only when $N=0$.

\paragraph{Round schedule.}
\label{subsec:iterative-posttraining}
The first round starts from the RLVR checkpoint; subsequent rounds use the preceding student's checkpoint according to Equ.~\ref{equ:osad-refresh}.
The content and structure weights $(\alpha_c,\alpha_s)$ are $(0.50,0.10)$ in round~1, $(0.60,0.10)$ in round~2, and $(0.80,0.08)$ in rounds~3--6, progressively emphasizing content supervision.
The learning rate is $2\times10^{-7}$ for rounds~1--2 and $1.5\times10^{-7}$ for rounds~3--6.

\section{Evaluation Details}
\label{sec:app-eval}

\begin{appintro}
This appendix defines the metrics of every benchmark, describes the evaluation toolkit we implement for OmniParsingBench, and lists the resulting evaluation set.
\end{appintro}
\subsection{Metric Definitions}
\label{subsec:eval-protocol}

\subsubsection{OmniDocBench}
\label{subsec:eval-protocol-omnidocbench}
OmniDocBench v1.6~\citep{OmniDocBench} evaluates document parsing on pages annotated with text blocks, formulas, tables, and reading order.
We follow the official evaluation: each output is converted to Markdown, and the official pipeline matches predicted and reference text blocks, formulas, and tables one-to-one before computing the metrics below.
Text Edit is the normalized edit distance between matched predicted and reference text.
CDM~\citep{UniMERNet} renders the predicted and reference \LaTeX{} formulas and reports the character-level $F_1$ of the rendered images.
TEDS~\citep{PubTabNet} is a similarity based on the tree edit distance between the predicted and reference HTML tables, and TEDS-S ignores cell text.
Reading Order Edit is the normalized edit distance between the predicted component order and the ground-truth reading order.

\subsubsection{PureDocBench}
\label{subsec:eval-protocol-puredocbench}
PureDocBench~\citep{PureDocBench} evaluates the robustness of document parsing to image degradation.
Every page is generated from an HTML/CSS source, which also provides the structured ground truth, and is rendered in three matched versions: a clean rendering, a digitally degraded version, and a physically recaptured version (printed on paper and photographed).
Because the three tracks share identical content and reference answers, score differences isolate the effect of the image condition.
All component metrics---Text Edit, Formula CDM, Table TEDS, and Reading Order Edit---are computed exactly as on OmniDocBench (App.~\ref{subsec:eval-protocol-omnidocbench}), with Overall defined per track as the mean of $(1-\text{Text Edit})\cdot100$, Formula CDM, and Table TEDS; Avg$_3$ is the mean of the three track Overalls.

\subsubsection{OmniParsingBench}
\label{subsec:eval-protocol-omniparsing}

OmniParsingBench~\citep{Logics-Parsing-Omni} evaluates structured parsing on five splits: natural images, graphics (charts and geometric figures), audio, natural video, and text-rich video.
Each split is scored by two components.
\emph{Perception} measures the accuracy of the grounded fields of a parse, such as localization, element extraction, OCR, and ASR, against the ground truth.
\emph{Cognition} is measured by multiple-choice questions (CaptionQA) that an LLM judge answers by reading only the parse. Dimension accuracies are aggregated using the split-specific rules below; text-rich video first averages dimensions within the parsing and report groups, then averages the two group scores.
We use Qwen3-235B-A22B for all LLM judging in this benchmark, including Cognition and model-based Perception grader components.
\emph{Overall} is the mean of Perception and Cognition multiplied by the post-repair parse rate.
Because the benchmark is released without a complete evaluation toolkit, we implement these metrics following the benchmark definitions.
Before scoring, every output is converted to a common representation and schema repair is applied. Outputs that remain unparseable are excluded from the component metrics and penalized through the parse rate; successfully repaired outputs are included.
For each task $t$, let the parse rate $r_t\in[0,1]$ be the fraction of outputs that are parseable after repair if necessary:
\begin{equation}
 r_t=\frac{N_{\mathrm{parsed},t}}{N_{\mathrm{total},t}},\qquad
 O_t=r_t\,\frac{P_t+C_t}{2}.
 \label{equ:eval-omniparsing-overall}
\end{equation}

\paragraph{Natural image.}
Perception evaluates the localization and recognition of three kinds of targets: plain entities, knowledge entities, and text regions.
Predicted and reference targets of each kind are matched separately by the Hungarian algorithm at IoU $\geq0.5$.
The identity of each matched entity is then verified by an LLM judge, and a matched text region is counted as correct when its normalized similarity exceeds $0.7$ or when the judge finds the reference text in the caption.
Each Perception component is the recall after these checks.
Cognition covers the dimensions $\mathcal D_{\mathrm{img}}=$ \{theme, emotion, scene, count, attributes, location, text, style, composition\}:
\begin{equation}
 P_{\mathrm{img}}=\operatorname{Avg}(p_{\mathrm{plain}},p_{\mathrm{knowledge}},p_{\mathrm{text}}),\quad
 C_{\mathrm{img}}=\operatorname{Avg}_{d\in\mathcal D_{\mathrm{img}}}c_d,\quad
 O_{\mathrm{img}}=r_{\mathrm{img}}\operatorname{Avg}(P_{\mathrm{img}},C_{\mathrm{img}}).
    \label{equ:eval-image}
\end{equation}

\paragraph{Graphics.}
Charts and geometric figures are scored separately and then averaged.
For charts, RMS-$F_1$~\citep{DEPLOT} scores statistical charts by optimally matching table entries on header and value similarity, and Mermaid $F_1$ scores structural diagrams by averaging the $F_1$ of labeled nodes and labeled edges.
Chart Cognition covers OCR and data, visual elements, and logical reasoning:
\begin{equation}
 P_{\mathrm{chart}}=\operatorname{Avg}(p_{\mathrm{rms}},p_{\mathrm{mermaid}}),\quad C_{\mathrm{chart}}=\operatorname{Avg}(c_{\mathrm{ocr/data}},c_{\mathrm{visual}},c_{\mathrm{logic}}),\quad O_{\mathrm{chart}}=r_{\mathrm{chart}}\frac{P_{\mathrm{chart}}+C_{\mathrm{chart}}}{2}.
 \label{equ:eval-chart}
\end{equation}

For geometric figures, predictions are canonicalized and scored on positioning, existence of primitives, and relations, and geometry Cognition covers basic elements, geometric relations, and quantitative relations:
\begin{equation}
 P_{\mathrm{geo}}=0.3p_{\mathrm{pos}}+0.3p_{\mathrm{prim}}+0.4p_{\mathrm{rel}},\quad C_{\mathrm{geo}}=\operatorname{Avg}(c_{\mathrm{basic}},c_{\mathrm{relation}},c_{\mathrm{quant}}),\quad O_{\mathrm{geo}}=r_{\mathrm{geo}}\frac{P_{\mathrm{geo}}+C_{\mathrm{geo}}}{2}.
 \label{equ:eval-geometry}
\end{equation}
Graphics Perception and Cognition average the corresponding chart and geometry components; Graphics Overall averages their separately computed, parse-rate-adjusted Overall scores:
\begin{equation}
 P_{\mathrm{graphics}}=\operatorname{Avg}(P_{\mathrm{chart}},P_{\mathrm{geo}}),\quad
 C_{\mathrm{graphics}}=\operatorname{Avg}(C_{\mathrm{chart}},C_{\mathrm{geo}}),\quad
 O_{\mathrm{graphics}}=\operatorname{Avg}(O_{\mathrm{chart}},O_{\mathrm{geo}}).
 \label{equ:eval-graphics}
\end{equation}

\paragraph{Audio.}
Perception combines segmentation, event labeling, and transcription.
Predicted and reference segments are matched one-to-one by temporal IoU, and Time $F_1$ is averaged over the IoU thresholds $\{0.3,0.5,0.7\}$.
Key $F_1$ compares the event labels of overlapping segments.
The concatenated transcript is scored by WER or CER, which is converted to a score by $p_{\mathrm{asr}}=100\,\big(1-\min(1,e_{\mathrm{asr}})\big)$.
Cognition covers information extraction and recognition:
\begin{equation}
 P_{\mathrm{audio}}=\operatorname{Avg}(p_{\mathrm{time}},p_{\mathrm{key}},p_{\mathrm{asr}}),\quad
 C_{\mathrm{audio}}=\operatorname{Avg}(c_{\mathrm{info}},c_{\mathrm{recognition}}),\quad
 O_{\mathrm{audio}}=r_{\mathrm{audio}}\operatorname{Avg}(P_{\mathrm{audio}},C_{\mathrm{audio}}).
    \label{equ:eval-audio}
\end{equation}

\paragraph{Natural video.}
Time $F_1$ and the ASR score are computed as for audio.
For Key $F_1$, action and interaction phrases are matched semantically by an LLM, which also rates the descriptions of segment pairs at temporal IoU $\geq0.5$ from $0$ to $5$ to obtain $p_{\mathrm{visual}}$.
Cognition covers the dimensions $\mathcal D_{\mathrm{vid}}=$ \{object/attribute, event, emotion, audio-visual causality\}:
\begin{equation}
 P_{\mathrm{video}}=\operatorname{Avg}(p_{\mathrm{time}},p_{\mathrm{key}},p_{\mathrm{asr}},p_{\mathrm{visual}}),\quad
 C_{\mathrm{video}}=\operatorname{Avg}_{d\in\mathcal D_{\mathrm{vid}}}c_d,\quad
 O_{\mathrm{video}}=r_{\mathrm{video}}\operatorname{Avg}(P_{\mathrm{video}},C_{\mathrm{video}}).
 \label{equ:eval-video}
\end{equation}

\paragraph{Text-rich video.}
Time $F_1$ is computed as for natural video, and the ASR score as for audio.
For the OCR score, repeated adjacent screens are merged, and the ordered on-screen text is scored by one minus the clipped CER:
\begin{equation}
 P_{\mathrm{textvid}} = \operatorname{Avg}(p_{\mathrm{time}},p_{\mathrm{ocr}},p_{\mathrm{asr}}).
 \label{equ:eval-text-video-perception}
\end{equation}

Cognition averages two groups of dimensions, named as in Tab.~\ref{tab:omniparsing-textvideo}: the parsing dimensions $\mathcal D_{\mathrm{parse}}=$ \{audio information extraction, objects, OCR, attributes and states, events, audio-visual information synthesis, causal reasoning\} and the report dimensions $\mathcal D_{\mathrm{report}}=$ \{core summary, outline, deep content\}:
\begin{equation}
\begin{aligned}
 C_{\mathrm{parse}}&=\operatorname{Avg}_{d\in\mathcal D_{\mathrm{parse}}}c_d,\quad
 C_{\mathrm{report}}=\operatorname{Avg}_{d\in\mathcal D_{\mathrm{report}}}c_d,\\
 C_{\mathrm{textvid}}&=\operatorname{Avg}(C_{\mathrm{parse}},C_{\mathrm{report}}),\quad
 O_{\mathrm{textvid}}=r_{\mathrm{textvid}}\operatorname{Avg}(P_{\mathrm{textvid}},C_{\mathrm{textvid}}).
\end{aligned}
    \label{equ:eval-text-video}
\end{equation}

The leaderboard Avg. is the equal-weight mean of the five split Overall scores:
\begin{equation}
\mathrm{Avg}_{\mathrm{leaderboard}}
=\frac{O_{\mathrm{img}}+O_{\mathrm{graphics}}+O_{\mathrm{audio}}+O_{\mathrm{video}}+O_{\mathrm{textvid}}}{5}.
\label{equ:eval-omniparsing-average}
\end{equation}

\subsubsection{ChemOCR}
ChemOCR~\citep{ChemVLM} evaluates optical chemical structure recognition on $1{,}000$ image--SMILES pairs before scoring exclusions.
A single predicted SMILES is extracted from each output.
Following the official evaluator of ChemVLM~\citep{ChemVLM}, RDKit~\citep{RDKit} parses the predicted and reference SMILES with \texttt{MolFromSmiles}, creates their default \texttt{RDKFingerprint} bit vectors, and computes Tanimoto similarity, $s_i=|A_i\cap B_i|/|A_i\cup B_i|$, where $A_i$ and $B_i$ are the sets of active fingerprint bits.
In practice, the official scorer excludes $10$ samples with unparseable reference SMILES from the $1{,}000$-sample benchmark. On the remaining $N=990$ samples, we report $\mathrm{Avg.\ Sim.}=100\sum_i s_i/N$ and $\mathrm{Tani@1.0}=100\sum_i\mathbf{1}[s_i=1]/N$, which measure the overall similarity and the number of perfectly correct samples, respectively.

\subsubsection{PDMX-Synth}
\label{subsec:eval-protocol-pdmx}
PDMX-Synth~\citep{LEGATO} evaluates full-page optical music recognition on $800$ rendered score images paired with ABC transcriptions.
Following LEGATO~\citep{LEGATO}, we report the normalized ABC edit distance at the character, symbol, and line levels (CER, SER, and LER).

\subsection{Evaluation Set}
\label{subsec:app-eval-set}

The camera-video subset is excluded because of annotation issues; of the remaining $3{,}874$ samples, $136$ have unavailable media, which leaves $3{,}738$ (Tab.~\ref{tab:omniparsing-gt-coverage}).
No sample IDs were added, and the Cognition labels of the retained samples are unchanged.

\begin{table}[!htbp]
\centering
\small
\setlength{\tabcolsep}{8pt}
\caption{\textbf{OmniParsingBench sample counts within our evaluation scope.} Natural video counts only the general subset.}
\label{tab:omniparsing-gt-coverage}
\begin{tabular}{lrrrrrr}
\toprule
 & Natural image & Graphics & Audio & Natural video & Text-rich video & \textbf{Total} \\
\midrule
Original       & 1{,}000 & 1{,}000 & 1{,}014 & 601 & 259 & \textbf{3{,}874} \\
Media excluded & 0       & 0       & 76      & 51  & 9   & \textbf{136} \\
Kept           & 1{,}000 & 1{,}000 & 938     & 550 & 250 & \textbf{3{,}738} \\
\bottomrule
\end{tabular}
\end{table}

\section{Additional Encoder Ablation Curves}
\label{sec:app-ablations}

\begin{appintro}
This section provides additional vision and ASR training curves for the two encoder studies in Subsec.~\ref{subsec:eval-analysis}.
\end{appintro}

\begin{figure}[!t]
\centering
\includegraphics[width=\textwidth]{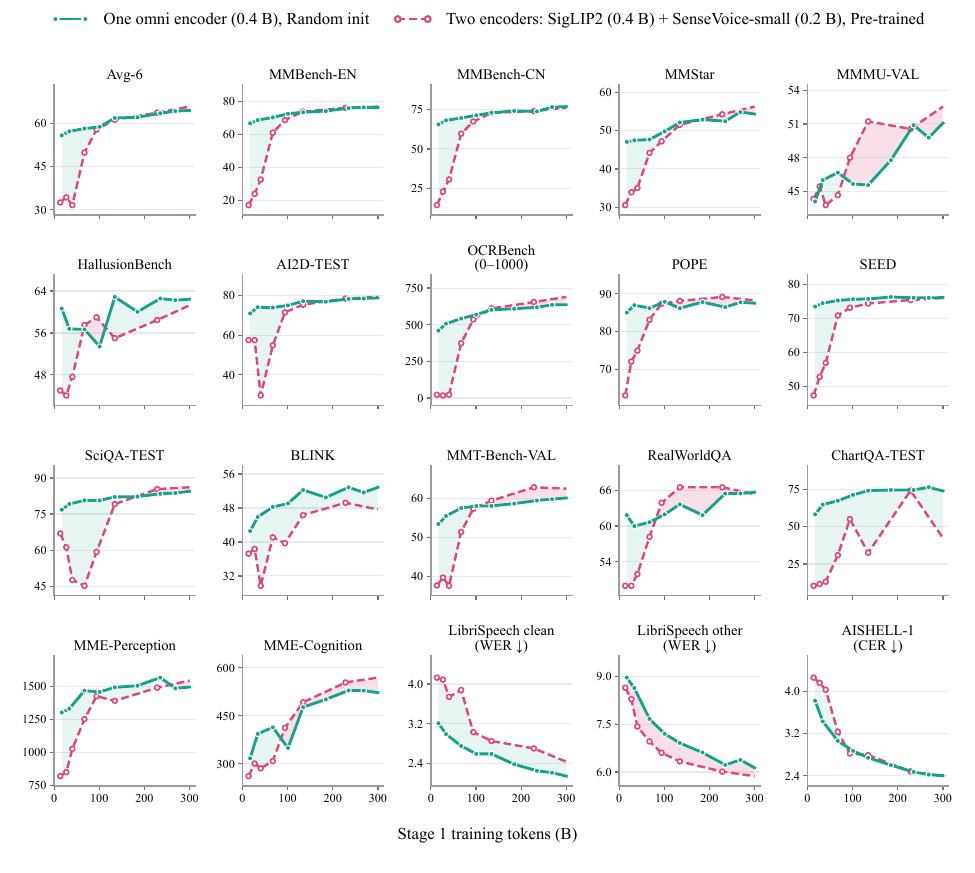}
\caption{
    \textbf{Unified encoder versus separate towers.}
    Green solid curves denote the randomly initialized unified encoder ($0.4$\,B); pink dashed curves denote pre-trained SigLIP-2~\citep{SigLIP-2} and SenseVoice-small~\citep{FunAudioLLM} towers ($0.6$\,B combined) with random projectors.
    Panels report individual vision benchmarks, the general multimodal average (Avg-6), and ASR error rates.
}
\label{fig:ablation-encoder-arch}
\end{figure}

\begin{figure}[!t]
\centering
\includegraphics[width=\textwidth]{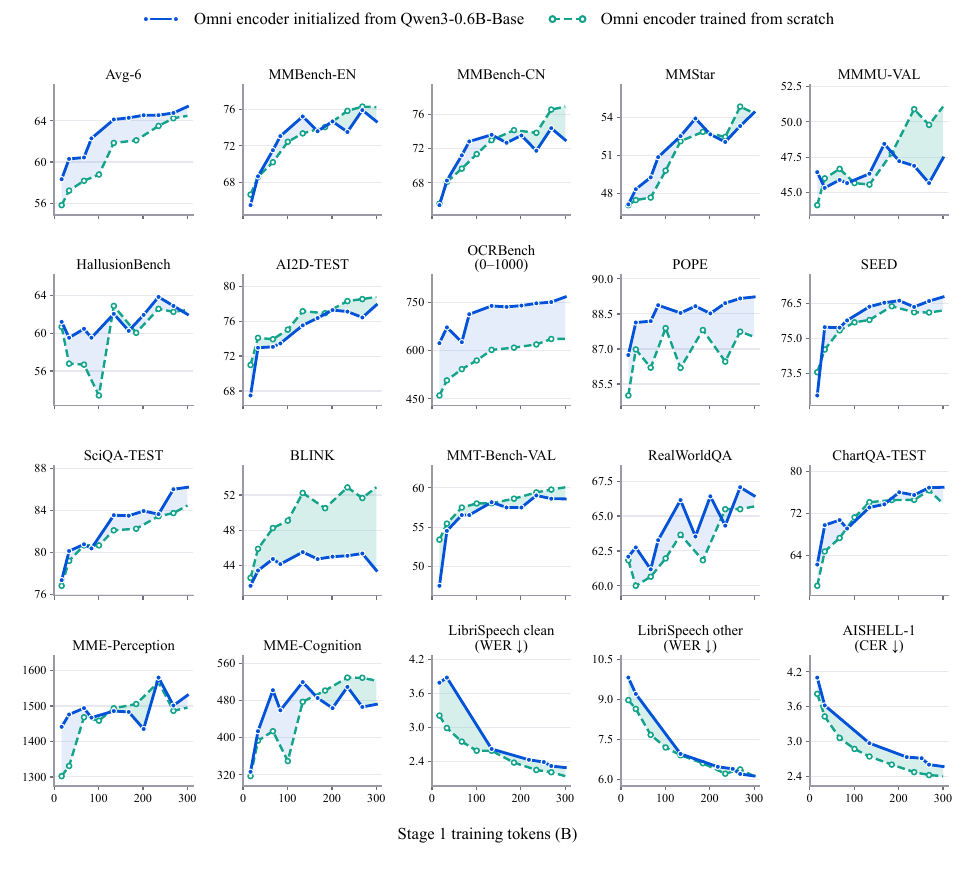}
\caption{\textbf{Encoder initialization on vision and ASR benchmarks.}
Blue solid curves denote the encoder initialized from Qwen3-0.6B-Base, and green dashed curves denote the encoder trained from scratch. Shaded bands indicate the performance gap, colored by the better-performing branch. Lower WER and CER indicate better ASR performance.}
\label{fig:ablation-encoder-init}
\end{figure}

Fig.~\ref{fig:ablation-encoder-arch} compares the randomly initialized unified encoder with separate pre-trained vision and speech towers.
Fig.~\ref{fig:ablation-encoder-init} compares text-LM initialization with training the unified encoder from scratch across vision and ASR benchmarks.

Across both figures, blue denotes the text-initialized unified encoder, green the randomly initialized unified encoder, and pink the pre-trained towers.
Circular markers indicate evaluated checkpoints, and shaded bands indicate which branch performs better between the curves.
The horizontal axis counts Stage~1 training tokens in billions.
Higher vision scores and lower ASR WER or CER indicate better performance.

\FloatBarrier
\section{Qualitative Examples}
\label{sec:app-examples}

\begin{appintro}
Each example in this appendix comprises the input, the parsed elements or segments, and the rendered parse. 
\end{appintro}

\newcommand{\casesep}{\par\vspace{14pt}}

\begin{figure}[!hp]
\centering
\includegraphics[width=0.93\textwidth,height=0.86\textheight,keepaspectratio]{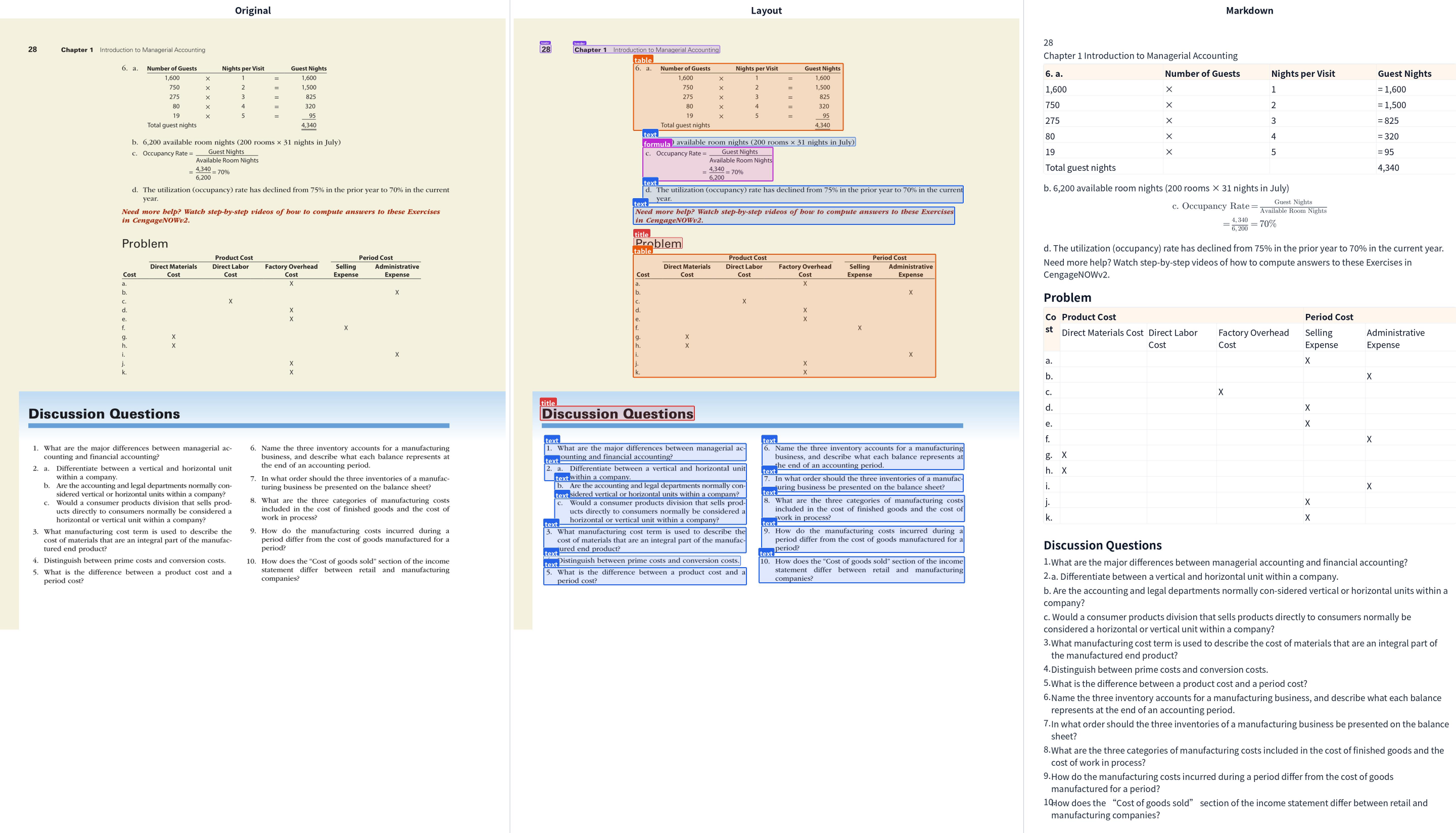}
\caption{\textbf{Document.}\enspace  A conventional single-column page: the parse recovers one linear reading order over headings, body paragraphs, and inline elements.}
\label{fig:case-doc-single-column}
\casesep
\includegraphics[width=0.93\textwidth,height=0.86\textheight,keepaspectratio]{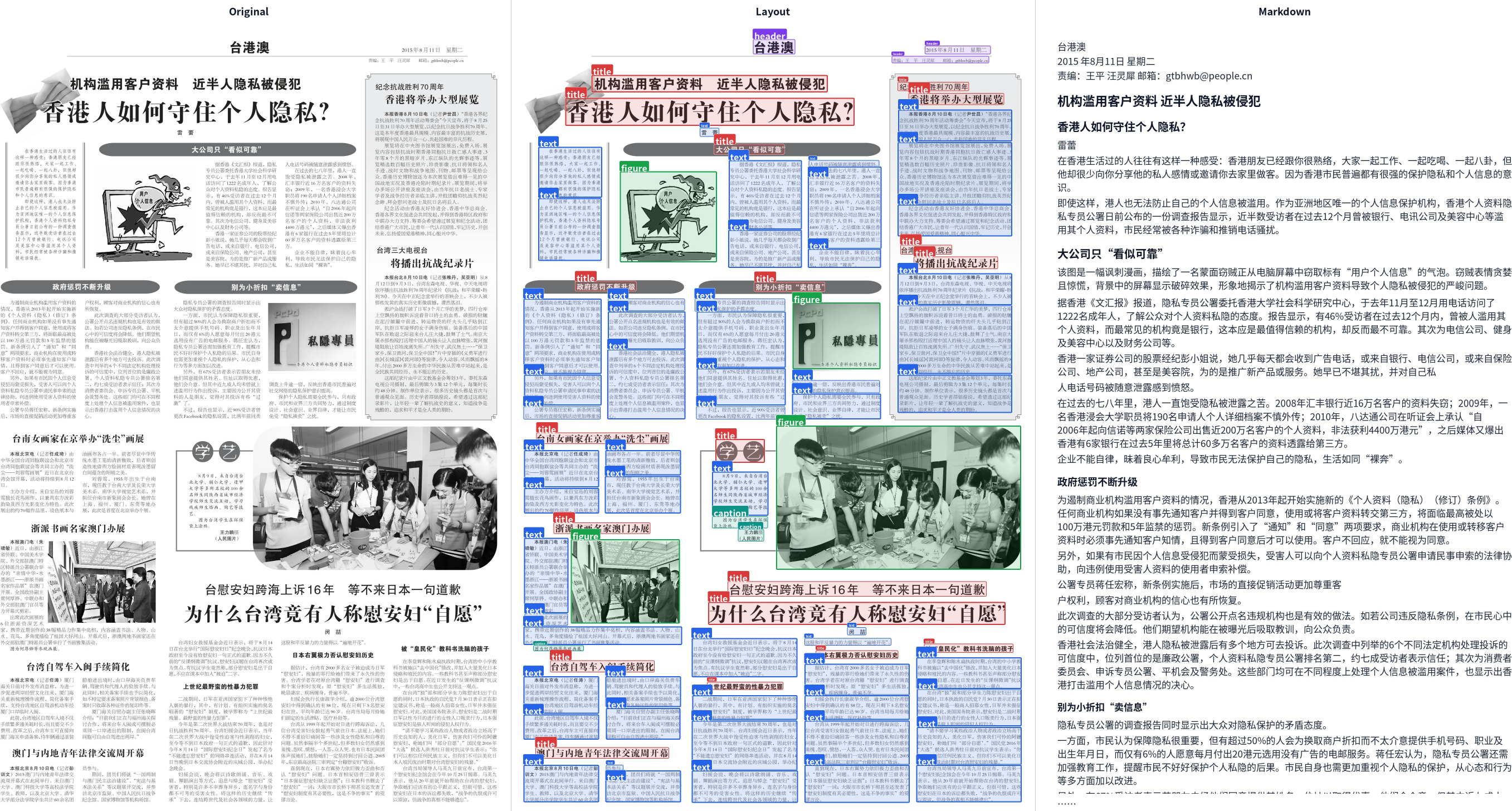}
\caption{\textbf{Document.}\enspace A newspaper page, the hardest layout class here: irregular multi-column blocks, headlines spanning several columns, captioned photographs, and articles that continue across blocks, all of which the parse has to segment and order.}
\label{fig:case-doc-newspaper}
\end{figure}

\begin{figure}[!p]
\centering
\includegraphics[width=\textwidth,height=0.86\textheight,keepaspectratio]{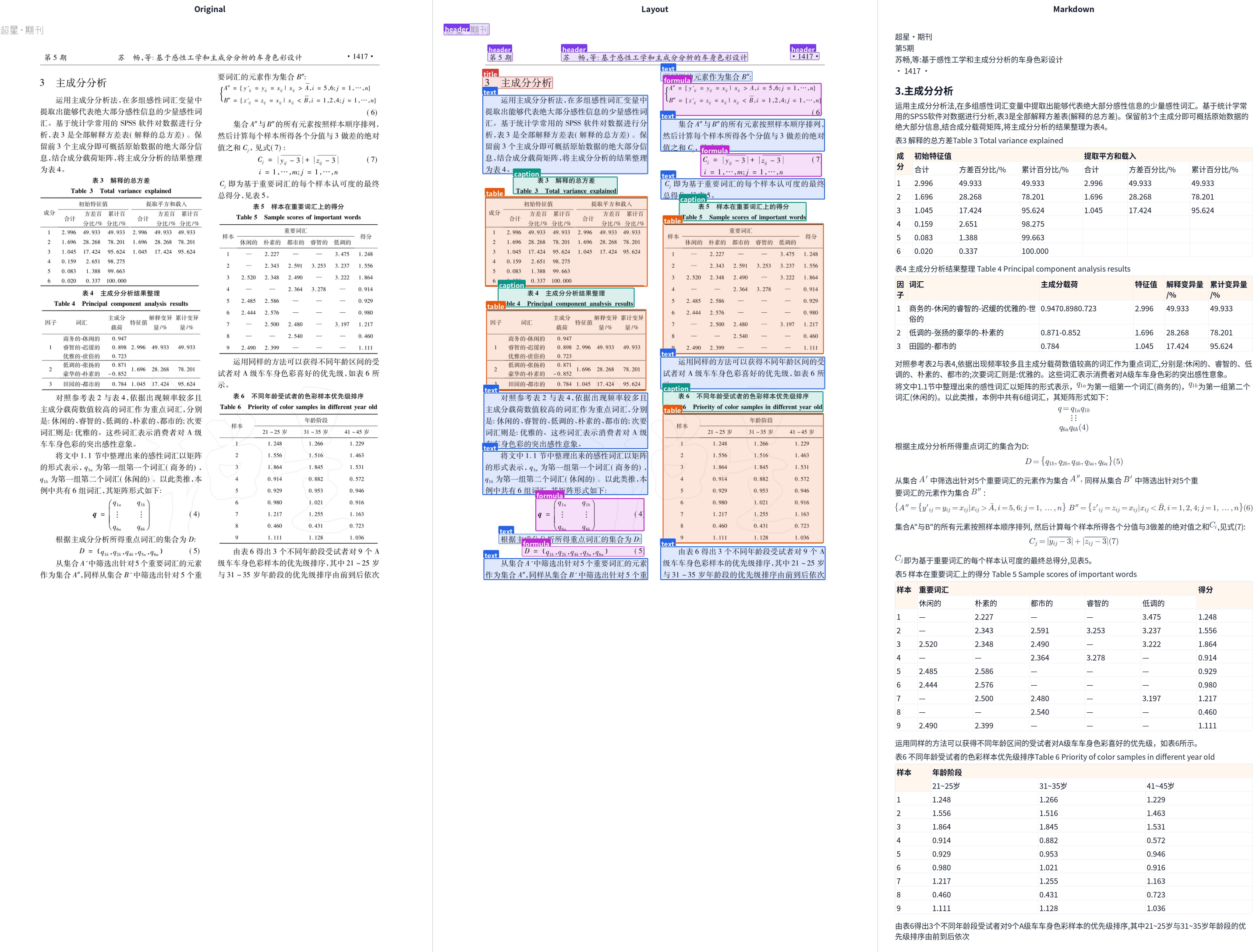}
\caption{\textbf{Document.}\enspace A two-column page combining several element categories---text, formulas, tables, and figures. The parse has to assign every element to its column and serialize the two columns into a single reading order rather than interleaving them line by line.}
\label{fig:case-doc-two-column}
\casesep
\includegraphics[width=1.0\textwidth,keepaspectratio]{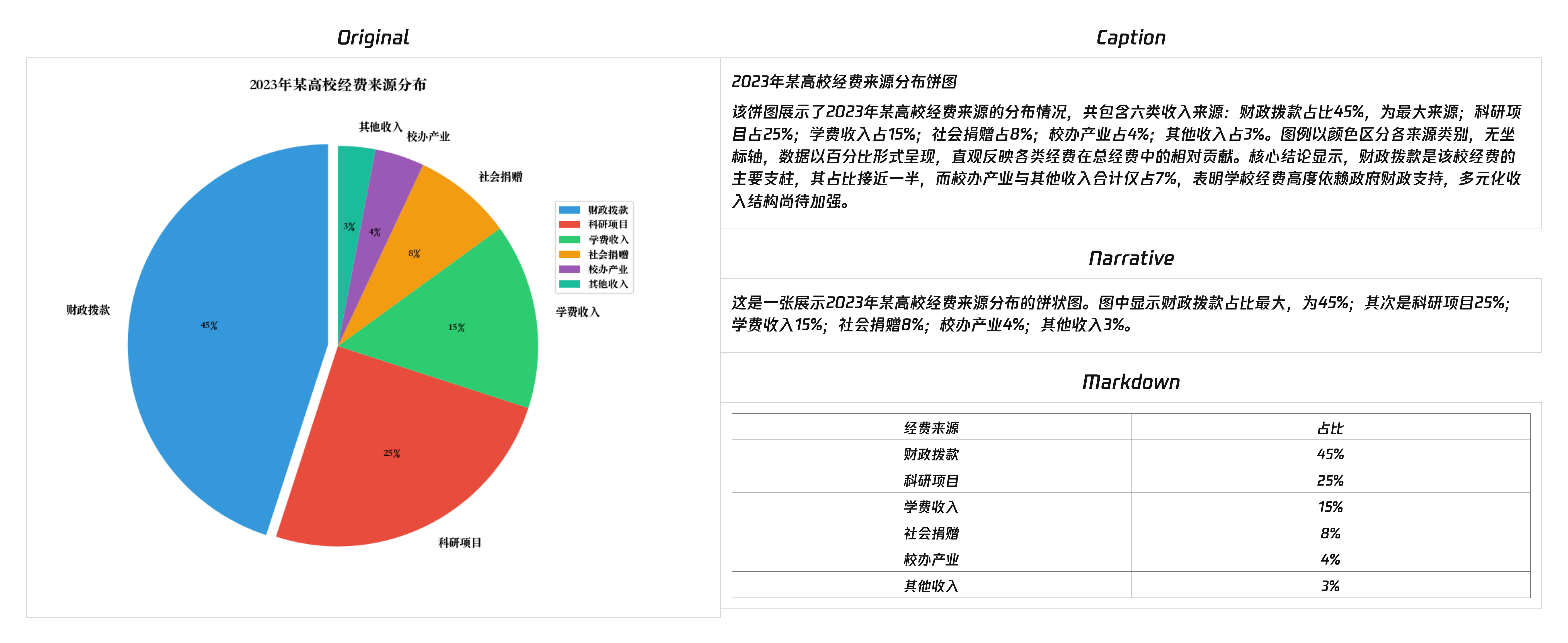}
\caption{\textbf{Chart.}\enspace Pie Chart of University Funding Sources.}
\label{fig:case-chart-221}
\end{figure}

\begin{figure}[!p]
\centering
\includegraphics[width=1.0\textwidth,keepaspectratio]{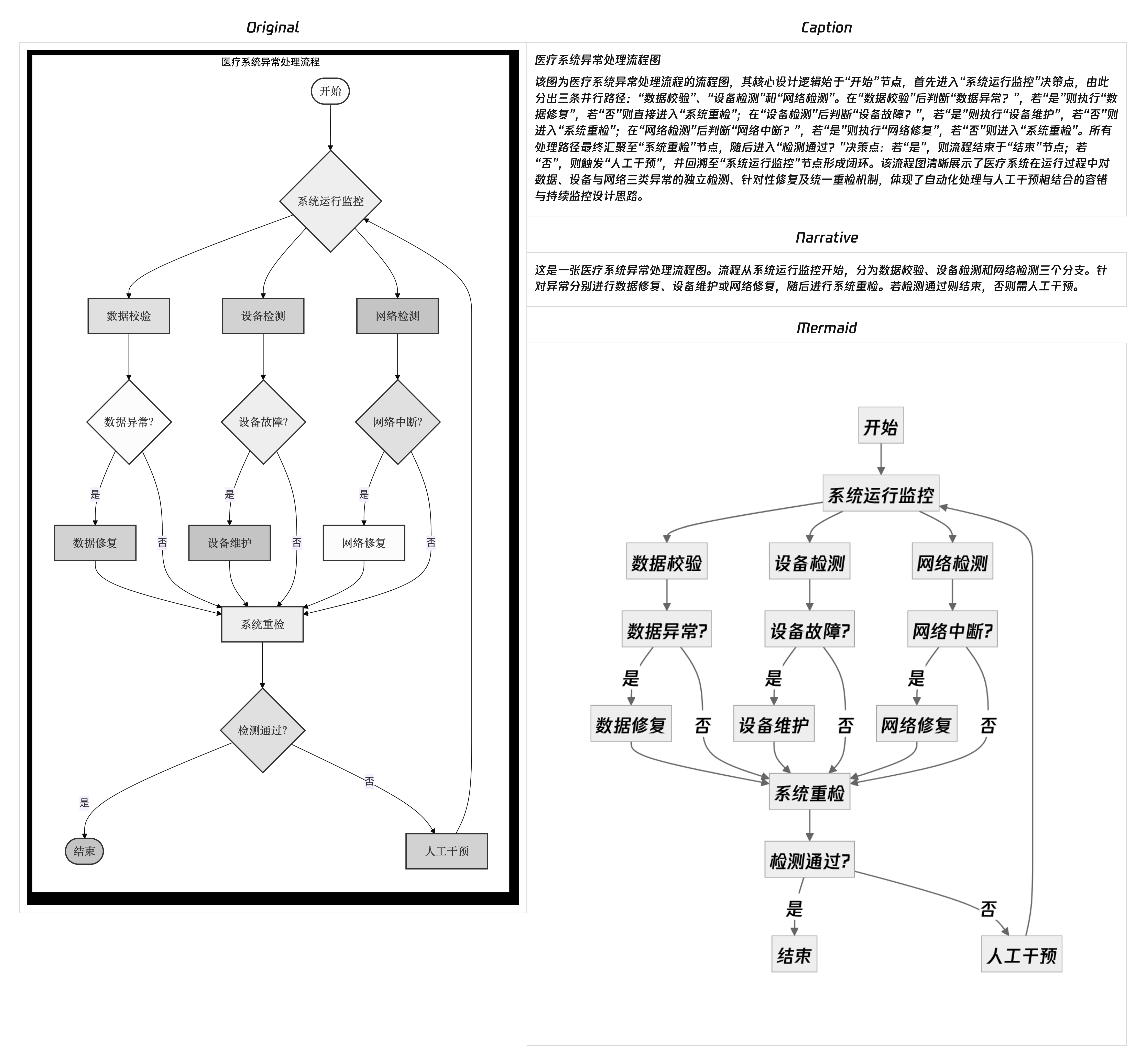}
\caption{\textbf{Flowchart.}\enspace Anomaly-Handling Workflow of a Medical System.}
\label{fig:case-flowchart-19}
\casesep
\includegraphics[width=1.0\textwidth,keepaspectratio]{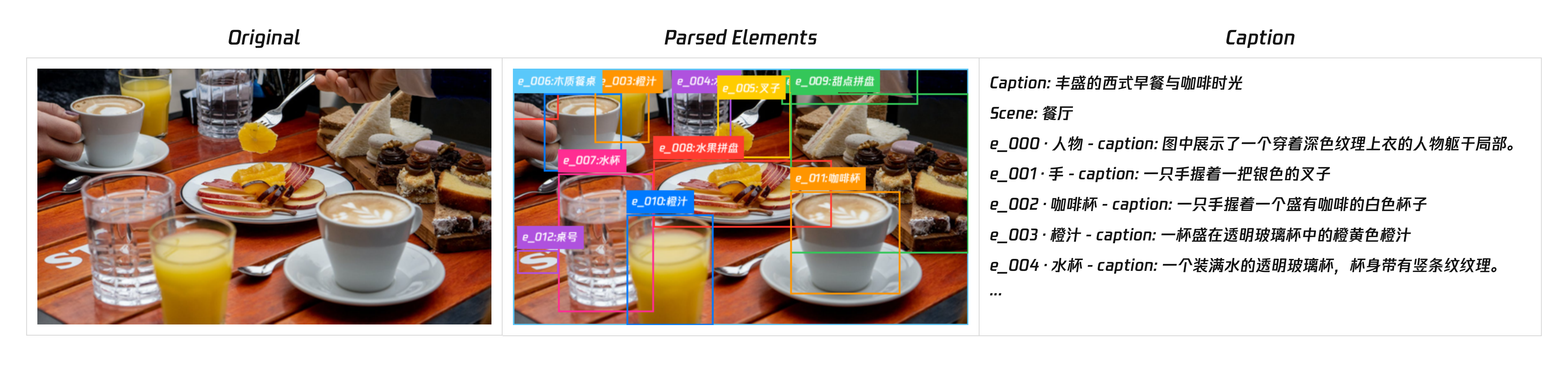}
\caption{\textbf{Natural Image.}\enspace Western Breakfast Spread with Coffee.}
\label{fig:case-natural-image-554}
\end{figure}

\begin{figure}[!p]
\centering
\includegraphics[width=1.0\textwidth,keepaspectratio]{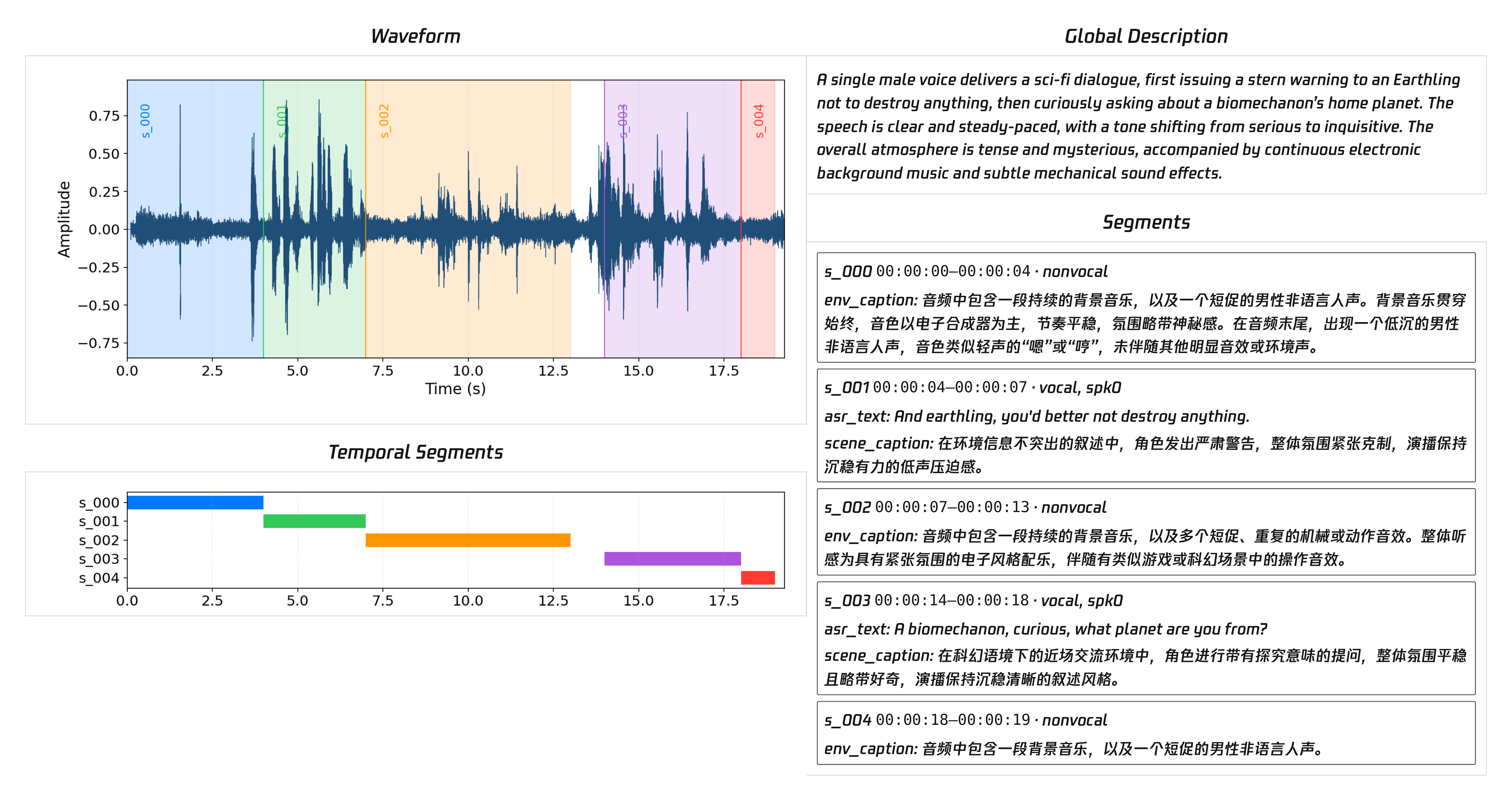}
\caption{\textbf{Audio.}\enspace Sci-Fi Dialogue with Mechanical Sound Effects.}
\label{fig:case-audio-692}
\casesep
\includegraphics[width=1.0\textwidth,keepaspectratio]{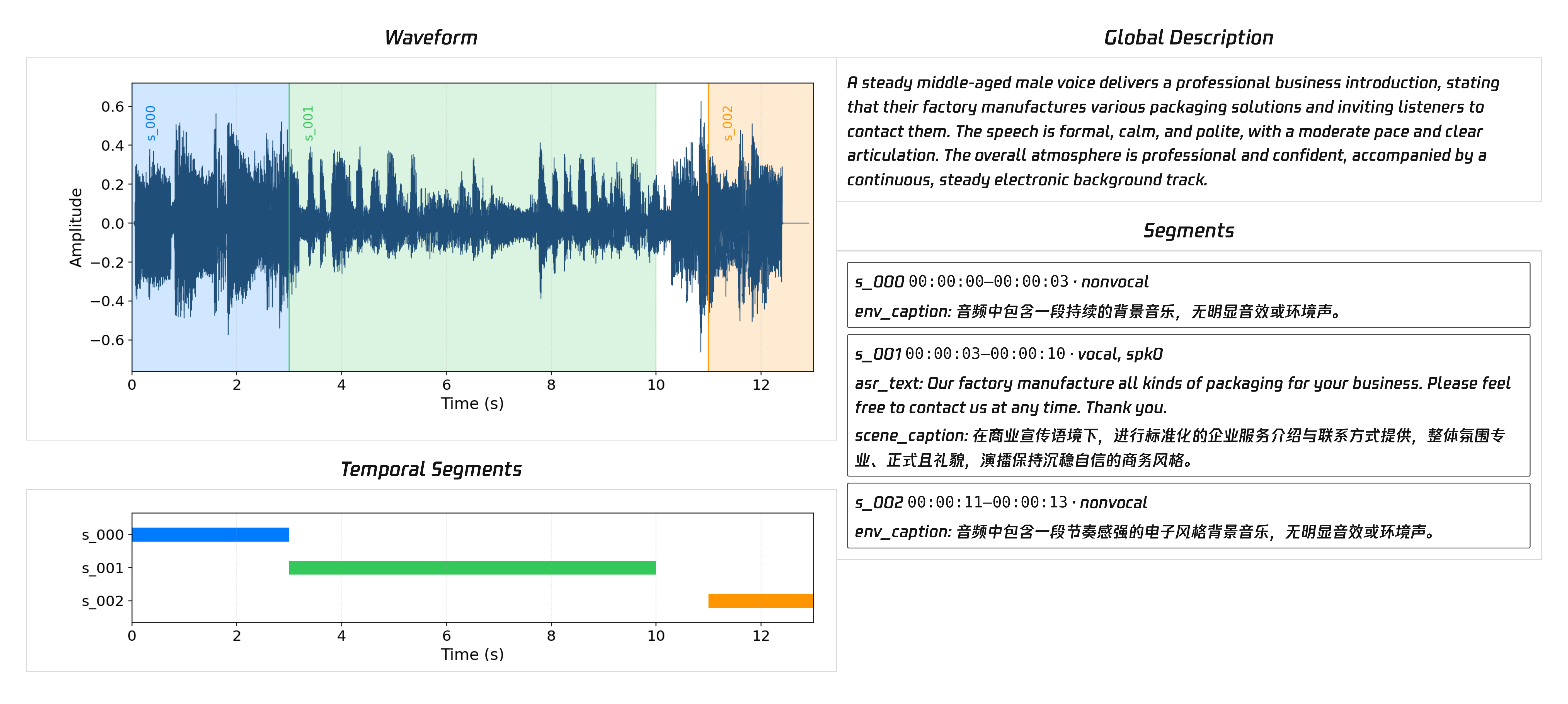}
\caption{\textbf{Audio.}\enspace Factory Packaging Services Introduction.}
\label{fig:case-audio-207}
\end{figure}

\begin{figure}[!p]
\centering
\includegraphics[width=1.0\textwidth,keepaspectratio]{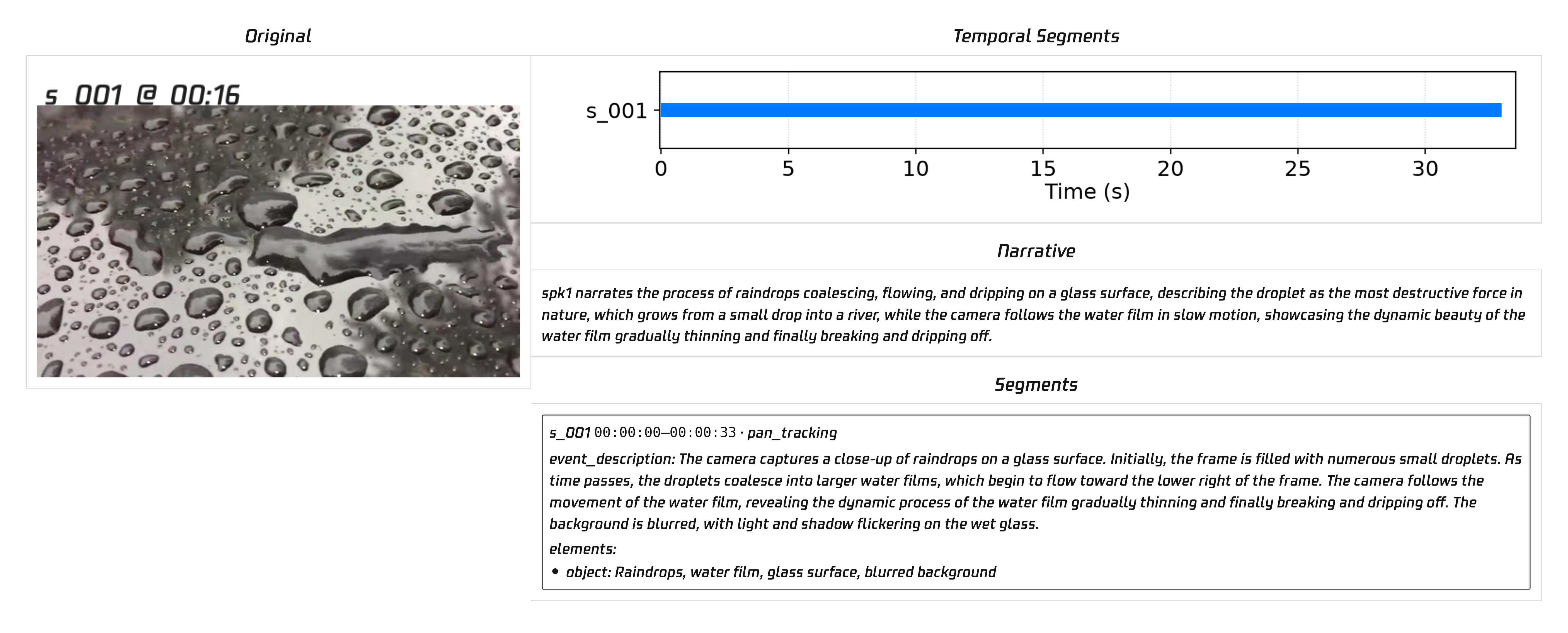}
\caption{\textbf{Natural Video.}\enspace Raindrops Coalescing on a Glass Surface.}
\label{fig:case-natural-video-3}
\casesep
\includegraphics[width=1.0\textwidth,keepaspectratio]{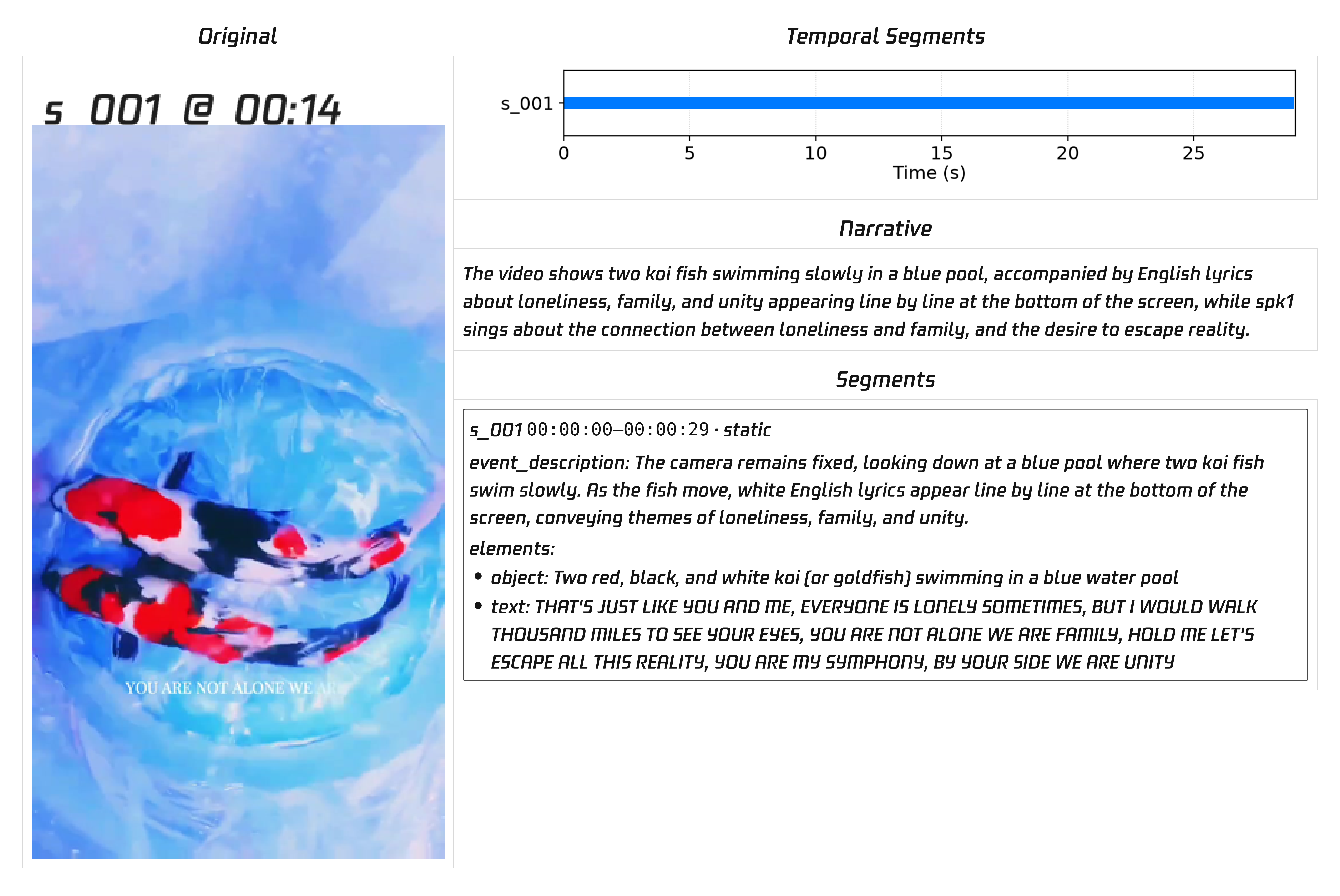}
\caption{\textbf{Natural Video.}\enspace Koi Fish Swimming in a Pool with Song Lyrics.}
\label{fig:case-natural-video-14}
\end{figure}

\begin{figure}[!p]
\centering
\renewcommand{\casesep}{\par\vspace{8pt}}
\includegraphics[width=1.0\textwidth,keepaspectratio]{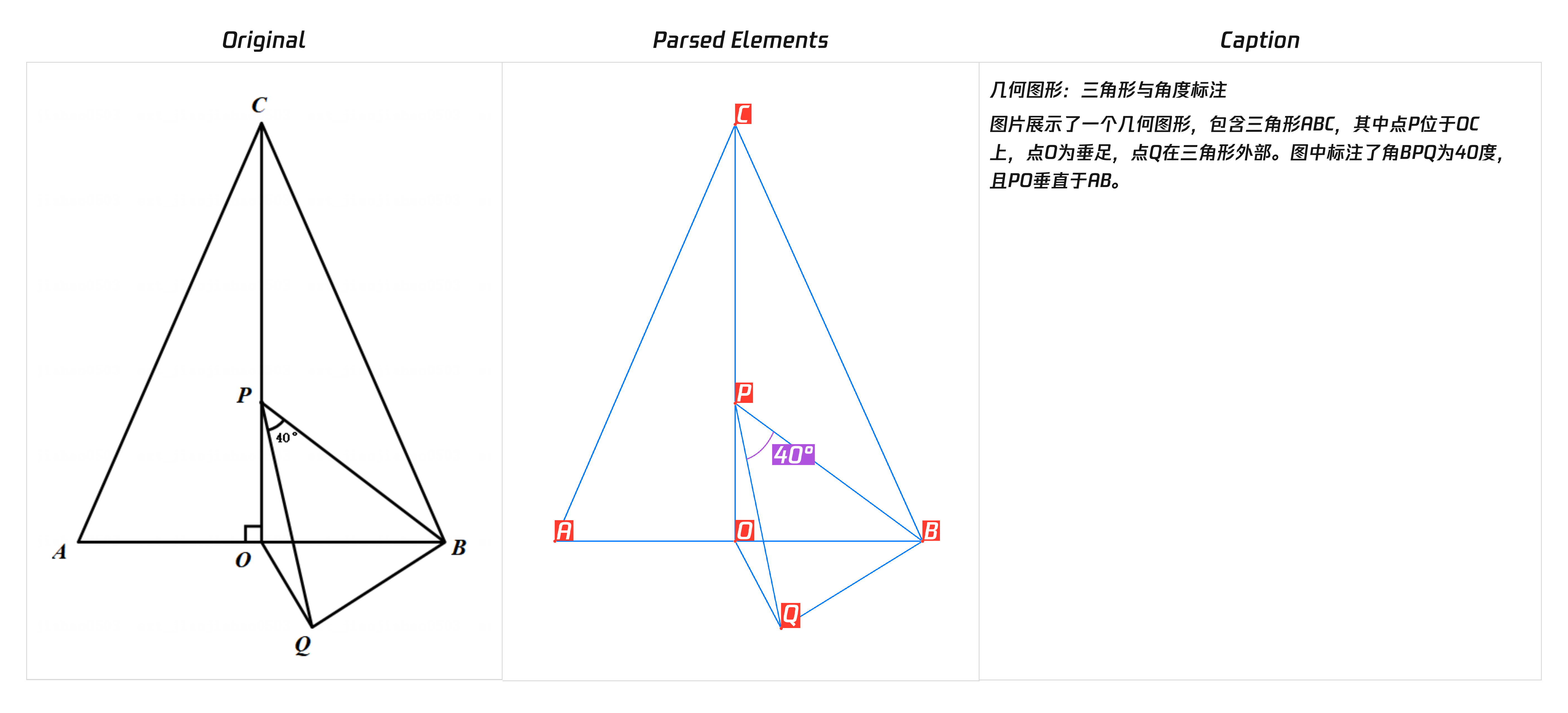}
\caption{\textbf{Geometry.}\enspace Triangle with Perpendicular Foot and Angle Annotation.}
\label{fig:case-geometric-771}
\casesep
\includegraphics[width=1.0\textwidth,keepaspectratio]{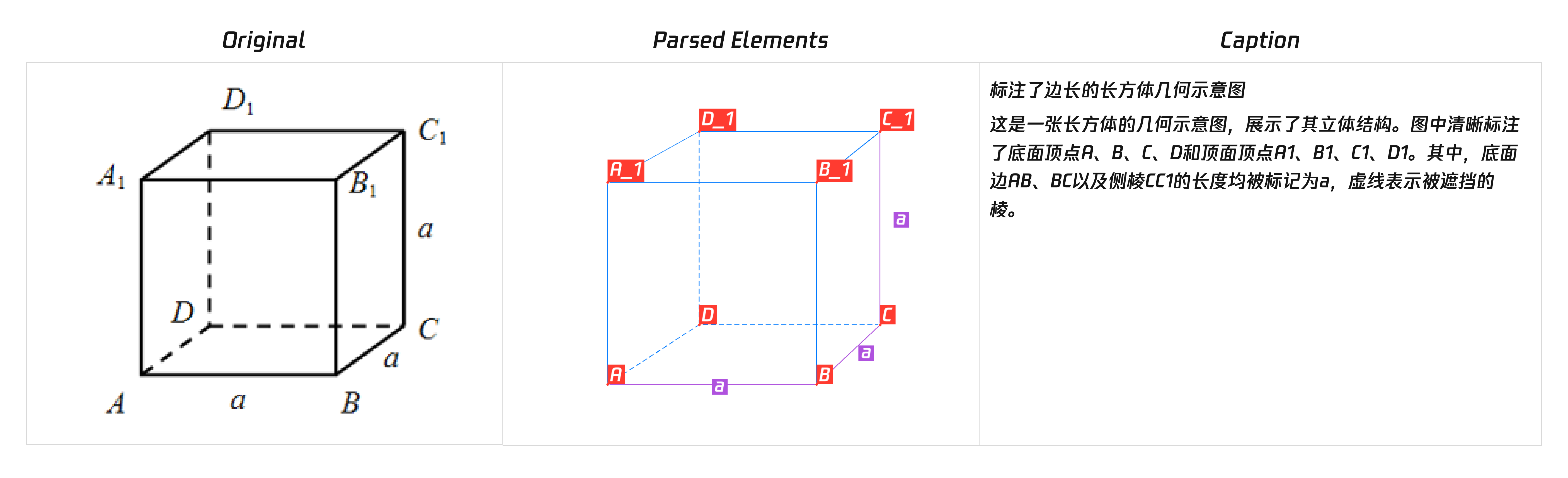}
\caption{\textbf{Geometry.}\enspace Cuboid with Labeled Vertices and Edge Lengths.}
\label{fig:case-geometric-616}
\casesep
\includegraphics[width=1.0\textwidth,keepaspectratio]{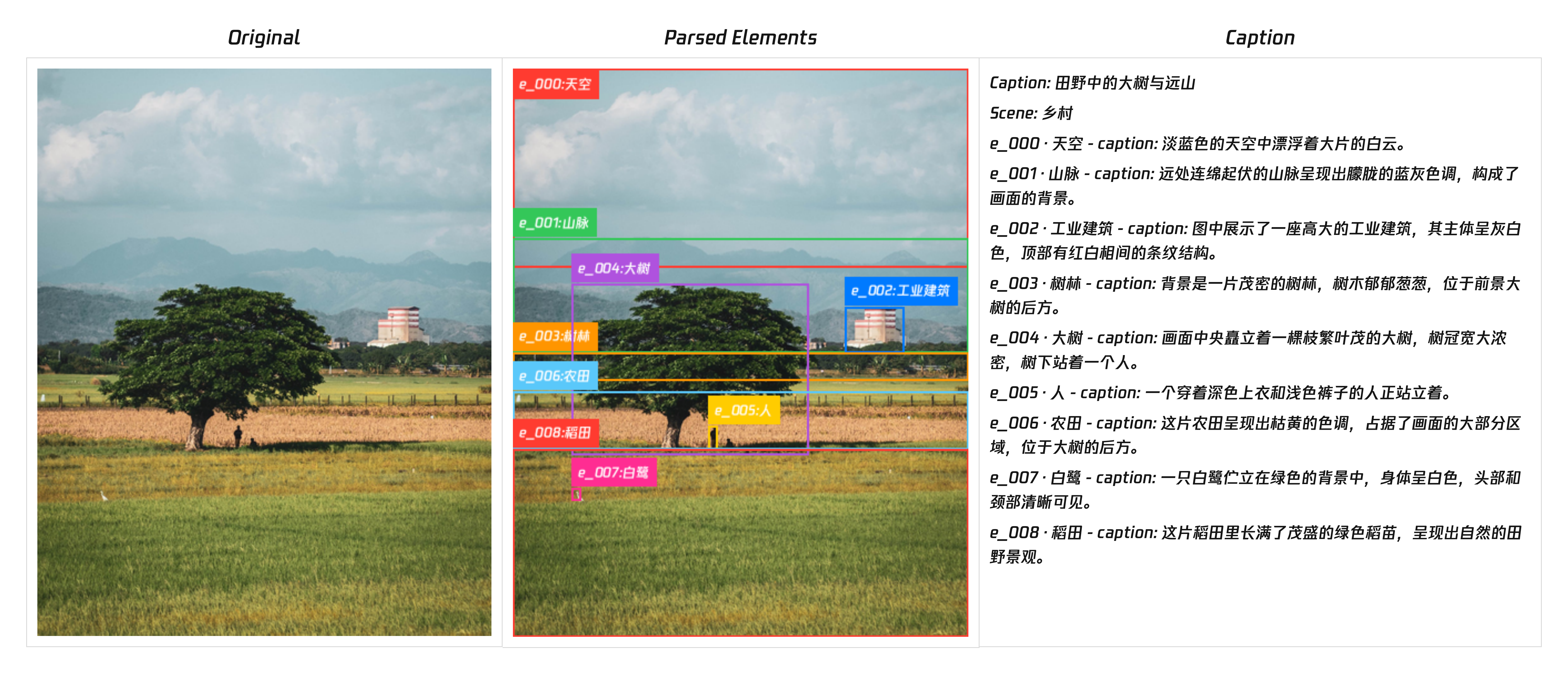}
\caption{\textbf{Natural Image.}\enspace Rural Landscape with a Large Tree and Distant Mountains.}
\label{fig:case-natural-image-716}
\end{figure}

\begin{figure}[!htbp]
\centering
\includegraphics[width=\textwidth,height=0.86\textheight,keepaspectratio]{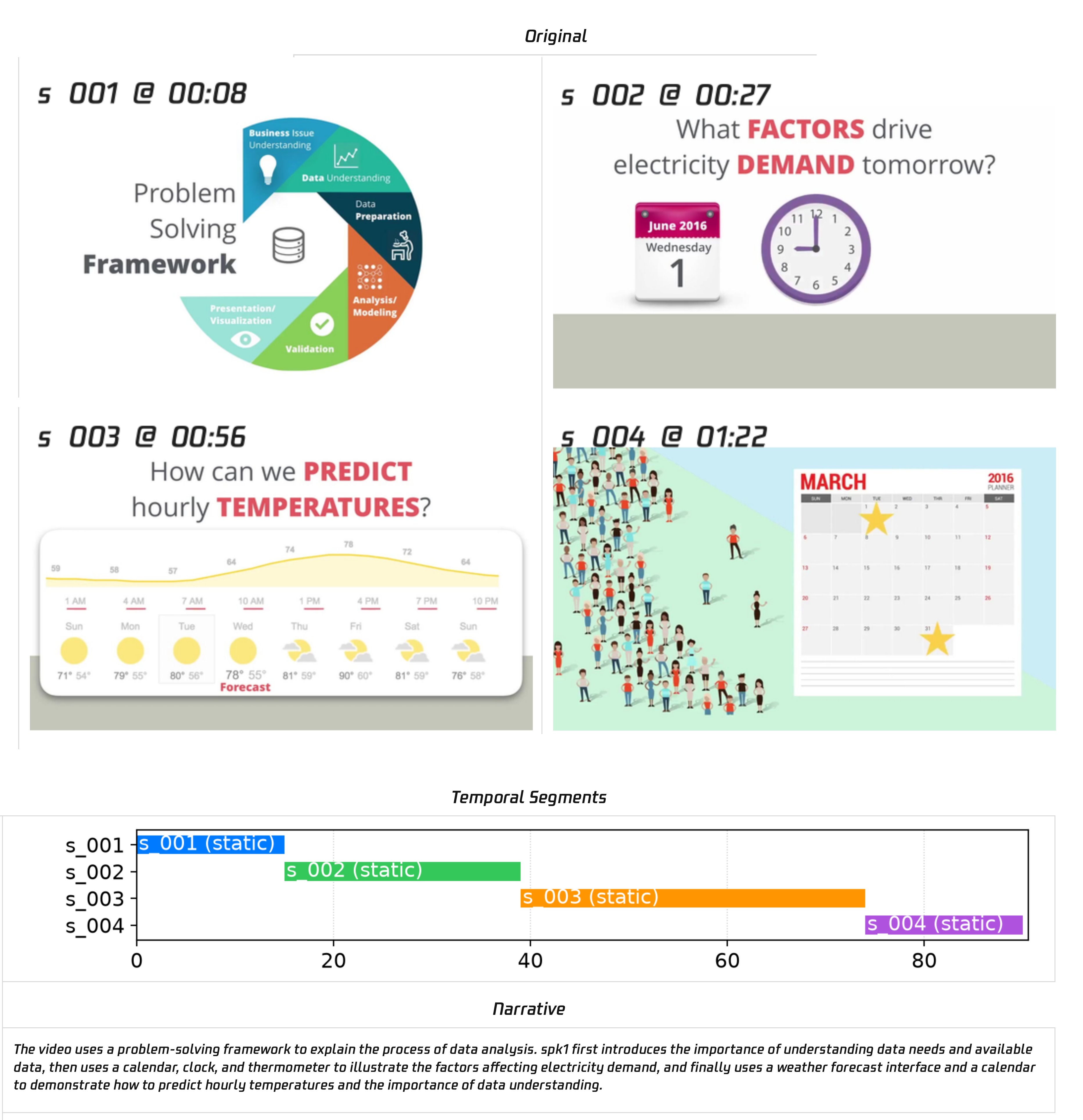}
\caption{\textbf{Text-Rich Video.}\enspace Data Analysis Problem-Solving Framework. Frames, timeline, and narrative; the report and segments follow on the next page.}
\label{fig:case-textrich-video-41}
\end{figure}
\begin{figure}[!htbp]
\ContinuedFloat
\centering
\includegraphics[width=\textwidth,height=0.86\textheight,keepaspectratio]{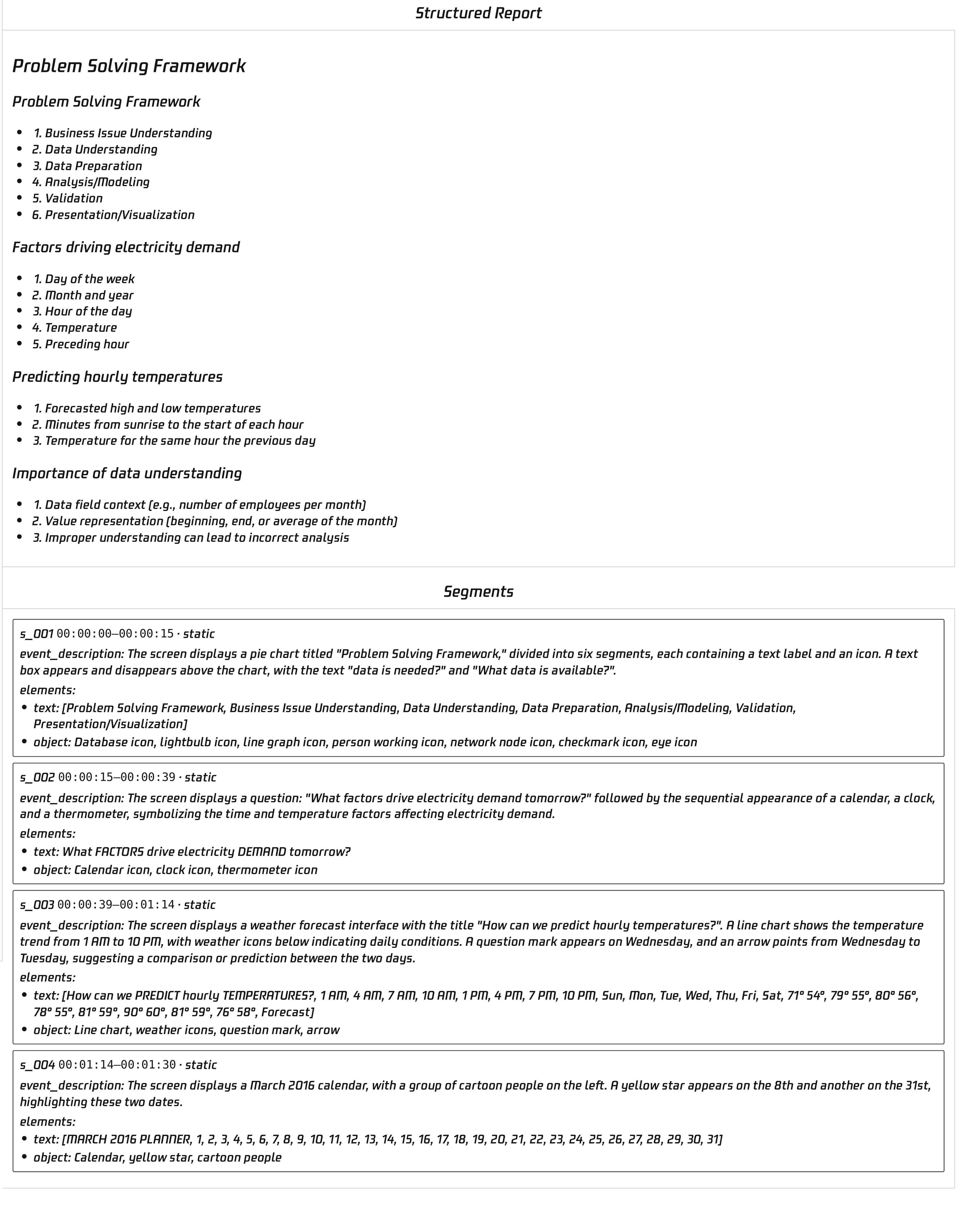}
\caption[]{\textbf{Text-Rich Video.}\enspace Data Analysis Problem-Solving Framework (continued).}
\end{figure}

\begin{figure}[!htbp]
\centering
\includegraphics[width=\textwidth,height=0.86\textheight,keepaspectratio]{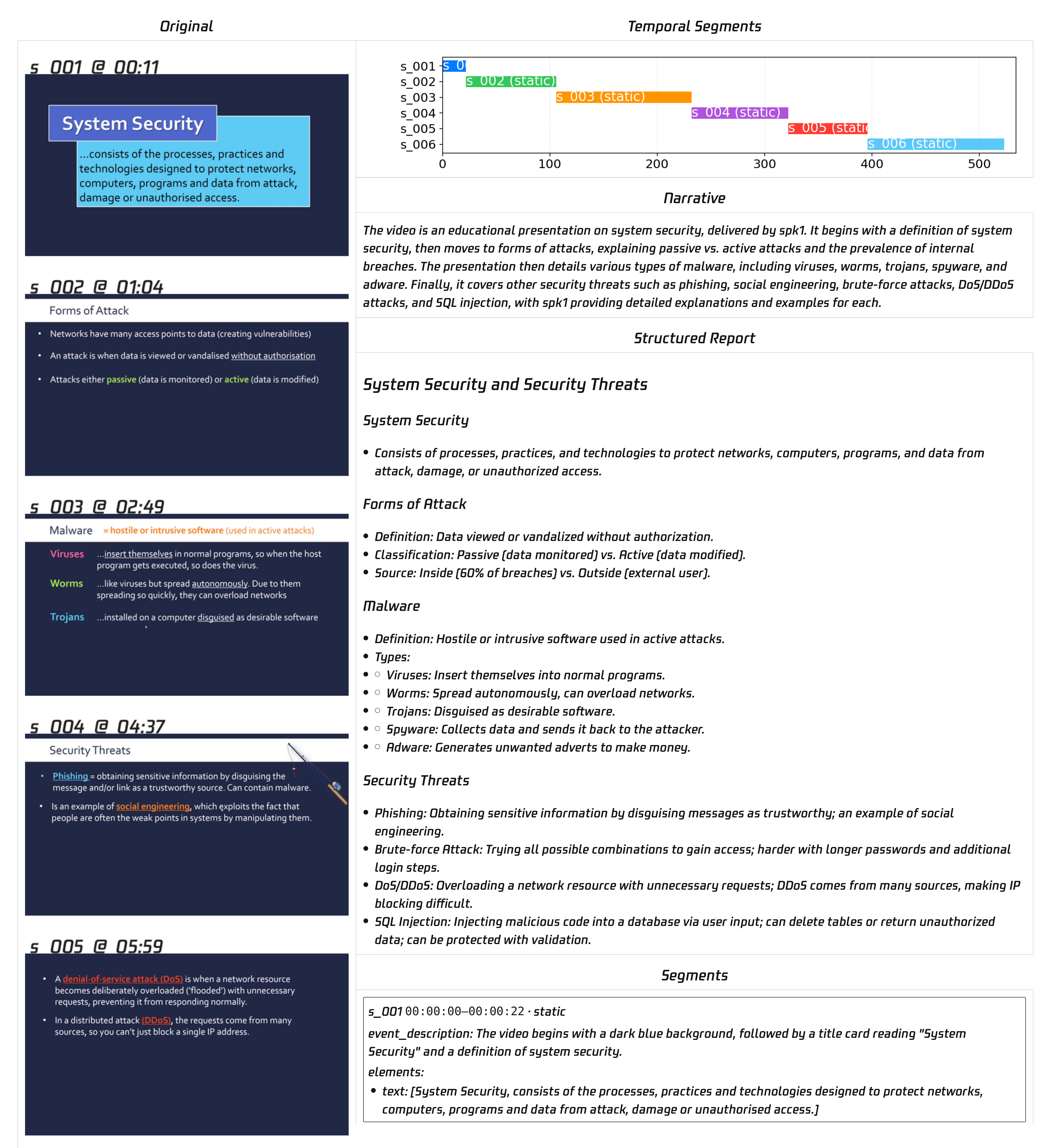}
\caption{\textbf{Text-Rich Video.}\enspace System Security Course on Attack Types and Threats. Frames, timeline, narrative, report, and the first segment; the last frame and the remaining segments follow on the next page.}
\label{fig:case-textrich-video-230}
\end{figure}
\begin{figure}[!htbp]
\ContinuedFloat
\centering
\includegraphics[width=\textwidth,height=0.86\textheight,keepaspectratio]{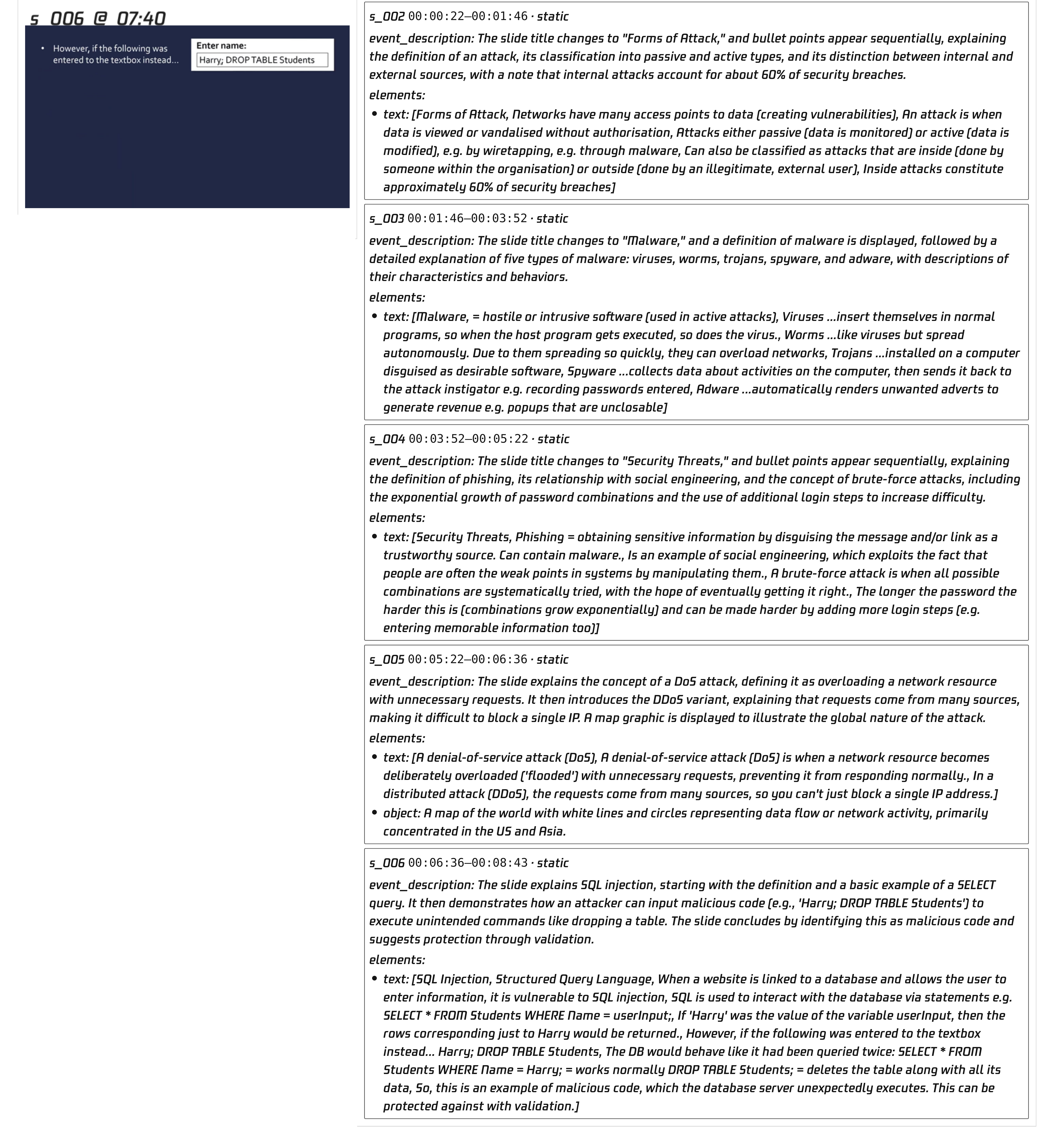}
\caption[]{\textbf{Text-Rich Video.}\enspace System Security Course on Attack Types and Threats (continued).}
\end{figure}

\end{document}